# Evidence and Intervention: A Coupled Active-Inference Extension of Rational Speech Act Models

Yonghyeon Gwon[1], Elliot Murphy[2,3,*], Chun Kee Chung[1,*]

1. Neuroscience Research Institute, Medical Research Center, Seoul National University, Seoul, Republic of Korea.

2. Vivian L. Smith Department of Neurosurgery, McGovern Medical School, UTHealth, Houston, TX, USA

3. Texas Institute for Restorative Neurotechnologies, UTHealth, Houston, TX, USA

* Corresponding authors: elliot.murphy@uth.tmc.edu; chungc@snu.ac.kr

**Abstract**

Identical utterance choices can arise from different communicative causes, and identical interpretations can leave different traces in what a listener learns. Rational Speech Act (RSA) models treat interpretation as inference over speaker meaning, but standard one-shot RSA does not intrinsically distinguish these causal update targets. We develop a coupled active-inference model of dialogue in which a listener's likelihood for an utterance is the policy distribution attributed to the speaker, placing the speaker's expected free energy within the listener's variational free energy. Each utterance is therefore both evidence about and an intervention on a partner. The model separates four updates that RSA approaches typically collapse: inference about the partner's current pragmatic state; learning of partner-specific parameters; prospective evaluation of clarification or repair; and a policy prior shaped by habit and a context-sensitive price of time. Under one-step, exact-inference restrictions, the model recovers the RSA speaker and listener, with RSA as the restricted single-turn limit of the coupled process. Outside these restrictions, the updates obey distinct rules and timescales, predicting which adaptations persist, remain partner-specific, or transfer. Worked examples show audience design reversing after clarification, self-confirming misunderstanding in which both interlocutors have low free energy while disagreeing about reference, and rational closing before uncertainty is resolved. By explaining these effects through active inference, pragmatics becomes a theory of communicative control as much as interpretation. Consistent with critiques of equating natural language with communication, the model treats communication as a downstream use of linguistic structure and recasts production and interpretation as coupled inference: speaking is both an intervention on a partner and an epistemic action that samples evidence for the speaker's model of that partner.

## 1. Introduction

"A relationship, I think, is like a shark. It has to constantly move forward or it dies."
— *Annie Hall* (1977), dir. Woody Allen

Ostensive-inferential linguistic communication requires interlocutors to infer the hidden causes of each other's observable behavior and to select actions whose consequences depend on how those actions will be interpreted (Scott-Phillips, 2015; Sperber & Wilson, 1995). A listener who hears an utterance may need to infer its intended referent, disambiguate its syntactic structure, represent its discourse goal and common-ground commitments, and assess its social relevance (Clark, 2013; Unger & Buschmeier, 2025). A speaker faces the complementary problem of selecting an utterance given a communicative goal and a model of the listener. The Rational Speech Act (RSA) framework formalizes this interdependence by linking a literal listener $L_0$, a pragmatic speaker $S_1$, and a pragmatic listener $L_1$ in a recursive Bayesian model (Degen, 2023; Frank & Goodman, 2012; Goodman & Frank, 2016).

Contemporary RSA is substantially broader than its original reference-game formulation. It supports graded pragmatic inference and provides a common formal language for aspects of semantic processing, prior beliefs and communicative utility. Recent extensions of RSA also address joint inference over lexica and world states, question-answer sequences, speaker-specific adaptation, and convention formation (Degen, 2023; Schuster & Degen, 2020). However, in a given RSA model, the state space, utterance alternatives, semantics, utility or cost function, and recursion reasoning depth are typically specified *before* the resulting production or interpretation distribution is calculated.

Here, we utilize active inference techniques to offer a candidate architecture for rendering these distinctions explicit. Under the free-energy principle (FEP), variational free energy (VFE) can be used to characterize approximate inference about the hidden causes of observations that have already occurred, whereas expected free energy (EFE) evaluates policies in terms of their predicted epistemic and preferred-outcome consequences (Friston, 2010; Millidge et al., 2021; Parr et al., 2022). Model-parameter learning and learning of a prior over policies can then be represented separately from both current-state inference and prospective policy evaluation. This domain-general formalism is not, by itself, an *explanation* of pragmatic behavior. It acquires specific explanatory power and predictive content only when the relevant hidden states, likelihoods, transitions, policy set, and update rules have been specified.

Human communication routinely succeeds despite a mismatch between what linguistic expressions conventionally encode and what speakers use them to accomplish. Interlocutors infer referents, lexical senses, discourse goals, speech-act force, social stance, and the relevance of an utterance to a developing activity. They also use response timing, repair, uptake, and non-verbal action as evidence about what has been understood. Pragmatic inference is therefore not only a hearer's recovery of a speaker's private intention – it is also an interactional process in which each move changes both the evidence available to the participants and the space of appropriate subsequent moves (Bohn & Frank, 2025; Elder & Haugh, 2024).

Several traditions isolate different parts of this problem. Gricean and neo-Gricean approaches emphasize rational and cooperative reasoning over alternatives. Relevance Theory emphasizes ostension, contextual effects, and the search for an interpretation worth the processing effort. Dynamic and discourse-semantic approaches model the evolution of common ground, commitments, discourse referents, and the question under discussion. Lastly, dialogue research treats grounding, repair, turn-taking, and partner adaptation as constitutive of communication rather than as peripheral corrections to sentence interpretation (Buz et al., 2016; Clark & Schaefer, 1989; Fedzechkina et al., 2012; Levinson, 2016; Lewis, 1979; Murray, 2014; Noveck & Reboul, 2008; Roberts, 2012; Sauerland & Schumacher, 2016). No single formalism automatically subsumes all of these traditions, but any process model of discourse should make contact with the distinctions they motivate.

The RSA framework provides the most influential probabilistic formalization of speaker-listener interdependence. It links a literal listener, a pragmatic speaker, and a pragmatic listener through Bayesian inference and softmax choice, and has been extended beyond simple reference games to lexical uncertainty, scalar and nonliteral interpretation, politeness, over-informativeness, speaker-specific adaptation, and convention formation (Bergen et al., 2016; Degen, 2023; Degen et al., 2020; Goodman & Stuhlmüller, 2013; Yoon et al., 2020). RSA is not intrinsically restricted to static or one-shot problems. Nevertheless, a particular RSA model commonly begins by fixing its state space, alternatives, semantics, utility, cost, and recursion depth, after which it characterizes a distribution over utterances or interpretations. Additional assumptions are required to distinguish current-state inference from parameter learning, to specify how an unexpected response changes the next action, and to represent habits or longer-horizon policies.

Here, we ask whether active inference supplies a useful common architecture for these additional distinctions. Variational free energy (VFE) characterizes inference about hidden causes of observations and learning of generative-model parameters; expected free energy (EFE) evaluates counterfactual policies in terms of anticipated information gain and preferred outcomes; and a prior over policies can encode context-sensitive or learned action biases (Da Costa et al., 2020; Friston, 2010; Parr et al., 2022; Sajid et al., 2021; Smith et al., 2022). These ingredients do not themselves constitute a theory of pragmatics. Explanatory content arises only after linguistically and interactionally motivated states, observations, transitions, preferences, and policies have been specified. Our proposal is therefore a constrained process architecture for pragmatic interaction, not a derivation of pragmatic theory from the free-energy principle alone.

A second, computational distinction is needed here. Exact Bayesian updating is straightforward when an experiment supplies a small set of states and utterance alternatives. Yet, in less constrained discourse possible referents, lexical interpretations, the force of speech acts, discourse goals, social stances, and partner models combine rapidly. A process-level account must therefore distinguish the exact posterior implied by a generative model from the tractable posterior $q$ that a bounded agent can represent or compute.

Previous work has used active inference to model turn-taking, interpersonal model alignment, and hierarchical linguistic exchanges (Friston & Frith, 2015b; Friston et al., 2020; Vasil et al., 2020). Murphy et al. (2024a) relate the FEP to the internal construction of syntactic objects through explicit complexity and compression measures – this concerns the generation of natural language constituency structures, but not the various actions that they make pragmatically available. The present model begins downstream of that problem: it assumes that structured utterances first generated by natural language syntax are already available (Murphy 2026), and asks how they function as evidence about hidden discourse-relevant causes.

### 1.1. The grammatical-pragmatic division of labor

The present proposal concerns communicative use rather than the internal construction of linguistic objects. It assumes that grammar and lexical-semantic knowledge have already made structured utterance candidates available. Syntax determines constituency and dependency relations; lexical and compositional semantics constrain the interpretations conventionally associated with those structures; prosody and other aspects of externalization provide additional observable cues. The active inference model begins when an interlocutor must treat one of these structured outputs as evidence about discourse-relevant causes and evaluate possible verbal or non-verbal responses.

This scope is compatible with the distinction between the design of the language faculty and the use of its outputs in ostensive-inferential communication (Murphy, 2019, 2020, 2021, 2023). Communicative pressures may shape externalization, audience design, convention formation, and the deployment of linguistic expressions without implying that Merge or the syntax-semantics interface are optimized for communication. The model developed here therefore makes no claim that free-energy minimization derives grammatical principles or explains why human syntax has its particular formal properties – this is a topic addressed in Murphy et al. (2024a). It instead targets the downstream problem of how agents use grammatically generated objects to alter one another's beliefs, commitments, and courses of action.

In what we assume below, low VFE means that an observed action is well explained under an agent's generative model. It does not mean that the proposition expressed by the utterance is *true*. A confidently held but mis-specified partner model may assign high likelihood to a false, deceptive, ironic, or systematically misunderstood utterance. Truth conditions, lexical entailments, presuppositions, and conventional speech-act constraints must therefore enter through the independently motivated structure of the likelihood and state space. Pragmatic active inference operates over these constraints rather than replacing them.

### 1.2. Relations to major traditions in pragmatics

#### 1.2.1. Gricean and probabilistic pragmatics

RSA inherits a recognizably Gricean idea: an interpreter reasons about why a rational agent selected one communicative action rather than relevant alternatives. Probabilistic pragmatics makes this reasoning graded by combining prior beliefs, a semantics, a choice rule, and a utility function. The present model retains these core assumptions, and its departure will be architectural rather than ‘anti-Gricean’: under our model, the partner’s choice distribution becomes part of the observing agent’s generative model, and the variables updated after a response are separated into current pragmatic state, persistent partner parameters, policy evaluation, and habit. This separation is intended to address current challenges concerning boundedness, scalability, incrementality, and learning that have already been emphasized within RSA itself (Degen, 2023; Franke, 2011; Franke & Jäger, 2016; Qing & Franke, 2015).

#### 1.2.2. Relevance Theory

Our model also has a natural, but limited, relation to Relevance Theory. Both approaches treat an utterance as ostensive evidence whose interpretation depends on inferred causes and contextual consequences rather than on decoding alone. A policy that is expected to resolve consequential uncertainty or produce a desirable downstream effect may have lower EFE, while policies that require additional time or exceed listener capacity may be disfavored. These parallels should not be stated as identities. Relevance-theoretic “cognitive effects” are not simply Shannon information gain, “processing effort” is not automatically the KL-complexity term in VFE, and the presumption of optimal relevance is not equivalent to minimizing an arbitrarily specified EFE. The active-inference account should instead be presented as a possible process-level implementation space in which relevance-theoretic hypotheses can be rendered explicit and compared. Recent attempts to probabilize Relevance Theory make this comparison especially timely (Unger & Buschmeier, 2025).

#### 1.2.3. Dynamic context, QUD, and commitments

Dynamic pragmatics treats discourse as an evolving record rather than a sequence of isolated intention-recovery problems. Assertions propose changes to a commitment space; questions organize possible continuations; presuppositions and discourse referents alter what later expressions can felicitously depend on; and the question under discussion constrains relevance and information structure (Farkas & Bruce, 2010; Lewis, 1979; Murray, 2014; Roberts, 2012). The corresponding variables should not be hidden inside a single private ‘intention’ state, and in the model we develop below we therefore distinguish an interaction-level discourse context – including QUD, commitments, turn or sequence position, and joint activity – from the partner's private referential, interpretive, goal, and stance variables. Participants may of course hold uncertain or divergent beliefs about the public record; 'interaction-level' does not mean infallibly shared.

#### 1.2.4. Grounding, repair, and sequence organization

Dialogue is jointly managed through evidence of understanding. A contribution is not fully grounded merely because a speaker has produced an informative form; acknowledgment, relevant next action, repair initiation, reformulation, and acceptance provide graded evidence that the contribution has entered the interactional record (Barr, 2008; Clark & Schaefer, 1989; Garrod & Pickering, 2004; Pickering & Garrod, 2004). Conversation analysis further shows that repair and closing are organized sequences with preferences and transition-relevant positions, not simply private decisions to ask another question or stop speaking (Dingemanse et al., 2015; Schegloff et al., 1977; Schegloff & Sacks, 1973). Active inference is useful here because policies can extend across several actions, but the model must exploit that capacity rather than reduce every discourse move to a one-step utterance.

#### 1.2.5. Audience design and partner adaptation

Speakers and listeners adapt to partner-specific lexical precedents, uncertainty expressions, pronunciations, expertise, and feedback. Some effects appear immediately; others persist across repeated interaction and become conventionalized within a dyad (Brennan & Clark, 1996; Brown-Schmidt, 2012; Hawkins et al., 2020; Metzing & Brennan, 2003; Schuster & Degen, 2020). This literature supplies direct empirical motivation for our distinction between a rapidly changing posterior over the partner's present state and slower learning of partner-model parameters. It also supplies a benchmark: a new formalism must predict when adaptation is partner-specific, when it generalizes to a category of partners, and when it survives a change in task or temporal pressure.

#### 1.2.6. Social meaning, indirectness, and noncooperation

Communicative preferences are not exhausted by accurate transfer of a target belief. Speakers may trade informativeness against kindness, face, plausible deniability, status, identity, or strategic risk; they may mislead, deceive, withhold, or coordinate action without seeking detailed representational alignment (Cummins, 2025; Lee & Pinker, 2010; Pinker et al., 2008; Yoon et al., 2020). These phenomena motivate a plural preference model. Listener alignment can be one valued outcome, but task success, commitment management, social consequences, and action coordination must be allowed to compete with it. This is also where the framework can move beyond a default assumption of cooperation: the generative model may attribute different preferences, reliabilities, and degrees of cooperativity to different partners.

## 1.3. Active inference, social coupling, and communication

The application of active inference to communication predates the present proposal and already contains several distinct strands. Friston and Frith (2015a, 2015b) model interpersonal exchange as reciprocally coupled inference in which agents' generative models become coordinated through action. Friston et al. (2020) develop a deep, factorized generative model for question-answer dialogue and show how hierarchical message passing can support linguistic exchange. Vasil and colleagues interpret human communication as active inference directed toward interpersonal model alignment (Vasil et al., 2020). Related work situates social expectations, cultural conventions, and shared affordances within generative models that are distributed across agents and environments (Constant et al., 2018; Constant et al., 2019; Ramstead et al., 2016; Veissière et al., 2020).

This literature also contains important internal criticism. Tison and Poirier (2021) argue that cooperative communication should not be reduced to alignment of internal models: its primary function may be to reorganize a shared field of affordances and coordinate joint action. Gallagher and Allen (2018) and Hipólito and van Es (2022) likewise emphasize embodied, enactive, and interactional alternatives to accounts centered on detached mindreading. These critiques are directly relevant to any attempt to make alignment the objective of communication. We therefore do not introduce a separate alignment functional: preferred outcomes may concern actions, commitments, social relations, environmental changes, or a task-specific listener state, as specified by the model under study.

The present discussion's intended contribution within this literature is rather specific. First, we connect probabilistic pragmatics to coupled active inference by defining the likelihood of a partner's observable action from the policy distribution attributed to that partner. Second, we distinguish current-state inference, partner-model learning, policy evaluation, and habitual priors within one sequential architecture. Third, we give an explicit restricted correspondence with RSA. Fourth, we distinguish outcome-valued components, contextual temporal policy bias, and learned habit even when they are statically equivalent within a one-step RSA score. These claims should be evaluated through formal equivalence proofs, parameter-recovery analyses, longitudinal transfer, and empirical model comparison rather than through the generality of the free-energy principle.

## 1.4. Model overview and contributions

The model we will develop here makes such a specification for multi-turn communication. An utterance has a dual status: it is an action and part of a policy for the speaker, while constituting an observation for the addressee. The hidden pragmatic state is factorized rather than reduced to a single undifferentiated 'intention'. Agent $A$ models an observed action by agent $B$ through the policy posterior that $A$ attributes to $B$. That attributed posterior depends on $A$'s estimates of $B$'s beliefs, preferences, policy precision, habitual biases, and model of $A$. Observing $B$'s action therefore initiates a recurrent sequence in which $A$ updates its beliefs about $B$, predicts $B$'s responses to its own candidate policies, evaluates those policies and selects its next action:

$$\text{observed partner action} \rightarrow \boldsymbol{q_A}\left(\boldsymbol{z_t^B}, \boldsymbol{\theta_{A\leftarrow B}}\right) \rightarrow \boldsymbol{q_A}\left(\boldsymbol{o_{t+1}^A} \mid \boldsymbol{\pi_t^A}\right) \rightarrow \boldsymbol{G_A}\left(\boldsymbol{\pi_t^A}\right) \rightarrow \boldsymbol{q_A}\left(\boldsymbol{\pi_t^A}\right) \rightarrow \text{next action.} \quad (\mathbf{1})$$

Equation (1) summarizes the recurrent inference-action loop from the perspective of agent $A$. Figure 1 depicts the schema of the model we will outline.

After observing an utterance or other action produced by agent $B$, $A$ updates $q_A(z_t^B, \theta_{A\leftarrow B})$: its beliefs about $B$'s current pragmatic state $z_t^B$, together with the parameters $\theta_{A\leftarrow B}$ of its longer-term generative model of $B$. Using this updated partner model, $A$ predicts the observations that would follow from each candidate policy $\pi_t^A$, represented by $q_A(o_{t+1}^A \mid \pi_t^A)$. These predicted outcomes determine the EFE, $G_A(\pi_t^A)$, of each policy, which evaluates both its expected information gain and its tendency to produce preferred outcomes. EFE, together with prior or habitual policy biases and any evidence supplied by the current interaction, determines the posterior $q_A(\pi_t^A)$ over policies. Selecting or sampling from this posterior generates $A$'s next verbal or non-verbal action, which in turn becomes an observation for $B$, continuing the coupled interaction. The arrows indicate computational dependencies rather than deterministic transformations or a requirement that all updates occur sequentially or at the same timescale.

Naturally, the two components of $q_A(z_t^B, \theta_{A\leftarrow B})$ may change at different rates. Beliefs about a currently intended referent may change after a single observation, whereas partner-specific likelihoods, preferences, reliability estimates and habitual priors may require evidence accumulated across multiple interactions. Conversational repair, clarification, partner adaptation, persistent misunderstanding and discourse termination can therefore arise from distinct updates within the same generative architecture.

We believe that the relationship here to RSA approaches is immediate and constructive: Under explicit restrictions (e.g., a fixed and finite state and utterance space, a fixed partner model, exact posterior inference, stable preferences and precision, and an EFE parameterization chosen to match RSA utility) the active-inference policy posterior reproduces the RSA speaker distribution $S_1$, and the corresponding Bayesian update reproduces the pragmatic-listener distribution $L_1$. This establishes a restricted embedding of the RSA calculation, even if it does not show that the FEP *uniquely* derives RSA from independent first principles.

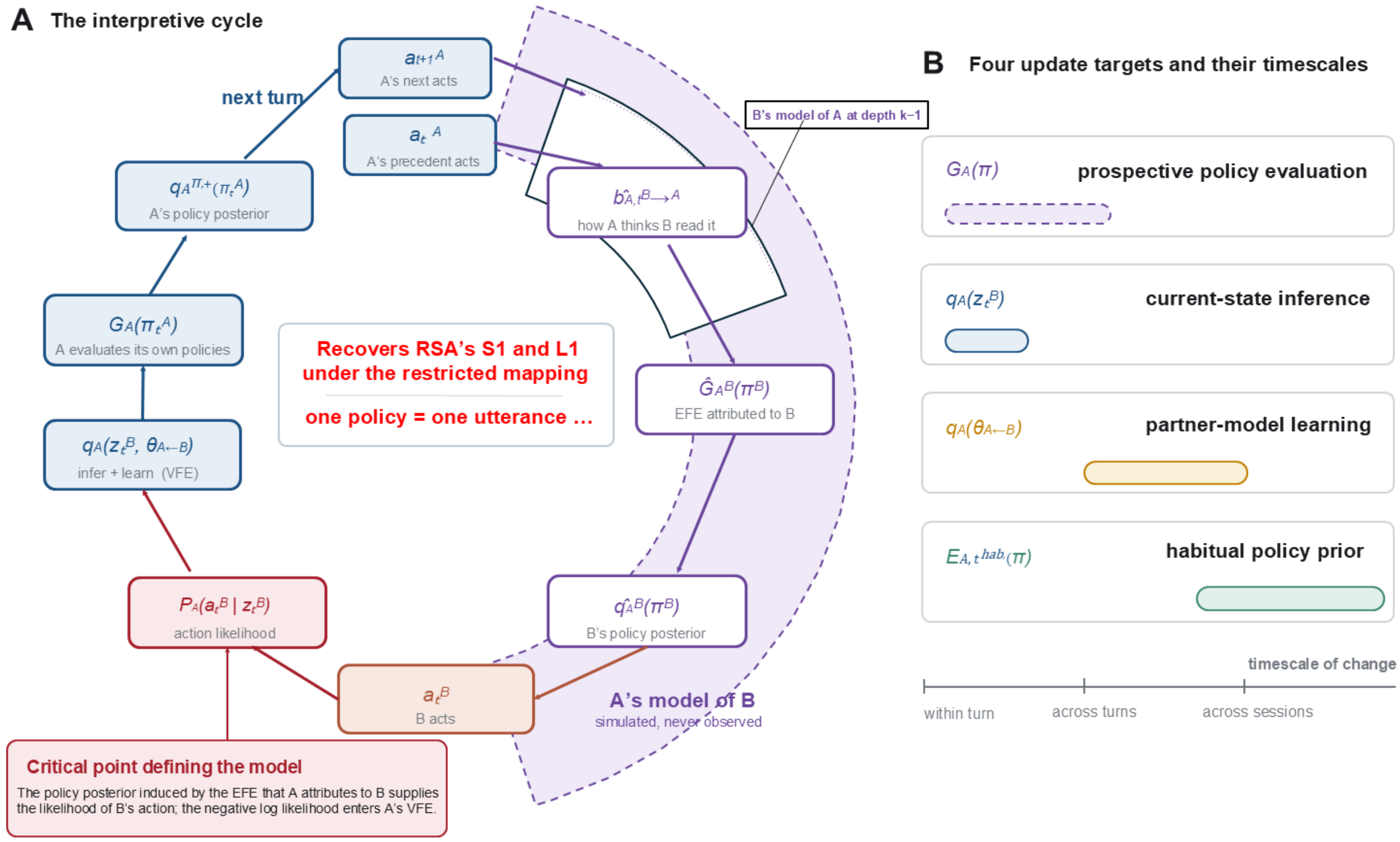

**Figure 1: Discourse as coupled active inference. (A)** The interpretive cycle from the perspective of agent A. A's preceding action $a_{t-1}^{A}$ is interpreted by B, and A represents that interpretation as the nested belief $\hat{b}_{A,t}^{B\to A}$. Conditional on this attributed reading, A represents the expected free energy $\hat{G}_A^B(\pi^B)$ that B assigns to its candidate policies and the resulting policy posterior $\hat{q}_A^B(\pi^B)$. The shaded field denotes quantities attributed to B within A's generative model and therefore not directly observed; the inner plate represents B's model of A, at which the recursion is truncated. B's overt action $a_t^B$ is observed by A. The action likelihood $P_A(a_t^B \mid z_t^B)$ is induced by the policy posterior that A attributes to B: for temporally extended policies, it is obtained by marginalizing over policies beginning with the observed action, whereas under the one-step approximation the action and policy can be identified. The negative log likelihood enters A's variational free energy and updates $q_A(z_t^B, \theta_{A\leftarrow B})$, comprising inference about B's current state and learning about the relatively persistent partner model. A then evaluates its own candidate policies through $G_A(\pi_t^A)$, forms the policy posterior $q_A^{\pi,+}(\pi_t^A)$, and selects its next action $a_{t+1}^{A}$, continuing the coupled cycle. Under the one-policy-one-utterance, exact-inference, fixed-recursion, and restricted parameter-mapping conditions stated in Section 3.12, the attributed speaker distribution and the corresponding Bayesian update recover the RSA speaker $S_1$ and pragmatic listener $L_1$. **(B)** Four model quantities positioned according to their characteristic temporal signatures. Expected free energy $G_A(\pi)$ is a prospective policy evaluation that is recomputed as beliefs and context change. Current-state inference $q_A(z_t^B)$ may resolve within a single turn, whereas learning about relatively persistent partner-model quantities $q_A(\theta_{A\leftarrow B})$ typically accumulates across turns and interactions. The habitual policy prior $E_{A,t}^{\text{hab}}(\pi)$ reflects learned policy history, may persist across sessions, and may decay over longer timescales. These positions indicate characteristic rather than mandatory temporal boundaries: the distinction between inference and learning is determined by the quantity being updated, not by a universal timescale. The context-sensitive temporal factor $\Phi_{A,t}^{\text{time}}(\pi \mid c_t)$, which combines with $E_{A,t}^{\text{hab}}(\pi)$ to form the current normalized policy prior, is treated separately in Section 4 and is not shown.

A second contribution concerns the causal allocation of terms that can be observationally equivalent in a static RSA score. We do not propose an exhaustive psychological decomposition of RSA cost. Instead, outcome-valued task, social, and effort consequences are represented through factorized prior preferences and PV; listener capacity modifies the comprehension likelihood; the duration-dependent price of reaching a communicative benefit is represented by a context-sensitive temporal policy factor that modulates policy prior support, whereas repeated policy selection changes a learned habitual policy prior; and the normalized product of these contributions defines the policy prior used for selection. Under restricted one-step conditions, the temporal policy factor can reproduce the same probabilities as an RSA cost. When a deadline is removed, a partner changes, or a policy is repeatedly practiced, however, the allocation generates different immediate and carryover predictions. This is a testable modeling commitment rather than a consequence that follows uniquely from the FEP.

Stated plainly, the proposal makes three claims against which a theory of pragmatics can be tested. First, interpretation and production are one computation: the likelihood a listener assigns to an utterance is the choice rule the listener attributes to the speaker, so every utterance is at once evidence about the speaker and an intervention on the listener's next move. Second, pragmatic knowledge is organized by update target and timescale rather than by a single score: reference, sense, and force are resolved within a turn; partner reliability, preferences, and convention familiarity are learned across turns; habits of expression are acquired across conversations; and the price of time is set by the current deadline. Third, RSA is the single-turn fixed point of this process. It is recovered exactly under the restrictions of Section 3.12, and it is precisely away from that fixed point – after unexpected feedback, a change of partner, or the removal of a deadline – that the model predicts what a static score cannot: which adaptations persist, which are partner-specific, and which transfer.

In the following section, we introduce the VFE, EFE, learning, and policy-selection quantities used in the model. We construct the coupled two-agent architecture and state the restricted correspondence with RSA. Section 4 then distinguishes static score equivalence from causal allocation, separating factorized outcome value, likelihood constraints, a contextual price of time, and learned habit. The toy examples illustrate RSA-equivalent choice, benefit per unit time, partner-induced policy reversal, stable misunderstanding, and discourse termination. Lastly, we derive empirically distinguishable predictions and clarify the scope and limitations of the proposed embedding.

## 2. Active Inference Quantities

We use a discrete partially observable Markov decision process only to distinguish the variables optimized by inference, learning, and policy selection. Let $s_t$ denote hidden states, $o_t$ observations, $a_t$ actions, $\theta$ relatively persistent model parameters, and $\pi = (a_t, \ldots, a_{t+H-1})$ a policy over a finite horizon. Conditional on a policy, a minimal generative model factorizes as

$$\boldsymbol{P(o_{1:T}, s_{1:T} \mid \pi, \theta) = P(s_1 \mid \theta) \prod_{\tau=1}^{T} P\,(o_\tau \mid s_\tau, \theta) \prod_{\tau=1}^{T-1} P\,(s_{\tau+1} \mid s_\tau, a_\tau, \theta).} \qquad (2)$$

Given observations already obtained, an agent represents a tractable posterior $q(s_t)$ and minimizes variational free energy

$$\boldsymbol{F_t[q] = \mathbb{E}_{q(s_t)}[ln q(s_t) - ln P(o_t, s_t \mid \theta)] = D_{KL}[q(s_t) \parallel P(s_t \mid o_t, \theta)] - ln P(o_t \mid \theta).} \qquad (3)$$

For a fixed model and observation, minimizing $F_t$ tightens a bound on surprisal by bringing $q$ toward the model posterior; it does not change the evidence of an observation that has already occurred. Learning instead changes θ, for example through a point-estimate update

$$\boldsymbol{\theta_{t+1} = \theta_t - \kappa_\theta \nabla_\theta F_t,} \qquad (4)$$

where $\kappa_\theta$ is a parameter-learning rate. The distinction between inference and learning is therefore defined by the variable being updated, not by a universal temporal boundary, although state beliefs will often change faster than partner-model parameters.

Policies concern future observations that have not yet been sampled. At a future time $\tau$, expected free energy is written here as

$$\boldsymbol{G_\tau(\pi) = -\mathbb{E}_{q(o_\tau, s_\tau \mid \pi)} D_{KL}[q(s_\tau \mid o_\tau, \pi) \parallel q(s_\tau \mid \pi)] - \mathbb{E}_{q(o_\tau \mid \pi)} ln P^\star(o_\tau)}$$
$$\boldsymbol{= -IG_\tau(\pi) - PV_\tau(\pi).} \qquad (5)$$

$$\boldsymbol{G(\pi) = \sum_{\tau=t+1}^{t+H} G_\tau\,(\pi).} \qquad (6)$$

The first term is expected information gain about hidden states; the second evaluates predicted outcomes under a prior-preference distribution $P^\star$. The signs follow the convention that policies with greater information gain and more preferred predicted outcomes have lower EFE. Equivalent risk-ambiguity decompositions and the full derivation are given in Appendix A. Neither decomposition determines which linguistic states or communicative outcomes should be represented.

The two occurrences of $q$ in Equation (5) have different temporal roles. In the notation introduced explicitly in Section 3.9, $q(s_\tau \mid \pi) = q^-(s_\tau \mid \pi)$ is the predictive belief before a possible future outcome, whereas $q(s_\tau \mid o_\tau, \pi) = q^+(s_\tau \mid o_\tau, \pi)$ is the counterfactual posterior that would be obtained after that outcome. When exact inference is unavailable, $q^+$ is represented by the optimized member of a variational family obtained through an inner counterfactual-VFE inversion, whose prior is the policy-conditioned predictive belief $q^-$. EFE is the separate outer prospective score that averages the epistemic and preferred-outcome consequences of these possible posterior updates.

A posterior over policies may combine a policy prior $E_t(\pi)$, policy-specific evidence already available, and prospective EFE:

$$\boldsymbol{q_t(\pi) \propto E_t(\pi) exp\left[-F_{\pi,t} - \gamma_t G_t(\pi)\right]}. \qquad (7)$$

Here $F_{\pi,t}$ scores policy-specific evidence already available, $G_t(\pi)$ scores future consequences, and $\gamma_t$ is policy precision. In the one-step examples below, $F_{\pi,t}$ is common across alternatives and cancels. A prior E and an EFE-based policy ranking must remain formally distinct: a frequently selected policy can retain high prior probability even when it is no longer prospectively optimal. In Section 4, we parameterize this prior as the normalized product of a learned habitual baseline and a context-sensitive temporal factor. The habitual component is updated from policy history, whereas the temporal factor changes with the current deadline or opportunity context.

## 3. Discourse as Coupled Active Inference

In this section, we will formalize discourse as a coupled active-inference process in which two agents observe one another's actions, infer the hidden causes of those actions, and select subsequent actions by anticipating the other agent's response. Under the active inference framework, perception updates an internal posterior so as to explain currently available observations, whereas action changes which observations will subsequently be sampled. Applied to discourse, the other agent's behavior becomes an observation to be explained, while one's own verbal or non-verbal behavior becomes part of a policy that shapes the other agent's subsequent response.

The central proposal we will develop here is not merely that discourse participants select utterances by minimizing expected free energy – this is already part and parcel of existing models of communication. More specifically, we explore how the likelihood assigned to another agent's action is itself modeled as that agent's policy posterior. This policy posterior depends on how the other agent is believed to represent the present agent, what the other agent is expected to learn from subsequent interaction, and which future outcomes the other agent is believed to prefer. The other agent's EFE is therefore not added externally as an independent objective. Rather, it enters the observer's model through the likelihood of the other agent's action and is consequently embedded within the observer's own variational and expected free-energy functionals.

### 3.1. A coupled perception-action loop between two agents

First, we begin with some elementary assumptions about the formal nature of dialogue. Let $\mathrm{A}$ and $\mathrm{B}$ denote two discourse participants. Each participant may perform both verbal and non-verbal actions. The action of agent $\mathrm{i} \in \{\mathrm{A}, \mathrm{B}\}$ at time $\mathrm{t}$ is written as

$$\boldsymbol{a_t^i} = \left(\boldsymbol{a_t^{i,ver}}, \boldsymbol{a_t^{i,nonver}}\right), \qquad (8)$$

where $a_t^{i,ver}$ denotes a verbal action, such as an utterance, and $a_t^{i,nonver}$ denotes a non-verbal action, such as closing a window, pointing toward an object, handing an object to the other agent, or changing physical position[1] [2].

An action and a policy are not strictly identical. An action is an event performed at a particular time, whereas a policy is a temporally extended sequence of actions:

$$\boldsymbol{\pi}_t^i = \left(\boldsymbol{a}_t^i, \boldsymbol{a}_{t+1}^i, \dots, \boldsymbol{a}_{t+H-1}^i\right). \qquad (9)$$

For a one-step model in which an entire conversational turn is treated as a single macro-action, however, we may use the approximation

$$\boldsymbol{\pi}_t^i \equiv \boldsymbol{a}_t^i. \qquad (10)$$

This approximation provides a direct correspondence with RSA models in which the choice of a single utterance is treated as the relevant decision. We emphasize that this is a limiting case utilized primarily for establishing a numerical correspondence with static RSA. In real-world discourse, policies retain a temporal horizon H>1, allowing agents to evaluate communicative moves based on their long-term impact on common ground and task goals.

The action of one agent becomes an observation for the other. Under the simplest direct-observation approximation,

$$o_t^B = a_t^A, \qquad o_{t+1}^A = a_{t+1}^B. \qquad (11)$$

More generally, the other agent observes the sensory consequences of an action rather than the action as an abstract variable. We may therefore write

$$\boldsymbol{o}_t^B \sim \boldsymbol{P}_B\left(\boldsymbol{o}_t^B \mid \boldsymbol{a}_t^A, \boldsymbol{e}_t\right), \qquad (\mathbf{12})$$

where $e_t$ denotes the state of the shared environment. If $A$ closes a window, for example, $B$ may observe the movement of $A$, the visual and auditory consequences of the window closing, and subsequent changes in temperature or environmental noise. Equation (11) is therefore a simplifying limit in which this sensory transformation is suppressed.

---

[1] Prosodic and non-segmental cues, including intonation, stress, voice quality, and speech rate, are treated as components of verbal action in the broad sense adopted here. Their independent likelihood structures and temporal dynamics are not developed in the present paper.

[2] While Eq. 8 treats action as a simple tuple, we assume that the generative model assigns distinct precision weights to verbal and non-verbal components. Non-verbal components can be detailed continuous variable, but we will simplify them as a discrete bulk of actions on each discrete turn, which sharing the same *t* with verbal components.

The interaction between the two agents forms a recursive perception-action cycle:

$$\boldsymbol{a}_t^A \rightarrow \boldsymbol{o}_t^B \rightarrow \boldsymbol{q}_B \rightarrow \boldsymbol{\pi}_{t+1}^B \rightarrow \boldsymbol{a}_{t+1}^B \rightarrow \boldsymbol{o}_{t+1}^A \rightarrow \boldsymbol{q}_A \rightarrow \boldsymbol{\pi}_{t+2}^A \rightarrow \cdots. \qquad (13)$$

Each participant interprets the other's action by inferring its hidden causes and evaluates the next policy conditional on the resulting posterior. This is a special case of the recursively coupled action-perception loops used in active-inference models of interpersonal communication (Friston et al., 2020; Parr et al., 2022).

### 3.2. Inferring discourse states from another agent's actions

The latent causes of an observed action should not be collapsed into a single undifferentiated speaker intention. We distinguish three levels: the structured linguistic form; the interaction-level discourse context in which the action occurs; and the partner-private states. We also distinguish parameters that may have generated action from latent spaces.

Let $y_t^B$ denote the structured form made available by grammar and externalization, including lexical material, syntactic structure, compositional content, prosody, gesture, and relevant non-verbal action features. Let the interaction-level context be

$$\boldsymbol{c}_t^{AB} = \left(\boldsymbol{Q}_t^{AB}, \boldsymbol{C}_t^{AB}, \boldsymbol{\tau}_t^{AB}, \boldsymbol{\mathcal{A}}_t^{AB}\right), \qquad (\mathbf{14})$$

where $Q_t^{AB}$ is the current question under discussion, $C_t^{AB}$ is a commitment or common-ground record, $\tau_t^{AB}$ is the turn and sequence position, and $\mathcal{A}_t^{AB}$ is the joint activity or task. These variables are interaction-level rather than intrinsically private, but an agent may remain uncertain about them or represent a different version from its partner.

The current partner-private pragmatic state is factorized as

$$\boldsymbol{x}_t^B = \left(\boldsymbol{r}_t^B, \boldsymbol{\ell}_t^B, \boldsymbol{f}_t^B, \boldsymbol{d}_t^B, \boldsymbol{k}_t^B, \boldsymbol{v}_t^B, \boldsymbol{g}_t^B\right), \qquad (\mathbf{15})$$

where $r_t^B$ is the relevant world or referential hypothesis (e.g., whether the intended referent is the red cup or the blue cup), $\ell_t^B$ a lexical or contextual interpretation (e.g., whether bank refers to a financial institution or a river bank), $f_t^B$ speech-act force (e.g., whether Can you close the window? functions as a question or a request), $d_t^B$ a discourse-level goal (e.g., whether the current discourse is organized around establishing a location or a price), $k_t^B$ common-ground increment (e.g., whether the location of the projector has been accepted as shared), $v_t^B$ affective or social stance (e.g., whether an evaluative remark is sincere or ironic), and $g_t^B$ the proximal interaction/task goal (e.g., obtaining a piece of information or inducing the partner to perform an action). The current partner-relevant latent state combines the private and interaction-level factors:

$$\boldsymbol{z}_t^B = \left(\boldsymbol{x}_t^B, \boldsymbol{c}_t^{AB}\right). \qquad (1)$$

Let $h_t^A$ denote the interaction history available to A,

$$\boldsymbol{h_t^A} = \left(\boldsymbol{o_{1:t}^A}, \boldsymbol{a_{1:t-1}^A}\right). \qquad (2)$$

The likelihood of the observable action is constrained by linguistic form, discourse state, and history:

$$\boldsymbol{P_A}\left(\boldsymbol{a_t^B} \mid \boldsymbol{z_t^B}, \boldsymbol{y_t^B}, \boldsymbol{h_{t-1}^A}\right). \qquad (3)$$

This notation does not assume that the factors are independent. It identifies a structured hypothesis space in which dependencies can be specified, learned, and tested rather than hidden inside a single intention variable.

After observing $a_t^B$, agent $A$ forms the posterior

$$\boldsymbol{q_A}\left(\boldsymbol{z_t^B} \mid \boldsymbol{a_t^B}, \boldsymbol{h_{t-1}^A}\right). \qquad (4)$$

The corresponding VFE is

$$\boldsymbol{F_{A,t}}[\boldsymbol{q_A}] = \mathbb{E}_{\boldsymbol{q_A}(\boldsymbol{z_t^B})}\left[\boldsymbol{ln q_A}\left(\boldsymbol{z_t^B}\right) - \boldsymbol{ln P_A}\left(\boldsymbol{a_t^B}, \boldsymbol{z_t^B} \mid \boldsymbol{h_{t-1}^A}\right)\right]. \qquad (5)$$

Factorizing the joint distribution into a predictive prior and a likelihood gives

$$\begin{aligned}\boldsymbol{F_{A,t}}[\boldsymbol{q_A}] = \boldsymbol{D_{KL}}&\left[\boldsymbol{q_A}\left(\boldsymbol{z_t^B}\right) \parallel \boldsymbol{P_A}\left(\boldsymbol{z_t^B} \mid \boldsymbol{h_{t-1}^A}\right)\right] \\ &-\mathbb{E}_{\boldsymbol{q_A}(\boldsymbol{z_t^B})}\boldsymbol{ln P_A}\left(\boldsymbol{a_t^B} \mid \boldsymbol{z_t^B}, \boldsymbol{h_{t-1}^A}\right).\end{aligned} \qquad (6)$$

The first term measures the departure of $A$'s updated belief from its predictive prior. The second measures how well each hypothesis about $B$ explains the observed action.

### 3.3. Exact Bayesian inference and a tractable posterior $\boldsymbol{q}$

If the latent state space is small and finite, $A$ may calculate the exact Bayesian posterior:

$$\boldsymbol{P_A}\left(\boldsymbol{z_t^B} \mid \boldsymbol{a_t^B}, \boldsymbol{h_{t-1}^A}\right) = \frac{\boldsymbol{P_A}\left(\boldsymbol{a_t^B} \mid \boldsymbol{z_t^B}, \boldsymbol{h_{t-1}^A}\right)\boldsymbol{P_A}\left(\boldsymbol{z_t^B} \mid \boldsymbol{h_{t-1}^A}\right)}{\sum_{\tilde{z}} \boldsymbol{P_A}\left(\boldsymbol{a_t^B} \mid \tilde{z}, \boldsymbol{h_{t-1}^A}\right)\boldsymbol{P_A}\left(\tilde{z} \mid \boldsymbol{h_{t-1}^A}\right)}. \qquad (22)$$

In a simple experimental reference game, the relevant world states and utterances may be supplied by the experimenter as small finite sets. In that case, every likelihood value and the complete normalization term can be evaluated directly.

The hypothesis space of natural discourse is considerably larger. Possible referents, lexical senses, speech-act forces, discourse goals, social stances, partner models, and background assumptions combine combinatorially. It is therefore generally implausible to assume that a cognitive agent explicitly enumerates and normalizes over every possible latent configuration. The agent may instead represent a tractable approximate posterior:

$$\boldsymbol{q_A}\left(\boldsymbol{z_t^B} \mid \boldsymbol{a_t^B}, \boldsymbol{h_{t-1}^A}\right) \approx \boldsymbol{P_A}\left(\boldsymbol{z_t^B} \mid \boldsymbol{a_t^B}, \boldsymbol{h_{t-1}^A}\right). \qquad \mathbf{(23)}$$

One possible approximation restricts posterior support to a contextually available hypothesis set

$$\widetilde{\mathcal{Z}}_t^A \subseteq \mathcal{Z} . \qquad (7)$$

The corresponding support-restricted posterior is

$$\boldsymbol{q_A\left(z_t^B \mid a_t^B, h_{t-1}^A\right) = \frac{1\left[z_t^B \in \widetilde{\mathcal{Z}}_t^A\right] P_A\left(a_t^B \mid z_t^B, h_{t-1}^A\right) q_A\left(z_t^B \mid h_{t-1}^A\right)}{\sum_{\tilde{z} \in \widetilde{\mathcal{Z}}_t^A} P_A\left(a_t^B \mid \tilde{z}, h_{t-1}^A\right) q_A\left(\tilde{z} \mid h_{t-1}^A\right)}}. \qquad (\mathbf{25})$$

However, $q$ should not be understood merely as a function that removes candidate states. Tractability may also be obtained through a learned recognition mapping or any other tractable approximation. The distinction between $P_A(z_t^B \mid a_t^B)$ and $q_A(z_t^B \mid a_t^B)$ is therefore substantive. The former is the posterior mathematically implied by $A$'s generative model, whereas the latter is the posterior that $A$ can actually represent or compute. Under a sufficiently expressive variational family and sufficiently accurate optimization,

$$\mathrm{q_A}\left(\mathrm{z_t^B} \mid \mathrm{a_t^B}\right) \approx \mathrm{P_A}\left(\mathrm{z_t^B} \mid \mathrm{a_t^B}\right), \qquad (26)$$

and the distinction becomes negligible. Standard finite RSA updates can be recovered under this limiting condition. In a larger or more deeply nested model, however, the distinction between the exact posterior and the agent's tractable posterior becomes essential.

### 3.4. The basic correspondence with the RSA listener

A standard pragmatic listener under RSA theories observes an utterance and infers the speaker's state:

$$\mathrm{L_1(s \mid u) \propto S_1(u \mid s)P(s)}. \qquad (27)$$

In the present notation, the utterance corresponds to $a_t^B$, and the speaker state corresponds to $z_t^B$. We may therefore identify

$$\mathrm{P_A}\left(\mathrm{a_t^B = u \mid z_t^B = s}\right) \equiv \mathrm{S_1^{A \leftarrow B}(u \mid s)}, \qquad (28)$$

where $S_1^{A\leftarrow B}$ is not direct access to $B$'s internal probability distribution. It is the speaker model of $B$ represented within $A$'s generative model. The posterior update becomes

$$q_A(z_t^B \mid a_t^B) \propto S_1^{A\leftarrow B}(a_t^B \mid z_t^B) q_A(z_t^B). \qquad (29)$$

This has the same basic Bayesian structure as the RSA pragmatic listener (Degen, 2023). However, Equation (29) does not yet explain where the speaker likelihood $S_1^{A\leftarrow B}$ comes from. Standard RSA approaches commonly define it through a softmax transformation of a utility that depends on a literal listener and an utterance cost. The present model instead derives the likelihood from the active-inference policy posterior that $A$ attributes to $B$.

### 3.5. Learning a partner-specific generative model

Within a single turn, $A$ infers the current state $z_t^B$. Across multiple turns, it may also learn more persistent regularities concerning the behavior of $B$. This corresponds to the question: Under which hidden states does this partner tend to produce which actions? Let

$$\theta_{\mathrm{A\leftarrow B}} \qquad (30)$$

denote the collection of relatively persistent quantities represented in A's partner-specific generative model of B. The components of $\theta_{\mathrm{A\leftarrow B}}$need not have the same mathematical form: they may include scalar parameters, multidimensional likelihood or transition parameters, categorical partner-model hypotheses, and learned distributions over policies. The likelihood becomes

$$P_{\theta_{A\leftarrow B}}(a_t^B \mid z_t^B, h_{t-1}). \qquad (31)$$

In a fully Bayesian formulation, $A$ maintains a joint posterior over the current state of $B$ and the parameters of the partner model:

$$\boldsymbol{q_A\left(z_t^B, \theta_{A\leftarrow B} \mid h_t\right) \propto \; P_{\theta_{A\leftarrow B}}\left(a_t^B \mid z_t^B, h_{t-1}\right) \times q_A\left(z_t^B, \theta_{A\leftarrow B} \mid h_{t-1}\right).} \qquad \mathbf{(32)}$$

The update of

$$\boldsymbol{q_A\left(z_t^B\right)} \qquad \mathbf{(33)}$$

constitutes inference about the current state of the partner, whereas the update of

$$\boldsymbol{q_A(\theta_{A\leftarrow B})} \qquad \mathbf{(34)}$$

constitutes learning about the partner-specific generative mapping. Under a point-estimate approximation, parameter learning can be written as

$$\theta_{A\leftarrow B,t+1} = \theta_{A\leftarrow B,t} - \eta_\theta \nabla_{\theta_{A\leftarrow B}} F_{A,t}. \qquad (35)$$

The learned parameters may include

$$\theta_{A\leftarrow B} = \left(\theta_{A\leftarrow B}^{\mathrm{lik}}, \theta_{A\leftarrow B}^{\mathrm{trans}}, \rho_B, \eta_B, \gamma_B, m^B, E_B^{\mathrm{h}ab}\right). \qquad (36)$$

These parameters encode, respectively,

1. partner-specific likelihood regularities governing how B's pragmatic states and interaction context are associated with observable verbal and non-verbal behavior;
2. partner-specific transition regularities governing how B's latent pragmatic states tend to persist or change across conversational turns;
3. B's reliability or the reliability of the state-to-observation mapping;
4. B's relatively persistent preferences over predicted outcomes;

5. B's policy precision, governing the sensitivity of policy probability to differences in current policy scores;
6. B's relatively persistent partner-model type or convention-related hypothesis, such as whether B is familiar with a shorthand convention; and
7. B's learned habitual policy prior $E_B^{\mathrm{h}ab}$, representing historical baseline propensities over policies independently of immediate prospective evaluation.

State inference and partner-model learning may both be expressed within the same free-energy framework, but they concern different quantities. The former updates beliefs about the current partner state, whereas the latter updates relatively persistent partner-model quantities governing states, observations, preferences, policy selection, and habitual propensity. This distinction corresponds to the general active-inference distinction between inference over hidden states and learning of generative-model and policy parameters (Parr et al., 2022).

Crucially, we do not posit two independent policy priors. The model retains a single normalized prior over policies. Within A's model of B, $E_B^{\mathrm{h}ab}$ represents the relatively persistent habitual baseline learned from B's policy history. For the temporal analyses developed in Section 4, the current policy prior attributed to B is obtained by combining this learned baseline with the positive context-sensitive temporal factor $\Phi_{B,t}^{time}$ and normalizing across policies. The temporal factor changes with immediate deadline or opportunity conditions and is therefore not itself a persistent partner parameter or a second independently learned policy prior.

The present proposal does not require the claim that multi-turn RSA models have never represented changing beliefs, lexical learning, or adaptation to a partner. Rather, it provides a common formulation in which current-state inference, partner-specific parameter learning, and policy selection can be represented as distinct updates within the same coupled generative model.

### 3.6. The speaker B's policy posterior as listener A's likelihood

Our central formal proposal is that the likelihood of an observed action is induced by the policy distribution that A attributes to B. For a temporally extended policy, A must marginalize over every policy whose first action is compatible with the observation:

$$P_A(a_t^B \mid z_t^B, h_{t-1}^A) = \sum_{\pi_t^B \in \Pi_B} \mathbf{1}\,[\pi_t^B[0] = a_t^B]\hat{q}_A^B\big(\pi_t^B \mid z_t^B, \hat{b}_{A,t}^{B\to A}, h_{t-1}^A\big) \quad (37)$$

The hat and the subscript A indicate that this is A's representation of B's policy distribution, not direct access to the distribution actually represented by B. The indicator function selects policies beginning with the observed action. $\hat{b}_{A,t}^{B\to A}$denotes the nested belief that A attributes to B regarding A's own state (formally defined in Eq. 41).

Only under a one-step macro-action approximation, in which an entire conversational turn is identified with a policy, does the expression reduce to

$$P_A(a_t^B \mid z_t^B, h_{t-1}^A) = \hat{q}_A^B(\pi_t^B = a_t^B \mid z_t^B, \hat{b}_{A,t}^{B\to A}, h_{t-1}^A). \qquad (38)$$

The one-step identity is therefore a useful limiting case for comparison with RSA, whereas policy marginalization is the general action likelihood required by the coupled model.

A temporal conditioning constraint is required to prevent circular explanation. The policy distribution used to predict $a_t^B$ must be constructed from the information available before that action is observed:

$$\hat{q}_A^B(\pi_t^B \mid z_t^B, \hat{b}_{A,t}^{B\to A}, h_{t-1}^A). \qquad (39)$$

The same observed action cannot first enter a policy-specific evidence term and then be used to define its own likelihood. After $a_t^B$ is observed, A updates its posterior over B's state and model parameters:

$$q_A(z_t^B, \theta_{A\leftarrow B,t} \mid a_t^B, h_{t-1}^A). \qquad (40)$$

This posterior may then influence A's prediction of B's later actions and A's own next policy. The distinction is temporal: pre-action beliefs generate the likelihood; the observed action updates the partner model; and the updated model reorganizes future policy evaluation.

Before selecting an action, $B$ is assumed to interpret the preceding action of $A$. The belief about $A$ that $A$ attributes to $B$ is

$$\hat{b}_{A,t}^{B\to A}(x_t^A) := \hat{q}_A^B(x_t^A \mid a_{t-1}^A, h_{t-1}). \qquad (41)$$

This distribution answers the question 'How does $A$ believe that $B$ has interpreted $A$'s preceding action?' The next policy of $B$ is therefore conditioned not directly on an objectively available state of $A$, but on the posterior about $A$ that is attributed to $B$:

$$a_{t-1}^A \to \hat{b}_{A,t}^{B\to A} \to \hat{G}_A^B(\pi_t^B) \to \hat{q}_A^B(\pi_t^B) \to a_t^B. \qquad (42)$$

A general policy posterior may include policy-specific evidence, a prior over policies, and EFE:

$$\hat{q}_A^B(\pi_t^B) = \mathrm{softmax}[\ln \hat{E}_A^B(\pi_t^B) - \hat{F}_{A,\pi,t}^B - \hat{\gamma}_A^B \hat{G}_A^B(\pi_t^B)]. \qquad (43)$$

Here, $\hat{E}_A^B$ is the policy prior that $A$ attributes to $B$, $\hat{F}_{A,\pi,t}^B$ is the policy-specific variational free energy that $A$ attributes to $B$, and $\hat{\gamma}_A^B$ is the attributed policy precision.

If currently available observations provide equal evidence for the prospective policy alternatives, or if the policy-specific variational free-energy term is omitted under a one-step approximation, then

$$\hat{q}_A^B\left(\pi_t^B \mid z_t^B, \hat{b}_{A,t}^{B\to A}\right) = \frac{\hat{E}_A^B(\pi_t^B \mid z_t^B)\exp[-\hat{\gamma}_A^B \hat{G}_A^B\left(\pi_t^B \mid z_t^B, \hat{b}_{A,t}^{B\to A}\right)]}{\hat{Z}_A^B\left(z_t^B, \hat{b}_{A,t}^{B\to A}\right)}, \tag{44}$$

where

$$\begin{aligned}\hat{Z}_A^B\left(z_t^B, \hat{b}_{A,t}^{B\to A}\right) = \sum_{\tilde{\pi}^B} \hat{E}_A^B\,(\tilde{\pi}^B \mid z_t^B) \\ \times \exp[-\hat{\gamma}_A^B \hat{G}_A^B\left(\tilde{\pi}^B \mid z_t^B, \hat{b}_{A,t}^{B\to A}\right)]\end{aligned} \tag{45}$$

is the corresponding normalization term.

### 3.7. The listener A's variational free energy contains the speaker B's expected free energy

By applying the one-step approximation identity from (38) to the variational free energy in (21), we obtain

$$\begin{aligned}F_{A,t}[q_A] = D_{\mathrm{KL}}[q_A(z_t^B) \parallel P_A(z_t^B \mid h_{t-1})] \\ -\mathbb{E}_{q_A(z_t^B)} \ln \hat{q}_A^B\left(a_t^B \mid z_t^B, \hat{b}_{A,t}^{B\to A}\right).\end{aligned} \tag{46}$$

Taking the logarithm of Equation (44) gives

$$\begin{aligned}\ln \hat{q}_A^B\left(a_t^B \mid z_t^B, \hat{b}_{A,t}^{B\to A}\right) = \ln \hat{E}_A^B(a_t^B \mid z_t^B) \\ -\hat{\gamma}_A^B \hat{G}_A^B\left(a_t^B \mid z_t^B, \hat{b}_{A,t}^{B\to A}\right) \\ -\ln \hat{Z}_A^B\left(z_t^B, \hat{b}_{A,t}^{B\to A}\right).\end{aligned} \tag{47}$$

Substitution into Equation (46) yields

$$\begin{aligned}F_{A,t}[q_A] = D_{\mathrm{KL}}[q_A(z_t^B) \parallel P_A(z_t^B \mid h_{t-1})] \\ -\mathbb{E}_{q_A(z_t^B)} \ln \hat{E}_A^B(a_t^B \mid z_t^B) \\ +\mathbb{E}_{q_A(z_t^B)}[\hat{\gamma}_A^B \hat{G}_A^B\left(a_t^B \mid z_t^B, \hat{b}_{A,t}^{B\to A}\right)] \\ +\mathbb{E}_{q_A(z_t^B)} \ln \hat{Z}_A^B\left(z_t^B, \hat{b}_{A,t}^{B\to A}\right).\end{aligned} \tag{48}$$

This equation shows that the EFE attributed to $B$ appears inside the variational free energy used by $A$ to interpret the observed action.

Let the EFE attributed to $B$ be decomposed as

$$\hat{G}_A^B\left(a_t^B \mid z_t^B, \hat{b}_{A,t}^{B\to A}\right) = -\widehat{\mathrm{IG}}_A^B\left(a_t^B \mid z_t^B, \hat{b}_{A,t}^{B\to A}\right) - \widehat{PV}_A^B\left(a_t^B \mid z_t^B, \hat{b}_{A,t}^{B\to A}\right). \tag{49}$$

Equation (48) then becomes

$$
\begin{aligned}
F_{A,t}[q_A] = {} & \underbrace{D_{\mathrm{KL}}[q_A(z_t^B) \parallel P_A(z_t^B \mid h_{t-1})]}_{A\text{'s belief-update complexity}} \\
& - \underbrace{\mathbb{E}_{q_A(z_t^B)} \ln \hat{E}_A^B(a_t^B \mid z_t^B)}_{\text{attributed prior support for } B\text{'s action}} \\
& + \underbrace{\mathbb{E}_{q_A(z_t^B)} \ln \hat{Z}_A^B\left(z_t^B, \hat{b}_{A,t}^{B\to A}\right)}_{\text{normalization over } B\text{'s policies}} \\
& - \underbrace{\mathbb{E}_{q_A(z_t^B)}[\hat{\gamma}_A^B \widehat{\mathrm{IG}}_A^B\left(a_t^B \mid z_t^B, \hat{b}_{A,t}^{B\to A}\right)]}_{\text{epistemic value attributed to } B} \\
& - \underbrace{\mathbb{E}_{q_A(z_t^B)}[\hat{\gamma}_A^B \widehat{PV}_A^B\left(a_t^B \mid z_t^B, \hat{b}_{A,t}^{B\to A}\right)]}_{\text{preferred consequences attributed to } B}.
\end{aligned}
\qquad (50)
$$

Under the one-step assumptions above, this is the first central result of our model. It does not state that the two agents jointly minimize the simple sum $F_A + G_B$. Agent $A$ does not directly minimize the EFE of $B$. Rather, the EFE attributed to $B$ determines the likelihood of $B$'s action, and the negative log likelihood of that action appears in $A$'s VFE. Also, what is important is that the normalization term is generally state-dependent and itself contains the scores of all alternative policies. It is therefore not a removable constant: it makes the likelihood of the observed action depend on its EFE relative to the alternatives attributed to B.

What is not trivial is what the likelihood thereby depends on. Through $b_{A,t}^{B\to A}$ it depends on what A takes B to believe about A; through the attributed information gain it depends on what B is expected to learn about A from A's reply; and through $Z^{A\leftarrow B}$ it depends on the value of the observed action relative to the alternatives B is taken to have had. An RSA listener conditions on a speaker model in which the first of these is fixed by the recursion depth and the second is absent. In the coupled model all three are quantities A can be wrong about and can revise, which is what allows the same observed utterance to be interpreted differently by A on different turns and with different partners.

### 3.8. Expanding the attributed information gain and pragmatic value

Suppose that, after performing $a_t^B$, agent $B$ will observe a future response from $A$:

$$
o_{t+1}^B = a_{t+1}^A. \qquad (51)
$$

The information gain that $A$ attributes to $B$ is

$$
\begin{aligned}
\widehat{\mathrm{IG}}_A^B\left(a_t^B \mid z_t^B, \hat{b}_{A,t}^{B\to A}\right) = {} & \mathbb{E}_{\hat{q}_A^B(o_{t+1}^B \mid a_t^B, z_t^B, \hat{b}_{A,t}^{B\to A})} \\
& D_{\mathrm{KL}}[\hat{q}_A^B\left(x_{t+1}^A \mid o_{t+1}^B, a_t^B, z_t^B, \hat{b}_{A,t}^{B\to A}\right) \\
& \parallel \hat{q}_A^B\left(x_{t+1}^A \mid a_t^B, z_t^B, \hat{b}_{A,t}^{B\to A}\right)].
\end{aligned}
\qquad (52)
$$

The distribution on the right side of the KL divergence is the predictive belief that $A$ attributes to $B$ before $B$ receives the future response. The distribution on the left is the posterior that $A$ expects $B$ to form after observing that response.

The predictive belief is generated from the belief about $A$ currently attributed to $B$:

$$\hat{q}_A^B\left(x_{t+1}^A \mid a_t^B, z_t^B, \hat{b}_{A,t}^{B\to A}\right) = \sum_{x_t^A} \hat{P}_A^B\left(x_{t+1}^A \mid x_t^A, a_t^B, z_t^B\right) \times \hat{b}_{A,t}^{B\to A}(x_t^A). \tag{53}$$

Thus, the epistemic value of $B$'s action ultimately depends on how $B$ is believed to have interpreted the preceding action of $A$. The pragmatic value attributed to $B$ is

$$\widehat{PV}_A^B\left(a_t^B \mid z_t^B, \hat{b}_{A,t}^{B\to A}\right) = \mathbb{E}_{\hat{q}_A^B\left(o_{t+1}^B \mid a_t^B, z_t^B, \hat{b}_{A,t}^{B\to A}\right)} \ln \hat{P}_A^{\star B}\left(o_{t+1}^B \mid g_t^B, \eta_B\right). \tag{54}$$

Here, $g_t^B$ denotes the proximal interaction or task goal that A attributes to B at the current turn, and $\eta_B$ denotes A's current estimate of B's relatively persistent preference parameters. $\hat{P}_A^{\star B}$ is the prior-preference distribution that $A$ attributes to $B$; it may assign high probability to outcomes in which $A$ answers a question, agrees with a proposal, performs a desired action, resolves an ambiguity, or responds in a socially acceptable manner.

Substituting Equations (52) and (54) into Equation (50) gives

$$\begin{aligned} F_{A,t}[q_A] = {} & D_{\mathrm{KL}}\left[q_A(z_t^B) \parallel P_A(z_t^B \mid h_{t-1})\right] \\ & -\mathbb{E}_{q_A(z_t^B)} \ln \hat{E}_A^B\left(a_t^B \mid z_t^B\right) + \mathbb{E}_{q_A(z_t^B)} \ln \hat{Z}_A^B\left(z_t^B, \hat{b}_{A,t}^{B\to A}\right) \\ & -\mathbb{E}_{q_A(z_t^B)\hat{q}_A^B\left(o_{t+1}^B \mid a_t^B, z_t^B, \hat{b}_{A,t}^{B\to A}\right)} \hat{\gamma}_A^B \\ & \times D_{\mathrm{KL}}\left[\hat{q}_A^B\left(x_{t+1}^A \mid o_{t+1}^B, a_t^B, z_t^B, \hat{b}_{A,t}^{B\to A}\right) \parallel \hat{q}_A^B\left(x_{t+1}^A \mid a_t^B, z_t^B, \hat{b}_{A,t}^{B\to A}\right)\right] \\ & -\mathbb{E}_{q_A(z_t^B)\hat{q}_A^B\left(o_{t+1}^B \mid a_t^B, z_t^B, \hat{b}_{A,t}^{B\to A}\right)} \hat{\gamma}_A^B \\ & \times \ln \hat{P}_A^{\star B}\left(o_{t+1}^B \mid g_t^B, \eta_B\right). \end{aligned} \tag{55}$$

Equation (55) expands the action-likelihood contribution to A's VFE in terms of the policy evaluation that A attributes to B. The VFE used by $A$ contains contributions from

1. divergence between $A$'s posterior and predictive prior over the state of $B$;
2. policy-prior support that $A$ attributes to the observed action of $B$;
3. normalization over $B$'s policies;
4. information that $B$ is expected to obtain about $A$ through the subsequent response;
5. future outcomes that $B$ is believed to prefer.

Agent $A$ therefore interprets the observed action according to the following hypothesis: $B$ interpreted my preceding action in a particular way, formed a corresponding belief and goal, and selected the present action because it was relatively probable under the policy posterior induced by those beliefs, preferences, policy priors, and expected free energy. Suppressing the additional conditioning variables, the key dependency path is

$$\boxed{a_{t-1}^{A} \to \hat{b}_{A,t}^{B\to A} \to \hat{G}_{A}^{B}(a_t^B) \to \hat{q}_{A}^{B}(a_t^B) \to P_A(a_t^B \mid z_t^B) \to q_A(z_t^B \mid a_t^B)}. \qquad (56)$$

This is the formal sense in which the other agent's EFE is embedded within the present agent's perceptual inference.

### 3.9. Separating A's predictive and counterfactual posterior beliefs

The preceding derivation concerned A's interpretation of an action of B that had already been observed. We now turn to the prospective calculation in which A evaluates its own candidate policy $\pi_t^A$. The notation must distinguish beliefs formed before and after a possible future observation. We therefore use a local minus/plus convention:

- $q_{A,\tau}^{-}(z_\tau^B \mid \pi_t^A, h_t)$ is A's policy-conditioned predictive belief about B before A receives B's future response;
- $q_{A,\tau}^{+}(z_\tau^B \mid o_\tau^A, \pi_t^A, h_t)$ is the posterior that A would form after a particular possible response $o_\tau^A$;
- $q_A^{\pi,+}(\pi^A)$ is reserved for a posterior over policies, so that a state posterior and a policy posterior are not denoted by the same symbol.

Here, $h_t$ denotes the interaction history available to A immediately before selecting. The superscripts do not indicate positive and negative value. They indicate whether the relevant observation has been incorporated. Starting from A's posterior after the history actually observed through $t$, a candidate policy induces the predictive state belief

$$q_{A,\tau}^{-}(z_\tau^B \mid \pi_t^A, h_t) = \sum_{z_t^B} P_A\left(z_\tau^B \mid z_t^B, \pi_t^A, h_t\right) q_{A,t}(z_t^B \mid h_t), \qquad \tau > t. \qquad (57)$$

Passing this belief through A's likelihood gives the predictive outcome distribution

$$q_{A,\tau}^{-}(o_\tau^A \mid \pi_t^A, h_t) = \sum_{z_\tau^B} P_A\left(o_\tau^A \mid z_\tau^B, \pi_t^A, h_t\right) q_{A,\tau}^{-}(z_\tau^B \mid \pi_t^A, h_t),$$

$$o_\tau^A = a_\tau^B \text{ in the direct-observation limit.} \qquad (58\text{a})$$

$$\tilde{P}_A(o_\tau^A, z_\tau^B \mid \pi_t^A, h_t) := P_A(o_\tau^A \mid z_\tau^B, \pi_t^A, h_t) q_{A,\tau}^{-}(z_\tau^B \mid \pi_t^A, h_t) \qquad (58b)$$

For the inner counterfactual inversion, A's policy-conditioned predictive belief $q_{A,\tau}^{-}\left(z_\tau^B \mid \pi_t^A, h_t\right)$ serves as the prior over the future state. We therefore define the auxiliary prospective density in Equation (58b) by combining this predictive belief with A's likelihood. Its outcome marginal is exactly the predictive distribution in Equation (58a).

For every possible future outcome, A can then construct the posterior it would hold if that outcome were observed. In a finite model this is an exactly normalized Bayesian update; with a restricted variational family it is the optimized approximate posterior:

$$q_{A,\tau}^{+}(z_\tau^B \mid o_\tau^A, \pi_t^A, h_t) := \arg\min_{q \in \mathcal{Q}_A} F_{A,\tau}^{cf}[q; o_\tau^A, \pi_t^A, h_t]. \qquad (59a)$$

$$q_{A,\tau}^{+}(z_\tau^B \mid o_\tau^A, \pi_t^A, h_t) \approx \tilde{P}_A(z_\tau^B \mid o_\tau^A, \pi_t^A, h_t) = \frac{P_A(o_\tau^A \mid z_\tau^B, \pi_t^A, h_t) q_{A,\tau}^{-}(z_\tau^B \mid \pi_t^A, h_t)}{q_{A,\tau}^{-}(o_\tau^A \mid \pi_t^A, h_t)}. \qquad (59b)$$

Where $F_{A,\tau}^{cf}$ is the inner counterfactual VFE whose minimizer is the posterior under each hypothetical future outcome. The approximation sign becomes equality when the exact posterior lies in $\mathcal{Q}_A$ and the minimum is attained. The counterfactual VFE minimized in this inner update is

$$F_{A,\tau}^{cf}[q; o_\tau^A, \pi_t^A, h_t] = \mathbb{E}_{q(z_\tau^B)}\left[\ln q(z_\tau^B) - \ln \tilde{P}_{A,\tau}(o_\tau^A, z_\tau^B \mid \pi_t^A, h_t)\right]. \qquad (60a)$$

$$F_{A,\tau}^{cf}[q; o_\tau^A, \pi_t^A, h_t] = D_{KL}\left[q(z_\tau^B) \parallel \tilde{P}_{A,\tau}(z_\tau^B \mid o_\tau^A, \pi_t^A, h_t)\right] - \ln q_{A,\tau}^{-}(o_\tau^A \mid \pi_t^A, h_t). \qquad (60b)$$

Unlike the retrospective VFE in Equation (20), the counterfactual VFE in Equation (60) uses A's policy-conditioned predictive belief $q_{A,\tau}^{-}$ as its prior over the future state. For each possible future outcome $o_\tau^A$, minimizing $F_{A,\tau}^{cf}$ maps this predictive belief to the corresponding counterfactual posterior $q_{A,\tau}^{+}(z_\tau^B \mid o_\tau^A, \pi_t^A, h_t)$. When exact inference is available, this minimizer coincides with the normalized Bayesian update in Equation (59b); under a restricted variational family, it is its variational projection. A's expected information gain then quantifies the belief change induced by each possible outcome by comparing this counterfactual posterior with the predictive belief from which it was obtained:

$$IG_{A,\tau}(\pi_t^A) = \mathbb{E}_{q_{A,\tau}^{-}(o_\tau^A \mid \pi_t^A, h_t)} D_{KL}\left[q_{A,\tau}^{+}(z_\tau^B \mid o_\tau^A, \pi_t^A, h_t) \parallel q_{A,\tau}^{-}(z_\tau^B \mid \pi_t^A, h_t)\right]. \qquad (61)$$

The pragmatic value of the same policy is evaluated from A's perspective, using A's predictive outcome distribution and A's prior preferences:

$$PV_{A,\tau}(\pi_t^A) = \mathbb{E}_{q_{A,\tau}^{-}(o_\tau^A \mid \pi_t^A, h_t)} \ln P_A^{\star}(o_\tau^A \mid g_t^A, \eta_A). \qquad (62)$$

The value in Equation (62) need not be a single, undifferentiated utility. Let A's predicted future outcome comprise K dimensions:

$$o_\tau^A = \left(o_\tau^{A,(1)}, \dots, o_\tau^{A,(K)}\right). \qquad (62a)$$

Depending on the task, these dimensions may include task completion, the partner's comprehension or subsequent action, effort-related consequences, and social consequences. If A's prior-preference distribution factorizes as

$$P_A^\star(o_\tau^A) = \frac{1}{Z_A}\prod_{k=1}^{K} P_{A,k}^\star\left(o_\tau^{A,(k)}\right)^{w_{A,k}}, \qquad (62b)$$

then the logarithm converts the product into a sum and A's pragmatic value decomposes additively:

$$PV_{A,\tau}(\pi_t^A) = \sum_{k=1}^{K} w_{A,k}\, PV_{A,k,\tau}(\pi_t^A) - \ln Z_A,$$
$$PV_{A,k,\tau}(\pi_t^A) = \mathbb{E}_{q_{A,\tau}^-\left(o_\tau^{A,(k)} \mid \pi_t^A, h_t\right)} \ln P_{A,k}^\star\left(o_\tau^{A,(k)}\right). \qquad (62c)$$

The normalization term is common across policy alternatives and disappears under softmax normalization. This representation is formally similar to adding several value terms to an RSA utility, but it does not require the outcome dimensions to be psychologically independent. Where interactions among outcome dimensions are theoretically important, the joint preference distribution $P_A^\star(o_\tau^A)$ can be retained instead.

The decomposition in Equations (62a)-(62c) contains values assigned to predicted consequences of a policy. It does not include the residual temporal bias introduced in Section 4: the proposed model represents that residual policy-level difference through a separate context-sensitive temporal factor $\Phi_t^{time}(\pi)$ , which multiplicatively modulates the learned habitual baseline to form the normalized policy prior $E_t$.

Consequently,

$$G_{A,\tau}(\pi_t^A) = -IG_{A,\tau}(\pi_t^A) - PV_{A,\tau}(\pi_t^A).$$

$$G_{A,\tau}(\pi_t^A) = -\mathbb{E}_{q_{A,\tau}^-(o_\tau^A \mid \pi_t^A, h_t)} D_{KL}\left[q_{A,\tau}^+(z_\tau^B \mid o_\tau^A, \pi_t^A, h_t) \parallel q_{A,\tau}^-(z_\tau^B \mid \pi_t^A, h_t)\right]$$
$$- \mathbb{E}_{q_{A,\tau}^-(o_\tau^A \mid \pi_t^A, h_t)} \ln P_A^\star(o_\tau^A \mid g_t^A, \eta_A). \qquad (63)$$

This clarifies the relation between VFE and EFE. The optimized $q_A^+$ inside Equation (63) is obtained by a counterfactual VFE minimization, but the standard EFE is not the naive average of the ordinary VFE that A would realize after observing the future outcome:

$$G_{A,\tau}(\pi_t^A) \neq \mathbb{E}_{q_{A,\tau}^-(o_\tau^A \mid \pi_t^A, h_t)} F_{A,\tau}^{cf}\left[q_{A,\tau}^+; o_\tau^A, \pi_t^A, h_t\right] \quad \text{in general.} \qquad (64)$$

Under exact inner inference, the divergence term in Equation (60b) vanishes, so the counterfactual VFE for a possible outcome reduces to its predictive surprisal, $-\ln q_{A,\tau}^-(o_\tau^A \mid \pi_t^A, h_t)$.

Averaging this quantity over predicted outcomes yields the entropy of A's predictive outcome distribution, not expected free energy. Equation (63), by contrast, assigns value to expected belief change with the opposite sign and evaluates predicted outcomes under the prior-preference distribution $P_A^{\star}$. There are therefore two nested operations. The inner counterfactual VFE minimization constructs $q^{+}$ for each possible future outcome using $q^{-}$ as its prior. The outer EFE calculation then averages the epistemic and preference consequences of those possible posterior updates in order to rank policies. EFE should therefore not be identified either with the retrospective VFE in Equation (20) or with the expectation of the counterfactual VFE in Equation (60).

### 3.10. Generating B's response recursively inside A's EFE

Equation (58) remains incomplete until the response likelihood $P_A(o_\tau^A \mid z_\tau^B, \pi_t^A, h_t)$ has been specified. In the coupled model, that likelihood is generated by A's model of how B reads A's policy and then chooses a response. First, A's action becomes an observation for B. From A's perspective, B's reading is represented as a posterior obtained from an attributed predictive belief and attributed likelihood:

$$o_t^B \sim \hat{P}_A^B(o_t^B \mid a_t^A = \pi_t^A[0], e_t). \qquad (65a)$$

$$\hat{q}_{A,t}^{B,+}(x_t^A \mid o_t^B) := \arg\min_{q \in \hat{\mathcal{Q}}_A^B} \hat{F}_{A,t}^B[q; o_t^B]. \qquad (65b)$$

$$\hat{q}_{A,t}^{B,+}(x_t^A \mid o_t^B) \approx \frac{\hat{P}_A^B(o_t^B \mid x_t^A)\hat{q}_{A,t}^{B,-}(x_t^A)}{\sum_{\tilde{x}^A} \hat{P}_A^B(o_t^B \mid \tilde{x}^A)\hat{q}_{A,t}^{B,-}(\tilde{x}^A)}. \qquad (65c)$$

Thus $\hat{q}_A^{B,+}$ is not B's objectively available mental state. It is A's model of the posterior that B would form after observing A. It corresponds to $\hat{b}_{A,t}^{B\to A}$ in Equation (41). Conditional on this reading, every candidate response policy of B generates a predictive belief about A's subsequent state:

$$\hat{q}_{A,\tau+1}^{B,-}(x_{\tau+1}^A \mid \pi_\tau^B, z_\tau^B, o_t^B) = \sum_{x_\tau^A} \hat{P}_A^B(x_{\tau+1}^A \mid x_\tau^A, \pi_\tau^B, z_\tau^B)\hat{q}_{A,\tau}^{B,+}(x_\tau^A \mid o_t^B), \qquad (66)$$

After B acts, B anticipates a further response from A, $o_{\tau+1}^B = a_{\tau+1}^A$. Passing the predictive state belief in Equation (66) through the attributed likelihood, and applying the attributed analogue of the counterfactual inversion in Equations (58a)–(60b), gives $\hat{q}_{A,\tau+1}^{B,+}(x_{\tau+1}^A \mid o_{\tau+1}^B, \pi_\tau^B, z_\tau^B, o_t^B)$ and $\hat{q}_{A,\tau+1}^{B,-}(x_{\tau+1}^A \mid \pi_\tau^B, z_\tau^B, o_t^B)$. The information gain $\widehat{IG}_{A,\tau}^B$ A attributes to B is therefore

$$\widehat{IG}_{A,\tau}^B(\pi_\tau^B) = \mathbb{E}_{\hat{q}_{A,\tau+1}^{B,-}(o_{\tau+1}^B \mid \pi_\tau^B, z_\tau^B, o_t^B)} D_{KL}\left[\hat{q}_{A,\tau+1}^{B,+}(x_{\tau+1}^A \mid o_{\tau+1}^B, \pi_\tau^B, z_\tau^B, o_t^B) \parallel \hat{q}_{A,\tau+1}^{B,-}(x_{\tau+1}^A \mid \pi_\tau^B, z_\tau^B, o_t^B)\right]. \qquad (67)$$

The corresponding pragmatic value and EFE attributed to B are

$$\widehat{PV}^B_{A,\tau}(\pi^B_\tau) = \mathbb{E}_{\hat{q}^{B,-}_{A,\tau+1}(o^B_{\tau+1}\mid\pi^B_\tau,z^B_\tau,o^B_t)}\ln\hat{P}^{\star B}_A(o^B_{\tau+1}\mid g^B_\tau,\eta_B),$$
$$\hat{G}^B_{A,\tau}(\pi^B_\tau) = -\widehat{IG}^B_{A,\tau}(\pi^B_\tau) - \widehat{PV}^B_{A,\tau}(\pi^B_\tau). \qquad (68)$$

The same decomposition can be represented inside A's model of B. Let

$$\hat{o}^B_{\tau+1} = \left(\hat{o}^{B,(1)}_{\tau+1},\dots,\hat{o}^{B,(K)}_{\tau+1}\right), \qquad (68a)$$

and suppose that the prior preference A attributes to B factorizes as

$$\hat{P}^{\star B}_A(\hat{o}^B_{\tau+1}) = \frac{1}{\hat{Z}^B_A}\prod_{k=1}^{K}\hat{P}^{\star B}_{A,k}\left(\hat{o}^{B,(k)}_{\tau+1}\right)^{\hat{w}^B_{A,k}}. \qquad (68b)$$

Then the pragmatic value attributed to B can be written as

$$\widehat{PV}^B_{A,\tau}(\pi^B_\tau) = \sum_{k=1}^{K}\hat{w}^B_{A,k}\,\widehat{PV}^B_{A,k,\tau}(\pi^B_\tau) - \ln\hat{Z}^B_A. \qquad (68c)$$

This is not direct access to B's actual preferences. It is the factorized outcome value that A represents within its attributed model of B. A's own $PV_A$ and the attributed $\widehat{PV}^B_A$ may therefore have the same formal structure while retaining different predictive distributions, preference parameters, and weights.

Here the hatted IG and PV belong to B as represented within A, whereas the unhatted quantities in Equations (61) and (62) belong to A. With policy-specific evidence, policy priors, and prospective EFE kept formally separate, the posterior over B's policies that A attributes to B is

$$\hat{q}^{B,\pi,+}_{A,t}(\pi^B_\tau \mid o^B_t, z^B_\tau) = \frac{\hat{E}^B_A(\pi^B_\tau \mid z^B_\tau)\exp\left[-\hat{F}^B_{A,\pi,t} + \hat{\gamma}^B_A\left\{\widehat{IG}^B_{A,\tau}(\pi^B_\tau) + \widehat{PV}^B_{A,\tau}(\pi^B_\tau)\right\}\right]}{\hat{Z}^B_A(o^B_t, z^B_\tau)}. \qquad (69)$$

For illustration, suppose that A approximates B's response selection by a point decision. Conditional on a particular possible observation $o^B_t$ of A's action, A then assigns B the maximum-a-posteriori (MAP) response policy:

$$\pi^{B,*}_\tau = \underset{\pi^B_\tau}{\operatorname{argmax}}\,\hat{q}^{B,\pi,+}_{A,t}(\pi^B_\tau \mid o^B_t, z^B_\tau), \qquad a^{B,*}_\tau = \pi^{B,*}_\tau[0], \qquad o^A_\tau \sim P_A\left(o^A_\tau \mid a^{B,*}_\tau\right). \qquad (70)$$

Equation (70) is a MAP point-decision approximation to the full attributed policy posterior. It is also recovered in the high-policy-precision limit when that posterior concentrates on a unique maximizing policy. The approximation concerns policy selection only; the sensory consequences generated by the selected action may remain stochastic. The general model does not require either a unique observation by B or a point-valued policy choice. A therefore marginalizes over both the possible observations that B may receive from A's action and B's full attributed policy posterior conditional on each possible observation:

$$P_A(o_\tau^A \mid z_\tau^B, \pi_t^A, h_t) = \sum_{o_t^B} \hat{P}_A^B\,(o_t^B \mid a_t^A = \pi_t^A[0], e_t) \sum_{\pi_\tau^B} P_A\,(o_\tau^A \mid \pi_\tau^B[0])\hat{q}_{A,t}^{B,\pi,+}(\pi_\tau^B \mid o_t^B, z_\tau^B). \quad (71)$$

Equation (71) therefore retains two distinct sources of uncertainty represented within A's model: uncertainty about how B observes A's current action, and uncertainty about which response policy B selects given that observation. A deterministic observation channel is recovered when A assigns unit probability to one possible observation by B. The point-policy case is recovered when each conditional policy posterior is replaced by the degenerate distribution concentrated on the MAP policy defined in Equation (70).

Substituting Equation (69) into Equation (71), and then substituting the result into Equation (58), gives the full response distribution used by A:

$$\mathcal{N}_A^B(\pi_\tau^B, z_\tau^B; o_t^B) := \hat{E}_A^B(\pi_\tau^B \mid z_\tau^B)\exp\left[-\hat{F}_{A,\pi,t}^B + \hat{\gamma}_A^B\left\{\widehat{IG}_{A,\tau}^B(\pi_\tau^B) + \widehat{PV}_{A,\tau}^B(\pi_\tau^B)\right\}\right]. \quad (72a)$$

$$\mathcal{D}_A^B(z_\tau^B; o_t^B) := \sum_{\tilde{\pi}^B} \hat{E}_A^B\,(\tilde{\pi}^B \mid z_\tau^B)\exp\left[-\hat{F}_{A,\tilde{\pi},t}^B + \hat{\gamma}_A^B\left\{\widehat{IG}_{A,\tau}^B(\tilde{\pi}^B) + \widehat{PV}_{A,\tau}^B(\tilde{\pi}^B)\right\}\right]. \quad (72b)$$

$$q_{A,\tau}^-(o_\tau^A \mid \pi_t^A, h_t) = \sum_{z_\tau^B} q_{A,\tau}^-(z_\tau^B \mid \pi_t^A, h_t) \sum_{o_t^B} \hat{P}_A^B(\,o_t^B \mid a_t^A = \pi_t^A[0], e_t\,) \sum_{\pi_\tau^B} P_A\,(o_\tau^A \mid \pi_\tau^B[0]) \frac{\mathcal{N}_A^B(\pi_\tau^B, z_\tau^B; o_t^B)}{\mathcal{D}_A^B(z_\tau^B; o_t^B)}. \quad (72c)$$

Equations (65a)-(72c) show the entire recursive path. A's policy becomes B's observation; that observation updates the posterior about A that A attributes to B; this attributed posterior conditions B's predictive beliefs, IG, and PV; those quantities determine the policy posterior attributed to B; B's selected action becomes A's possible future observation. Finally, inserting the fully expanded distribution in Equation (72c) into A's own EFE gives

$$\begin{aligned} G_{A,\tau}(\pi_t^A) = -&\sum_{o_\tau^A} q_{A,\tau}^-\,(o_\tau^A \mid \pi_t^A, h_t) \sum_{z_\tau^B} q_{A,\tau}^+\,(z_\tau^B \mid o_\tau^A, \pi_t^A, h_t) \ln \frac{q_{A,\tau}^+(z_\tau^B \mid o_\tau^A, \pi_t^A, h_t)}{q_{A,\tau}^-(z_\tau^B \mid \pi_t^A, h_t)} \\ -&\sum_{o_\tau^A} q_{A,\tau}^-\,(o_\tau^A \mid \pi_t^A, h_t) \ln P_A^\star(o_\tau^A \mid g_t^A, \eta_A). \quad (73) \end{aligned}$$

Equation (73) is A's prospective counterpart to the VFE expansion in Equation (55). Its first contribution is the negative of A's expected information gain, comparing the counterfactual posterior $q_{A,\tau}^+$ with the policy-conditioned predictive belief $q_{A,\tau}^-$ from which that posterior is constructed. Its second contribution is the negative pragmatic-value term, which evaluates predicted responses under A's own prior-preference distribution.

In both contributions, the weighting distribution $q_{A,\tau}^-(o_\tau^A \mid \pi_t^A, h_t)$ is the fully marginalized response distribution constructed in Equation (72c). It integrates over A's uncertainty about B's future state, the possible observation that B receives from A's current action, and the response policy that A attributes to B conditional on that observation. B's attributed information gain, pragmatic value, current-evidence term, policy prior, and policy precision therefore affect A's expected free energy indirectly, by shaping B's attributed policy posterior and hence the distribution of responses that A expects to observe.

The levels remain nested rather than blended. A's information gain is not the information gain that A attributes to B, and A's pragmatic value is not the pragmatic value that A attributes to B. Likewise, the counterfactual posterior $q^{+}_{A,\tau}$in Equation (73) is A's own posterior over B's future state, constructed by the inner counterfactual inversion in Equations (59)-(60). The B-side quantities enter this calculation only through the response-generating distribution they induce. Thus, A evaluates its own candidate policies under its own expected free energy, while the consequences of those policies depend recursively on how A expects B to observe, interpret, evaluate, and respond to A's action.

### 3.11 Recursive interdependence without an alignment objective

The expansion above supplies the interactional connection required by the model without positing a separate alignment functional. Let $\mathcal{R}_B[\pi^A]$ denote the fully marginalized response distribution generated by B's attributed inference and policy selection in Equations (65a)-(72c), and let $\mathcal{R}_A[\pi^B]$ denote the symmetric construction from B's perspective. Then

$$\begin{aligned} G_A(\pi^A) \quad &= \mathcal{G}_A(q_A^-, q_A^+, P_A^\star; \mathcal{R}_B[\pi^A]), \\ G_B(\pi^B) \quad &= \mathcal{G}_B(q_B^-, q_B^+, P_B^\star; \mathcal{R}_A[\pi^B]), \\ q_A^{\pi,+} \to a^A \to o^B \to q_B^+ \quad &\to q_B^{\pi,+} \to a^B \to o^A \to q_A^+ \to \cdots . \end{aligned} \qquad (74)$$

Here, $\mathcal{G}_i$ denotes the EFE functional defined in Equation (63), evaluated using the response distribution induced by the other agent. The arrows in the final line denote generative or inferential dependencies rather than deterministic mappings.

Equation (74) is not the assertion that the agents minimize an additive quantity $G_A + G_B$, nor that their internal distributions must become identical. Each agent evaluates and selects policies according to its own EFE, while the future observations entering that EFE are generated by the other agent's posterior-guided actions. A change in A's policy therefore changes the distribution over B's possible observations, thereby changing B's posterior and subsequent response distribution; that response, in turn, changes A's posterior and later policy evaluation. Under compatible models and preferences, this coupling can be mutually stabilizing: an action that reduces A's EFE can elicit a response that also makes B's subsequent inference or policy evaluation easier. Under conflicting preferences or mis-specified partner models, the same coupling can instead stabilize disagreement or produce strategic divergence. Interdependence is therefore a dependency between generative processes and response likelihoods, not an assumed global alignment objective.

The goal and preference variables introduced earlier remain available inside $z^i$ and $P_i^{\star}$, but no additional section-specific alignment score is required. Any valued consequence - successful reference, task completion, avoidance of an unwanted result, or another outcome included by a substantive model - enters the relevant agent's PV through its own prior-preference distribution. This keeps the general architecture neutral about which outcomes a particular communicative task values.

### 3.12. A restricted correspondence with RSA

The recursive active-inference architecture developed above recovers the familiar RSA speaker and pragmatic-listener calculations only under explicit restrictions. The purpose of this section is not to claim that active inference uniquely derives RSA, but to show how the coupled construction developed in Sections 3.1–3.10 reduces, in a one-step exact-inference limit, to the standard RSA distributions.

The relevant comparison should begin with a general RSA utility rather than with utterance cost alone. Let

$$U_{\mathrm{RSA}}(u;s,c) = \lambda_I \ln L_0(s \mid u,c) - \lambda_C C_{\mathrm{RSA}}(u,c) + \sum_{k=1}^{K} \lambda_k V_k(u;s,c)$$

$$S_1(u \mid s,c) = \frac{\exp\{\alpha U_{\mathrm{RSA}}(u;s,c)\}}{\sum_{\tilde{u}} \exp\{\alpha U_{\mathrm{RSA}}(\tilde{u};s,c)\}} \qquad (75)$$

Here, the first term is the conventional RSA informativeness term, the second is utterance cost, and the remaining terms permit additional utility components licensed by the RSA model under comparison. No claim is made that these components must be social, psychologically additive, or uniquely implemented by a single active-inference quantity.

Now impose the following restrictions: (i) one utterance is one policy, $\pi^B = u$; (ii) the relevant state and utterance spaces are finite; (iii) the variational posteriors used in the comparison coincide with their exactly normalized Bayesian posteriors, and the corresponding priors are identified with those used in the RSA calculation, in particular $q_A^-(s \mid c) = P_{\mathrm{RSA}}(s \mid c)$; (iv) the partner model and recursion depth are fixed; (v) no generative-model parameter learning occurs within the one-step calculation; (vi) preferences and policy precision are fixed across the utterance alternatives being compared; and (vii) currently available policy evidence and the learned habitual baseline are uniform, or otherwise common, across the utterances being compared; a nonuniform contextual temporal factor may remain only when it is explicitly mapped to the temporal component of the RSA utility.

Applying these restrictions to the attributed B-side policy posterior in Equation (69) removes the temporal extension of the policy and freezes the partner-model and learning variables within the one-step calculation. The nested beliefs, transition model, likelihood, and preference parameters do not disappear. Once the partner model and recursion depth are fixed, they determine reduced policy-dependent quantities such as $\widehat{IG}_A^B(u;s,c)$ and $\widehat{PV}_A^B(u;s,c)$. Terms that are common across utterance alternatives contribute only an additive constant. Writing the common policy-specific evidence term as $F_0(s,c)$ and the fixed policy precision as γ, the resulting active-inference speaker score is

$$\mathcal{S}_{\text{AI}}(u;s,c) = \ln\hat{E}_A^B(u \mid s,c) - F_0(s,c) + \gamma\left[\widehat{IG}_A^B(u;s,c) + \widehat{PV}_A^B(u;s,c)\right] \quad (76a)$$

The corresponding attributed speaker distribution is therefore

$$\hat{q}_A^{B,\pi,+}(u \mid s,c) = \frac{\exp\{\mathcal{S}_{\text{AI}}(u;s,c)\}}{\sum_{\tilde{u}} \exp\{\mathcal{S}_{\text{AI}}(\tilde{u};s,c)\}} \quad (76b)$$

To establish the RSA correspondence, we now choose a restricted parameterization of the remaining policy-dependent terms. This is a parameter mapping for the RSA comparison, not a claim that the following identities hold for active inference in general.

Model the listener's response as its literal posterior over the target, and let the speaker's prior preference over that response be log-proportional to the probability the response assigns to the intended state. The reduced pragmatic value then carries informativeness together with any further outcome values, the policy prior carries utterance cost (through the contextual temporal factor of Section 4), and the reduced information gain – which concerns what B learns about A, not what A learns about the world – is common across the alternatives:

$$\gamma\, \text{PV}^{A\leftarrow B}(u;s,c) = \alpha\left(\lambda_I \ln L_0(s \mid u,c) + \sum_{k=1}^{K} \lambda_k\, V_k(u;s,c)\right) + K_V(s,c),$$

$$\ln E^{A\leftarrow B}(u \mid s,c) = -\alpha\lambda_C\, C_{\text{RSA}}(u,c) + K_E(c),$$

$$\gamma\, \text{IG}^{A\leftarrow B}(u;s,c) = K_I(s,c), \qquad (76c)$$

where $K_V(s,c)$, $K_E(c)$ and $K_I(s,c)$ are independent of the utterance alternative. Define $K(s,c) = K_I(s,c) + K_V(s,c) + K_E(c) - F_0(s,c)$. Substituting these relations into Equation (76a) gives

$$\mathcal{S}_{\text{AI}}(u;s,c) = \alpha\left[\lambda_I \ln L_0(s \mid u,c) - \lambda_C C_{\text{RSA}}(u,c) + \sum_{k=1}^{K} \lambda_k\, V_k(u;s,c)\right] + K(s,c)$$

$$\mathcal{S}_{\text{AI}}(u;s,c) = \alpha U_{\text{RSA}}(u;s,c) + K(s,c) \qquad (76d)$$

Because K(s,c) is common across utterance alternatives, it cancels under softmax normalization. Hence

$$\hat{q}_A^{B,\pi,+}(u \mid s,c) = \frac{\exp\{\alpha U_{\mathrm{RSA}}(u;s,c) + K(s,c)\}}{\sum_{\tilde{u}} \exp\{\alpha U_{\mathrm{RSA}}(\tilde{u};s,c) + K(s,c)\}}$$

$$\hat{q}_A^{B,\pi,+}(u \mid s,c) = \frac{\exp\{\alpha U_{\mathrm{RSA}}(u;s,c)\}}{\sum_{\tilde{u}} \exp\{\alpha U_{\mathrm{RSA}}(\tilde{u};s,c)\}} \qquad (76e)$$

$$\hat{q}_A^{B,\pi,+}(u \mid s,c) = S_1(u \mid s,c)$$

The second step uses the central likelihood construction from Section 3.6. Under the one-step identity, the policy posterior that A attributes to B is the likelihood that A assigns to B's observed utterance, so that $P_A(u \mid s,c) = \hat{q}_A^{B,\pi,+}(u \mid s,c) = S_1(u \mid s,c)$. Exact Bayesian updating by A therefore gives

$$q_A^{+}(s \mid u,c) = \frac{P_A(u \mid s,c)\, q_A^{-}(s \mid c)}{\sum_{\tilde{s}} P_A(u \mid \tilde{s},c)\, q_A^{-}(\tilde{s} \mid c)}$$

$$q_A^{+}(s \mid u,c) = \frac{S_1(u \mid s,c)\, P_{\mathrm{RSA}}(s \mid c)}{\sum_{\tilde{s}} S_1(u \mid \tilde{s},c)\, P_{\mathrm{RSA}}(\tilde{s} \mid c)} \qquad (76f)$$

$$q_A^{+}(s \mid u,c) = L_1(s \mid u,c)$$

Equations (76a)–(76f) therefore make the reduction explicit. Equation (69) first yields a one-step active-inference speaker score after the temporal, learning, and recursion variables have been fixed; a restricted parameter mapping then makes the policy-dependent part of that score equal to the RSA utility up to an utterance-independent constant; softmax normalization recovers the RSA speaker; and the policy-based likelihood identity from Section 3.6 converts that speaker distribution into the RSA pragmatic listener by ordinary Bayes' rule. The nested quantities introduced earlier are thus not discarded. Under the stated restrictions, they are either absorbed into the reduced pragmatic-value term and the policy prior or contribute only quantities common across the alternatives. The allocation also makes explicit what the coupled model adds to RSA. Informativeness is not an epistemic quantity for the speaker; it is a preference over what the listener will believe or do. The attributed information gain has no RSA counterpart: it is the value to the speaker of learning about the listener from the listener's reply, and it ceases to be constant across alternatives exactly when the speaker is uncertain about the partner (Section 5.3). This correspondence is restricted and static. It establishes equality of the normalized one-step speaker and listener distributions, but it does not identify a unique causal location for each reduced-form RSA term within the active-inference architecture.

### 3.13. Linguistic interpretation of the architecture

The value of the coupled architecture we have discussed is not that familiar pragmatic phenomena receive new names. It is that established linguistic distinctions are assigned to variables with different update targets, transition structures, and temporal signatures. Referential hypotheses, lexical interpretations, speech-act force, QUD, commitments, stance, and partner reliability are not interchangeable sources of uncertainty. A clarification about reference should update a different part of the model from evidence that a partner is sarcastic, unreliable, unfamiliar with a convention, or pursuing a different discourse goal.

1. **Reference and lexical disambiguation.** *Component:* current-state posterior over $r_t$ or $\ell_t$. *Prediction:* when represented as transient current-state factors, these beliefs can update rapidly within a turn and need not persist once the relevant context changes.
2. **Indirect speech acts.** *Component:* posterior over force $f_t$ plus action and social preferences. *Prediction:* interpretation and response-policy value vary with joint activity and social outcome preferences, even when literal informativeness is held fixed.
3. **QUD-sensitive relevance.** *Component:* interaction-level $Q_t^{AB}$ and the policy set. *Prediction:* the same utterance changes value when the active question changes.
4. **Assertion, acceptance, and commitment.** *Component:* the transition model for $C_t^{AB}$. *Prediction:* uptake changes the public record even when private belief remains uncertain.
5. **Irony and stance.** *Component:* posterior over $v_t$ and the partner-specific likelihood. *Prediction:* repeated partner-specific evidence can produce persistent expectations about stance through learning of the partner-specific likelihood.
6. **Clarification.** *Component:* EFE information gain about a selected latent factor. *Prediction:* questions target the factor whose expected resolution most changes future action.
7. **Repair.** *Component:* predicted mismatch plus a multi-action repair policy. *Prediction:* repair depends on the type and location of mismatch, not uncertainty in general.
8. **Partner adaptation.** *Component:* learning of $\boldsymbol{\theta}_{A \leftarrow B}$. *Prediction:* effects persist across turns and may remain partner-specific.
9. **Convention and shorthand.** *Component:* partner-specific learning together with a learned policy bias, developed formally in Section 4. *Prediction:* repeatedly successful forms can acquire persistent prior support; whether this yields shorter expressions depends on the available policy set and the temporal assumptions introduced in Section 4.
10. **Politeness and indirectness.** *Component:* social outcome preferences. *Prediction:* choice changes with face or relationship stakes even at fixed duration and informativeness.
11. **Deception and strategic ambiguity.** *Component:* divergent speaker and listener outcome preferences. *Prediction:* low EFE need not imply truth or belief alignment.

This mapping also clarifies what would count as evidence against the proposal. If manipulations designed to isolate outcome value, listener capacity, immediate temporal pressure, and policy history produce indistinguishable immediate and carryover effects, the proposed causal allocation gains no empirical advantage over a single flexible RSA score. If trial-by-trial responses are adequately predicted by a static RSA model and no parameter-learning, temporal-context, or policy-history term improves out-of-sample prediction, the coupled machinery is unnecessary for that task. Conversely, selective persistence after deadline removal, partner specificity, capacity-by-density interactions, factor-targeted clarification, and policy reversal after unexpected feedback would support the richer architecture.

## 4. Static Equivalence and the Causal Separation of Temporal Policy Bias

The purpose of this section is not to partition RSA cost into an exhaustive list of psychologically primitive sub-costs. In RSA, length, production difficulty, delay, listener burden, social risk, and learned convention can all enter a reduced-form policy score. This flexibility is not a defect. It becomes insufficient only when the scientific question concerns which variable changes under an intervention, which effect persists after the intervention is removed, and which effect transfers across partners or tasks.

The present proposal therefore separates two claims. First, several active-inference parameterizations can generate the same one-step choice distribution. Second, those parameterizations need not generate the same learning and transfer dynamics. We assign values of predicted consequences to factorized $PV$, expected uncertainty resolution to $IG$, constraints on outcome generation to the relevant likelihood, learned historical policy propensity to $E^{hab}$ and residual context-dependent temporal bias to $\Phi^{time}$. The normalized product of the latter two defines the policy prior E. This is a causal modeling commitment, not a claim that the FEP makes this allocation mathematically unique.

A general RSA speaker can be written in a form that keeps informativeness, additional values, and the specifically temporal component separate:

$$S_1(u \mid s,c) \propto \exp\left\{\alpha\left[\ln L_0(s \mid u,c) + \sum_{k=1}^{K} V_k\,(u;s,c) - C_{time}(u,c)\right]\right\}. \qquad (77)$$

Some influences conventionally entered as negative costs can be re-expressed as positive or negative outcome values $V_k$ on the same log-utility scale. The central active-inference score developed below has the corresponding form

$$\mathcal{S}_{A,t}(\pi) = -F_{A,t}(\pi) + \gamma_{A,t} IG_{A,t}(\pi) + \gamma_{A,t}\sum_{k=1}^{K} w_{A,k}\, PV_{A,k,t}(\pi) + \ln E_{A,t}^{hab}(\pi)$$
$$+ \ln\Phi_{A,t}^{time}(\pi \mid c_t). \qquad (78)$$

The distinction among these terms is not identified by their additive appearance in Equation (78). It is identified by their causal roles and update rules: PV is defined over predicted outcomes, $E^{hab}$ is a learned distribution over policies, and $\Phi^{time}$ is a context-dependent positive factor that modulates policy support without itself being learned from policy history.

### 4.1 Static reparameterization of E and PV

Using $G = -IG - PV$, a policy posterior can be written as

$$q_A(\pi \mid c) \propto \exp[\ln E_A(\pi \mid c) - F_A(\pi) + \gamma_A IG_A(\pi) + \gamma_A PV_A(\pi)]. \tag{79}$$

For any policy-dependent function $\delta(\pi)$, define a new normalized policy prior

$$E_A{}'(\pi \mid c) = \frac{E_A(\pi \mid c)\exp[\delta(\pi)]}{\sum_{\tilde{\pi}} E_A\,(\tilde{\pi} \mid c)\exp[\delta(\tilde{\pi})]}. \tag{80}$$

Define at the same time

$$PV_A'^{(\pi)} = PV_A(\pi) - \frac{\delta(\pi)}{\gamma_A}.\ (where\ \gamma_A > 0) \tag{81}$$

The transformed score differs from the original score only by the normalization constant in Equation (80), which is common across policies. The two parameterizations therefore produce the same policy posterior. A single cross-sectional choice distribution cannot by itself determine whether a policy-dependent bias came from $E$ or $PV$.

This static equivalence does not make the two terms generatively identical. $E(\pi \mid c)$ is a prior distribution over policies in a context. $PV(\pi)$ is an expected log preference over outcomes predicted under a policy. The former changes when a context-dependent policy propensity or learned policy history changes; the latter changes when prior preferences or the policy-to-outcome prediction changes. Their empirical distinction must therefore come from longitudinal data, transfer, and interventions rather than from fit to a single softmax distribution.

Nothing in this static reparameterization identifies the internal sources of E. In Sections 4.4–4.5, E is represented as a single normalized prior formed from $E^{hab}\Phi^{time}$. At a fixed context, only their product affects the current policy distribution; their separation is therefore identified by contextual intervention and learning history rather than by a single cross-sectional choice distribution.

### 4.2. Outcome-valued factors represented through PV

As introduced in Equations (62a)-(62c), factorized prior preferences yield an additive pragmatic value:

$$PV_A(\pi) = \sum_{k=1}^{K} w_{A,k}\, PV_{A,k}(\pi) - \ln Z_A. \tag{82}$$

The components can represent task completion, an appropriate partner response, effort or fatigue as a predicted consequence, social acceptance, or another outcome specified by the substantive model. A pressure that appeared as an RSA cost can therefore be represented as a low preference for its predicted consequences, without positing a separate psychological cost accumulator. Additivity follows only under the stated factorization of prior preferences; where outcome dimensions interact, the joint preference distribution should be retained.

Not every non-temporal influence belongs in $PV$. Listener capacity and channel noise alter the likelihood that a policy will generate successful comprehension. Expected resolution of uncertainty belongs to $IG$. Production difficulty belongs in $PV$ when the model values its effort-related consequences, but it may instead alter action likelihood or policy duration. These possibilities should be treated as rival causal models, not collapsed merely because they can produce similar choices.

Elapsed time could also be introduced as an outcome dimension and represented in $PV$. We do not claim that this is mathematically impossible. The narrower modeling decision made here is to reserve $PV$ for what terminal communicative consequences are achieved and to assign the residual policy bias concerning how quickly matched consequences are reached by the contextual temporal factor $\Phi^{time}$.

### 4.3. Terminal equivalence and the residual temporal difference

Let $\pi_S$ and $\pi_F$ be short and full policies. Suppose that $\pi_S$ and $\pi_F$ induce the same terminal predictive joint distribution and the same terminal counterfactual update mapping, and consider an EFE evaluation restricted to the terminal communicative outcome at $\tau$, then

$$q_A(s_\tau, o_\tau \mid \pi_S) = q_A(s_\tau, o_\tau \mid \pi_F). \qquad (83)$$

Under this strong terminal-equivalence condition, their terminal expected information gains are equal,

$$IG_A(\pi_S) = IG_A(\pi_F), \qquad (84)$$

their terminal pragmatic values are equal,

$$PV_A(\pi_S) = PV_A(\pi_F), \qquad (85)$$

and consequently

$$G_A(\pi_S) = G_A(\pi_F). \qquad (86)$$

The policies can nevertheless differ in duration:

$$T(\pi_S) < T(\pi_F). \qquad (87)$$

A terminal-outcome evaluation of $G$ does not distinguish these policies. A finite agent may nevertheless favor the policy that reaches the matched result earlier, because time allocated to one explanation cannot simultaneously be allocated to another utterance, topic, or non-linguistic action.

This is not a claim that temporally extended EFE can never encode time. Time-indexed preferences, discounting, or a finite horizon can distinguish policies before their terminal distributions coincide. Our restricted claim is instead that, once terminal communicative value is held fixed, a residual duration-sensitive policy bias can be represented through the contextual temporal factor $\Phi^{time}$, which modulates the learned habitual policy prior before normalization. The separation makes explicit whether an effect changes immediately with a deadline or persists because a policy has been learned.

### 4.4. Separating communicative benefit from the price of time

For a designated target $s^{\star}$, the pointwise target-evidence gain of a policy is

$$\Delta J(\pi; s^{\star}) = \ln q_{A\to B}(s^{\star} \mid e_{\pi}) - \ln q_{A\to B}(s^{\star}). \qquad (88)$$

This is not the expected mutual information $IG$. Its duration-normalized version should therefore be described as target-specific evidence gain per unit time:

$$R_J(\pi; s^{\star}) = \frac{\Delta J(\pi; s^{\star})}{T(\pi)}. \qquad (89)$$

A more general quantity begins with the prospective benefit of continuing with $\pi$ rather than stopping:

$$B_t(\pi) = G_t(\pi_{stop}) - G_t(\pi). \qquad (90)$$

The expected-free-energy benefit per unit time is

$$R_t^G(\pi) = \frac{B_t(\pi)}{T(\pi)}. \qquad (91)$$

A rate objective selects

$$\pi^{\star} \in \underset{\pi}{\operatorname{argmax}} \frac{B_t(\pi)}{T(\pi)}. \qquad (92)$$

For a finite policy set, define the optimal rate

$$\mu_t^{\star} = \max_{\pi} \frac{B_t(\pi)}{T(\pi)}. \qquad (93)$$

The Dinkelbach residual at this optimum satisfies

$$\max_{\pi}[B_t(\pi) - \mu_t^{\star} T(\pi)] = 0. \qquad (94)$$

If $B_t(\pi) = G_t(\pi_{stop}) - G_t(\pi)$, the same optimum can be written as

$$\pi^{\star} \in \underset{\pi}{\operatorname{argmin}}[G_t(\pi) + \mu_t^{\star} T(\pi)]. \qquad (95)$$

This is a normative equivalence at the optimum. It does not imply that a cognitive agent explicitly and exactly computes $\mu_t^{\star}$. A process model may instead use a context-sensitive estimate $\mu_t(c_t)$ shaped by deadlines, remaining horizon, and opportunity cost. We represent that estimated price of time as a temporal policy factor:

$$\Phi_t^{time}(\pi \mid c_t) = \exp[-\gamma_t \mu_t(c_t) T(\pi)]. \tag{96}$$

The total policy prior can then be factorized as

$$E_t(\pi \mid c_t) = \frac{E_t^{hab}(\pi)\Phi_t^{time}(\pi \mid c_t)}{\sum_{\tilde{\pi}} E_t^{hab}(\tilde{\pi})\Phi_t^{time}(\tilde{\pi} \mid c_t)}. \tag{97}$$

Equation (97) defines one normalized policy prior, not two. $E_t^{hab}$ supplies the learned baseline propensity, whereas $\Phi_t^{time}$ transiently tilts that baseline according to the current opportunity structure.

The policy posterior becomes

$$q_t(\pi \mid c_t) \propto E_t^{hab}(\pi)\Phi_t^{time}(\pi \mid c_t)\exp[-F_t(\pi) - \gamma_t G_t(\pi)]. \tag{98}$$

Informational and pragmatic benefit remains in $G_t$; the temporal factor changes the policy-level price of reaching those consequences, whereas $E_t^{hab}(\pi)$ carries historical propensity. For the terminal-equivalent policies in Equations (83)-(87), then,

$$\frac{q_t(\pi_S \mid c_t)}{q_t(\pi_F \mid c_t)} = \frac{E_t^{hab}(\pi_S)}{E_t^{hab}(\pi_F)}\exp\{\gamma_t \mu_t(c_t)[T(\pi_F) - T(\pi_S)]\}. \tag{99}$$

If their habitual priors are equal and $\mu_t > 0$, the shorter policy is preferred. When terminal benefits differ, $G$ and $\Phi_t^{time}(\pi \mid c_t)$ jointly determine the trade-off between benefit and duration.

In a restricted one-step RSA comparison, the temporal factor can be mapped to a conventional temporal cost by setting

$$\ln \Phi^{time}(u \mid c) = -\alpha C_{time}(u, c) + \mathrm{K}(\mathrm{c}), \tag{100}$$

where $\mathrm{K}(\mathrm{c})$ is common across utterance alternatives. Thus, in the restricted one-step comparison, the contextual temporal factor can be numerically identical to an RSA temporal cost. This equality does not identify the causal source of the bias; the interpretations diverge when temporal context or policy history is manipulated.

### 4.5. Contextual temporal policy factors and learned habitual priors

The two influences entering Equation (97) have different causal sources and update rules. The contextual temporal factor $\Phi_t^{time}(\pi|c_t)$ is determined by the current temporal context and may change immediately when a deadline or opportunity cost changes. By contrast, $E_t^{hab}(\pi)$ is a learned habitual policy prior that reflects accumulated policy selections and may persist after the temporal pressure has disappeared. Their normalized product determines the current policy prior $E_t(\pi \mid c_t)$. Because only this combined prior affects a single contemporaneous choice distribution, the two contributions are not identifiable from cross-sectional choice alone. Their distinction becomes empirically meaningful under intervention and learning: changes in temporal context directly alter $\Phi_t^{time}$, whereas repeated realized policy selections gradually alter $E_t^{hab}$.

To represent this history dependence, let $e_t(\pi) \geq 0$. denote the policy-specific habitual concentration available at turn t. A minimal update with forgetting is

$$e_{t+1}(\pi) = (1-\omega)e_t(\pi) + \kappa_E 1\{\pi_t = \pi\}, \qquad (101)$$

where $0 \leq \omega \leq 1$ controls forgetting, $\kappa_E \geq 0$ controls the strength of habit acquisition, and $\mathbf{1}[\pi_t = \pi]$ equals one for the policy actually selected at turn t and zero otherwise. The corresponding smoothed habitual policy prior is

$$E_{t+1}^{hab}(\pi) = \frac{e_{t+1}(\pi) + \alpha_0}{\sum_{\tilde{\pi}} [e_{t+1}(\tilde{\pi}) + \alpha_0]}, \qquad \alpha_0 \geq 0. \qquad (102)$$

This update does not compute temporal efficiency a second time. At turn $t$, policy selection is already determined jointly by the habitual baseline, the contextual temporal factor, current policy-specific evidence, and prospective expected free energy as in Equation (98).

The realized policy selected from this distribution is then accumulated in the habitual component. The relevant dependency can therefore be summarized as

$$\boxed{c_t \to \mu_t(c_t) \to \Phi_t^{time} \to q_t(\pi) \to \pi_t \to E_{t+1}^{hab}}. \qquad (103)$$

The temporal factor does not itself become a habit. Rather, it transiently changes the distribution from which policies are selected. If the same temporal context repeatedly biases selection toward a particular policy, the resulting realized selections can gradually increase that policy's habitual prior probability. Thus, removing a one-off deadline can immediately return $\Phi_t^{time}$ to unity, whereas a habitual bias acquired through repeated selections may persist and decay only gradually through $\omega$.

### 4.6. Listener capacity and over-compression

A high communicative benefit per unit time is not automatically successful. Compressing adequate evidence into too short an interval may exceed the listener's momentary processing capacity. Let $\ell_\pi(t)$ be the local processing load induced by policy $\pi$ and $\kappa_B$ the capacity A attributes to B. Cumulative overload can be written as

$$\mathcal{O}(\pi, \kappa_B) = \int_0^{T(\pi)} \max\{\ell_\pi(t) - \kappa_B, 0\} dt. \qquad (104)$$

A simple comprehension likelihood is

$$P(correct \mid \pi, \kappa_B, \nu) = \sigma[\beta_0 - \beta_1 \mathcal{O}(\pi, \kappa_B) - \beta_2 \nu], \qquad (105)$$

where $\nu$ denotes channel noise or a second processing burden. Capacity is not an additional utterance cost and is not itself a $PV$ component. It changes the policy-to-outcome likelihood and therefore changes the predicted distributions under which $IG$ and $PV$ are evaluated.

As a simplified hard-constraint approximation to the capacity-sensitive likelihood in Equation (105), a target-directed compression objective can be expressed as

$$\pi^\star = \underset{\pi}{\operatorname{argmax}} \frac{\Delta J(\pi; s^\star)}{T(\pi)} \quad \text{subject to} \quad \lambda(\pi) \leq \kappa_B, \qquad (106)$$

where $\lambda(\pi)$ summarizes the relevant local information-density or processing-load constraint.

A more general active-inference expression is

$$\pi^\star = \underset{\pi}{\operatorname{argmax}} \frac{G_t(\pi_{stop}; , \kappa_B, \nu) - G_t(\pi; , \kappa_B, \nu)}{T(\pi)} \qquad (107)$$

This avoids three extremes: a policy can be too short to resolve the relevant target, too dense for the listener to process, or too long relative to its marginal benefit. Communicative efficiency is therefore partner-relative rather than equivalent to minimizing utterance length.

### 4.7. Evidence arrival and the speed of variational updating

The claim that two terminal-equivalent policies can differ in convergence time requires an additional process theory. Let $\phi$ parameterize an approximate posterior and suppose that each effective evidence unit $n$ induces

$$\phi_{n+1} = \phi_n - \eta \nabla_\phi F_n. \qquad (108)$$

If policy $\pi$ supplies effective evidence at rate $r_\pi$, then

$$\frac{dn}{dt} = r_\pi. \qquad (109)$$

Under a continuous approximation,

$$\frac{d\phi}{dt} = \frac{d\phi}{dn}\frac{dn}{dt} \approx -\eta r_\pi \nabla_\phi F. \qquad (110)$$

If two policies eventually supply the same sufficient statistics, they may reach the same posterior fixed point,

$$q_{\infty}^{(\pi_S)}(s) = q_{\infty}^{(\pi_F)}(s). \qquad (111)$$

But, for the case in which the short policy delivers the relevant effective evidence at a higher rate, $r_{\pi_S} > r_{\pi_F}$ , then,

$$T_{conv}(\pi_S) < T_{conv}(\pi_F). \qquad (112)$$

The trajectory can therefore be the same when indexed by effective evidence but traversed at different speeds in clock time. If the longer policy contains diagnostic information absent from the shorter policy, their terminal posteriors need not be equal. This restricted process hypothesis does not follow from the FEP alone and is not required to justify the $E$ - $PV$ allocation; it is an additional empirical link between linguistic evidence arrival and belief-updating dynamics.

### 4.8. Experimental identification and stopping

The allocation above is useful only if it generates selective empirical signatures. Manipulating task or social outcome values at fixed duration should change PV immediately. Manipulating a one-off deadline while holding terminal communicative consequences fixed should directly alter the contextual temporal factor $\Phi_t^{time}$, with little carryover once that temporal context is removed. Repeated deadline training should repeatedly bias policy selection through $\Phi_t^{time}$; if this produces repeated realized selection of the same policy, those selections should gradually alter $E_t^{hab}$, producing residual compression after the deadline is removed. Manipulating listener capacity should interact with local density and partner identity through the comprehension likelihood rather than produce a partner-independent duration penalty. Terminal-equivalent policy pairs are particularly diagnostic because they hold the terminal G-based ranking constant while varying duration.

Stopping can be included explicitly in the policy set:

$$\pi_{stop} \in \Pi_t. \qquad (113)$$

Using the decomposition of the policy prior in Equation (97), define the policy log score, up to a policy-independent normalization constant, as

$$\mathcal{S}_t(\pi) = \ln E_t^{hab}(\pi) + \ln \Phi_t^{time}(\pi \mid c_t) - F_t(\pi) - \gamma_t G_t(\pi). \qquad (114)$$

Under maximum-a-posteriori policy selection, or in the high-policy-precision limit, stopping is selected when

$$\mathcal{S}_t(\pi_{stop}) \geq \max_{\pi \neq \pi_{stop}} \mathcal{S}_t(\pi). \qquad (115)$$

Under the rate formulation, no continuation is worthwhile when

$$\max_{\pi \neq \pi_{stop}} \frac{G_t(\pi_{stop}) - G_t(\pi)}{T(\pi)} \leq \mu_t, \qquad (116)$$

in the case of $T(\pi_{stop}) = 0$, $E_t^{hab}(\pi) = E_t^{hab}(\pi_{stop})$ and $F_t(\pi) = F_t(\pi_{stop})$ in Equation (114).

Thus, a clarification policy may still have positive expected information gain and even lower EFE than stopping, but it can nevertheless be rejected when its prospective EFE improvement per unit time is smaller than the current temporal opportunity rate.

In ordinary dialogue, however, closing is usually a jointly managed sequence rather than an instantaneous unilateral action. A minimal closing policy can be represented as

$$\pi_{close} = \left(a_{preclose}^{A}, a_{uptake}^{B}, a_{terminal}^{A/B}\right), \qquad (117)$$

and a repair sequence as

$$\pi_{repair} = \left(a_{initiate}^{A}, a_{solution}^{B}, a_{accept/reject}^{A}\right). \qquad (118)$$

Their later components depend on the partner's response and are therefore evaluated through the recursive partner model developed in Section 3. The unilateral stop policy remains a useful limiting case for tasks in which one agent can end the interaction directly.

The central claim of this section is consequently not that an RSA cost has been exhaustively decomposed. Static policy scores can be reparameterized, and the same one-step utterance distribution can arise from several allocations. The proposed model commits to one causally explicit allocation: outcome-valued communicative consequences enter $PV$, expected uncertainty resolution enters $IG$, outcome-generation constraints enter the likelihood, learned historical policy propensity enters $E^{hab}$, and residual context-sensitive duration factor enters $\Phi^{time}$. The normalized product $E^{hab}\Phi^{time}$ defines the single policy prior $E$. The value of this allocation lies not in static uniqueness, but in predicting different responses to temporal intervention, repeated selection, partner change, and transfer.

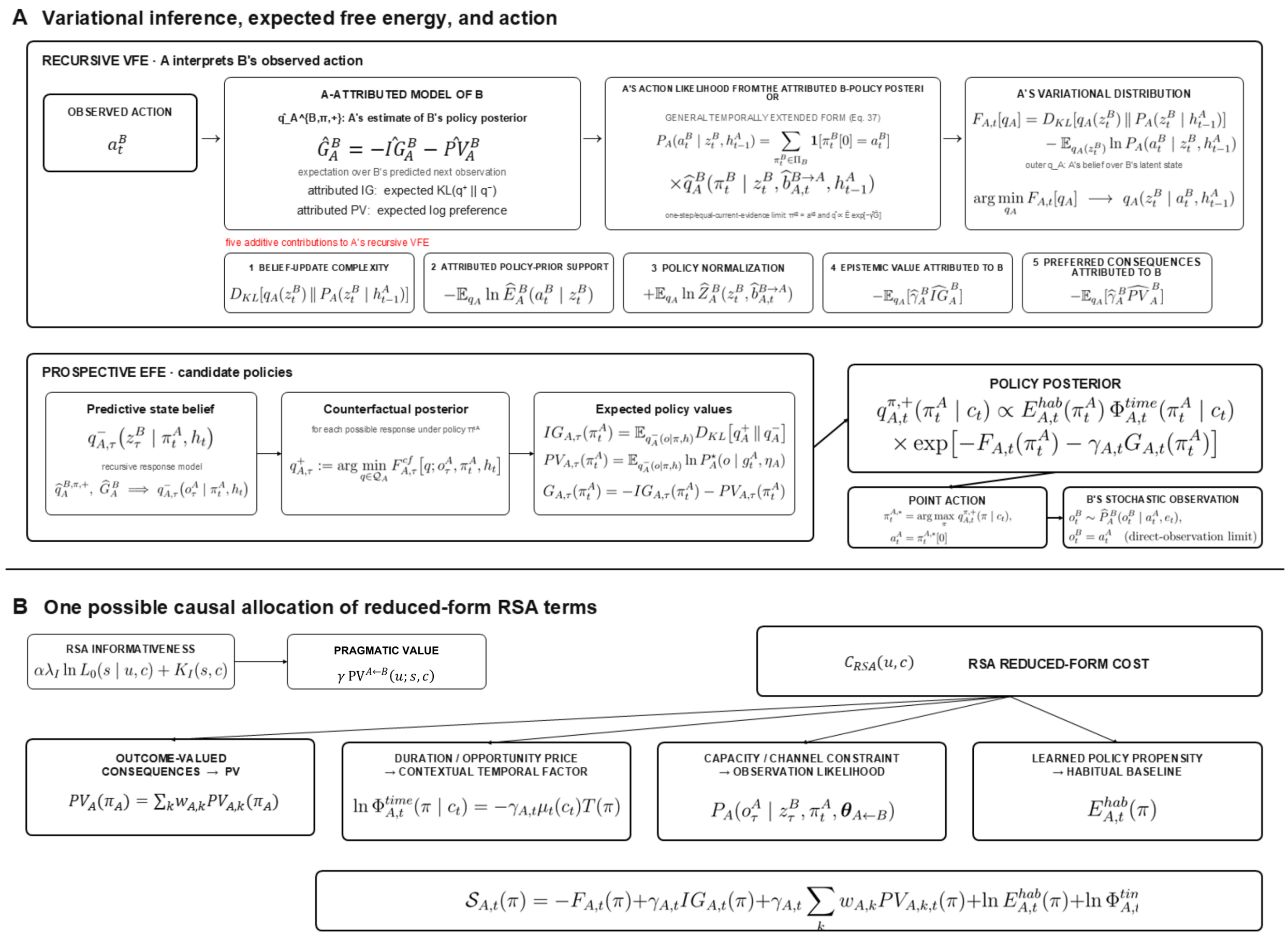

**Figure 2: Variational inference, expected free energy, and one possible causal allocation of reduced-form RSA terms. (A)** The upper pathway shows how A interprets B's observed action. In the general temporally extended model, the action likelihood $P_A(a_t^B \mid z_t^B, h_{t-1}^A)$ is obtained by marginalizing over the B-policy posterior $\hat{q}_A^B$ represented within A; only in the one-step macro-action limit can the observed action be identified directly with a policy. The negative log of this induced likelihood enters A's variational free energy (VFE). Under the one-step and equal-current-evidence assumptions used in Equation (50), the recursive VFE contains five additive contributions: A's belief-update complexity, policy-prior support attributed to B's observed action, normalization over B's alternative policies, epistemic value attributed to B, and preferred consequences attributed to B. The lower pathway shows A's prospective evaluation of its own candidate policies. The policy-conditioned predictive belief $q_{A,\tau}^-$ is mapped, for each possible future response, to the counterfactual posterior $q_{A,\tau}^+$ by minimizing the counterfactual VFE $F_{A,\tau}^{cf}$. Expected information gain and pragmatic value are then combined as $G_{A,\tau} = -IG_{A,\tau} - PV_{A,\tau}$. A's policy posterior combines the learned habitual baseline $E_{A,t}^{hab}(\pi)$ with the context-sensitive temporal factor $\Phi_{A,t}^{time}(\pi \mid c_t)$ before policy selection. The displayed point action is a MAP approximation, whereas B's observation of A may remain stochastic; direct observation is recovered as a limiting case. **(B)** Reduced-form RSA terms can be assigned to different locations in the active-inference architecture as a causal modeling commitment. RSA informativeness enters through attributed pragmatic value under the restricted affine mapping $\gamma\, \mathrm{PV}^{A\leftarrow B}(u; s, c) = \alpha\lambda_I \ln L_0(s \mid u, c) + K_V(s, c)$; attributed information gain has no RSA counterpart. Outcome-valued consequences enter pragmatic value; duration and opportunity price enter $\Phi_{A,t}^{time}$; capacity and channel constraints enter the observation likelihood; and learned policy propensity enters $E_{A,t}^{hab}$. These components contribute separately to the active-inference policy score in Equation (78). The allocation is neither unique nor exhaustive, and the restricted RSA correspondence does not imply a general identity between RSA informativeness and expected log preference.

## 5. Examples of Pragmatic Active Inference

The calculations in this section are thought experiments rather than empirical results. Numerical values are selected for transparency and are not fitted to behavioral data. The purpose of the toy examples is to show how the proposed variables interact and how the framework can reproduce standard RSA results under simplifying assumptions while generating different dynamics when inference, partner learning, habit, misconception, or stopping are included.

We distinguish two classes of examples:

1. Cases in which the proposed model produces the same choice as a traditional RSA model; and
2. Cases in which the inclusion of partner learning, temporal dynamics, or mutually coupled generative models produces behavior not represented in a conventional single-shot RSA calculation.

### 5.1. Examples for the factorized latent state and partner-model parameters

Recall the factorized partner latent state and the partner-model parameters introduced in Section 3:

$$z_t^B = (r_t^B, \ell_t^B, f_t^B, d_t^B, k_t^B, v_t^B, g_t^B, Q_t^{AB}, C_t^{AB}, \tau_t^{AB}, \mathcal{A}_t^{AB})$$

$$\theta_{A\leftarrow B} = \left(\theta_{A\leftarrow B}^{\text{lik}}, \theta_{A\leftarrow B}^{\text{trans}}, \rho_B, \eta_B, \gamma_B, m^B, E_B^{hab}\right). \qquad (119)$$

Table 1 provides a minimal example for each latent state factor.

**Table 1. Illustrative observations associated with different latent pragmatic factors.**

| Factor | Candidate states | Example observation | Primary inference |
|---|---|---|---|
| $r_t^B$ | red cup / blue cup | "The cup on the left" | referent |
| $\ell_t^B$ | financial bank / river bank | "I withdrew money from the bank" | lexical sense |
| $f_t^B$ | question / request | "Can you close the window?" followed by the speaker waiting | speech-act force |
| $d_t^B$ | establish location / establish price | "It's on the second floor" | discourse goal |
| $k_t^B$ | inferred acceptance / non-acceptance of the proposed common-ground increment | "Turn on the projector" | common-ground commitment |
| $v_t^B$ | sincere / ironic | "That went really well" after an obvious failure | social stance |

| Factor | Candidate states | Example observation | Primary inference |
|---|---|---|---|
| $g_t^B$ | information seeking / action induction | "Do you have a pen?" while presenting an unsigned form | current goal |
| $Q_t^{AB}$ | location QUD / price QUD | "Where is it?" versus "How much is it?" | question under discussion |
| $C_t^{AB}$ | location not grounded / location grounded | both speakers subsequently use "the second-floor projector" without repair | public commitment / common-ground record |
| $\tau_t^{AB}$ | first pair part / second pair part | "Yes" immediately following a polar question | turn / sequence position |
| $A_t^{AB}$ | reference game / action-coordination task | "Move the red block to the left slot" during a shared manipulation task | joint activity / task |

These factors are not assumed to be independent. For example, evidence that changes $Q_t^{AB}$ can change the relevance of a private discourse goal $d_t^B$, and an acknowledgment can jointly support $k_t^B$ while updating the interaction-level commitment record $C_t^{AB}$. The factorization therefore identifies distinct update targets without requiring a product of independent marginals.

For any one binary factor $z_j$, suppose that the prior is uniform:

$$P\left(z_j = 1\right) = P\left(z_j = 0\right) = 0.5. \qquad (120)$$

If the observation likelihood is

$$P\left(o \mid z_j = 1\right) = 0.9, \qquad P\left(o \mid z_j = 0\right) = 0.1, \qquad (121)$$

then

$$q\left(z_j = 1 \mid o\right) = \frac{0.9(0.5)}{0.9(0.5) + 0.1(0.5)} = 0.9. \qquad (122)$$

The same Bayesian structure may therefore update different pragmatic factors. In a realistic discourse model, these factors need not be independent, and one observation may update several components simultaneously. What differs is the target and expected persistence of the update: current private and interaction-level factors may change within a turn or sequence, whereas partner-model parameters accumulate across repeated evidence. A single observation may also update several components jointly.

Table 2 provides corresponding examples for the relatively persistent partner-model parameters in $\theta_{A \leftarrow B}$. These quantities need not be immutable, but they are expected to change more slowly and to carry over across local discourse states.

**Table 2. Illustrative observations associated with relatively persistent partner-model parameters.**

| Factor | Candidate states | Example observation | Primary inference |
|---|---|---|---|
| $\theta_{A\leftarrow B}^{\text{lik}}$ | high / low probability of a behavior under a given pragmatic state | B repeatedly uses the same shorthand under the same referential condition | partner-specific likelihood regularity |
| $\theta_{A\leftarrow B}^{\text{trans}}$ | high / low persistence of a latent state across turns | B's stance or interaction goal repeatedly persists versus changes across adjacent turns | partner-specific transition regularity |
| $\rho_B$ | reliable / unreliable | repeated mismatch between claims and observable facts | speaker reliability |
| $\eta_B$ | directness preference / politeness preference | repeated use of indirect requests | stable outcome preference |
| $\gamma_B$ | high / low policy precision | steep versus shallow changes in choice probability as controlled policy-score differences are varied | policy precision |
| $m^B$ | shorthand-familiar / shorthand-unfamiliar | immediate understanding versus clarification after shorthand | partner-model type / convention familiarity |
| $E_B^{\text{hab}}$ | shorthand-biased / full-form-biased habitual baseline | persistent preference for one policy when current evidence, prospective value, and temporal context are matched | learned habitual policy propensity |

Thus, the split between Tables 1 and 2 is not a split between variables that can and cannot be inferred. Both are uncertain within A's generative model. The distinction is temporal and causal: $z_t^B$ represent the current interaction and partner state, whereas $\theta_{A\leftarrow B}$ represents relatively persistent regularities.

Evidence can propagate across these levels. For example, a clarification after a shorthand expression may update the current referential hypothesis while also lowering A's posterior probability that B is shorthand-familiar. The model therefore permits coupling across factors and timescales while keeping the variables themselves formally distinct.

### 5.2. Example I: Recovering the RSA choice

Suppose the visual context contains three objects:

$$w_1 = \text{red square}, \qquad w_2 = \text{red circle}, \qquad w_3 = \text{blue square}. \qquad (123)$$

The speaker's target is $w_1$. Two utterances are available:

$$u_S = \text{“red”}, \qquad u_F = \text{“red square”}. \qquad (124)$$

The literal listener assigns

$$L_0(w_1 \mid u_S) = 0.5 \qquad (125)$$

and

$$L_0(w_1 \mid u_F) = 1. \qquad (126)$$

Let policy durations be

$$T(u_S) = 1, \qquad T(u_F) = 2. \qquad (127)$$

To isolate the RSA informativeness-duration trade-off, assume that no outcome value other than the listener's identification of the target differs between the two utterances, and that the speaker knows the target, so that the reduced information gain is common across the alternatives. Following the mapping in Section 3.12, represent RSA informativeness by the reduced pragmatic-value contribution, up to an utterance-independent constant:

$$\gamma\,\mathrm{PV}(u) = \alpha \ln L_0(w_1 \mid u) + K_V, \qquad \gamma\,\mathrm{IG}(u) = K_I. \qquad (128)$$

Here $K_V$ and $K_I$ are common across utterance alternatives and cancel under normalization. The EFE contribution is

$$G(u) = -\mathrm{IG}(u) - \mathrm{PV}(u) = -\frac{1}{\gamma}(\alpha \ln L_0(w_1 \mid u) + K_I + K_V). \qquad (129)$$

Represent the duration-dependent contribution using the contextual temporal factor introduced in Section 4:

$$\Phi^{\mathrm{time}}(u \mid c) = \exp[-\gamma\mu(c)T(u)]. \qquad (130)$$

Unlike a second policy prior, $\Phi^{\mathrm{time}}$ is a positive context-dependent factor. The single normalized policy prior remains $E(u \mid c)$, obtained by normalizing $E^{\mathrm{hab}}(u)\Phi^{\mathrm{time}}(u \mid c)$ across utterance alternatives, as in Equation (97). For this equivalence example, set the learned habitual baseline to be uniform, $E^{\mathrm{hab}}(u_S) = E^{\mathrm{hab}}(u_F) = 1/2$. Habitual support therefore contributes no relative preference between the utterances; the current temporal context supplies the duration-dependent tilt.

Set

$$\alpha = \gamma = 4, \qquad \mu(c) = 0.1, \qquad K_I = K_V = 0. \qquad (131)$$

This choice gives $\mathrm{PV}(u_S) = \ln 0.5$ and $\mathrm{PV}(u_F) = 0$, both nonpositive as required of an expected log preference; the common offsets do not affect the normalized policy distribution. Hold the current-evidence term $F(u)$ common across alternatives. From Equation (98), the policy posterior is proportional to $E^{\mathrm{hab}}(u)\Phi^{\mathrm{time}}(u \mid c)\exp[-F(u) - \gamma G(u)]$. Substituting Equations (128)-(131) and dropping terms common across utterances gives

$$q(u \mid c) \propto \exp[4 \ln L_0(w_1 \mid u) - 0.4T(u)]. \qquad (132)$$

For the short utterance,

$$W(u_S) = \exp[4\ln 0.5 - 0.4] = \exp[-3.1726] = 0.0419. \qquad (133)$$

For the full utterance,

$$W(u_F) = \exp[4\ln 1 - 0.8] = \exp[-0.8] = 0.4493. \qquad (134)$$

After normalization,

$$q(u_S \mid c) = \frac{0.0419}{0.0419 + 0.4493} = 0.0853, \qquad (135)$$

and

$$q(u_F \mid c) = 0.9147. \qquad (136)$$

This is identical to an RSA speaker with

$$S_1(u \mid w_1) \propto \exp\{4[\ln L_0(w_1 \mid u) - 0.1T(u)]\}, \qquad (137)$$

because

$$\gamma\mu(c)T(u) = 4[0.1T(u)] = 0.4T(u). \qquad (138)$$

The resulting policy distribution is therefore identical to the RSA speaker distribution in this restricted example. The equivalence arises from two distinct mappings: RSA informativeness is represented by the reduced pragmatic-value contribution up to an utterance-independent constant, whereas the RSA duration cost is reproduced by the contextual temporal factor $\Phi^{\text{time}}$. The learned habitual prior $E^{\text{hab}}$ is held uniform and therefore does not contribute to the policy ranking. This equality is static and does not identify a unique causal implementation of the RSA score (Section 4.1).

### 5.3. Example II: Partner-dependent time efficiency and policy adaptation

The preceding example showed that the proposed architecture can reproduce a static RSA choice distribution. The present example asks a different question: when can a shorter policy be more time-efficient, and how does uncertainty about the partner convert that efficiency contrast into an actual policy choice? We first use a reduced target-evidence calculation to isolate the role of time. We then embed the same qualitative familiar-versus-unfamiliar contrast in a fuller outcome model in which A is uncertain about B's partner type, evaluates policies through expected free energy, and updates that partner model after feedback.

#### 5.3.1. Reduced target-evidence comparison

First consider a shorthand-familiar listener B. Suppose B's prior probability for a designated target is

$$q_{B,0}(w^{\star}) = 0.50. \qquad (139)$$

A shorthand policy $\pi_S$ and a full policy $\pi_F$ are predicted to yield the terminal listener beliefs

$$q_B(w^{\star} \mid \pi_S, m_F) = 0.95, \qquad q_B(w^{\star} \mid \pi_F, m_F) = 0.99. \qquad (140)$$

where $m_F$ denotes the shorthand-familiar partner hypothesis. The corresponding target-directed evidence gains are

$$\Delta J(\pi_S; w^{\star}, m_F) = \ln 0.95 - \ln 0.50 = 0.6419. \qquad (141)$$

and

$$\Delta J(\pi_F; w^{\star}, m_F) = \ln 0.99 - \ln 0.50 = 0.6831. \qquad (142)$$

The full form therefore provides more terminal target-evidence gain:

$$0.6831 > 0.6419. \qquad (143)$$

The quantity compared here is pointwise evidence gain for a designated target, not expected mutual information over the full latent state or parameter space. A policy may provide greater target-evidence gain while providing less expected information about other latent variables, or vice versa. Accordingly, $\Delta J$ should not be identified with the expected information-gain term $IG$ entering expected free energy.

Now let

$$T(\pi_S) = 1, \qquad T(\pi_F) = 3. \qquad (144)$$

Using the target-evidence rate introduced in Section 4,

$$R_J(\pi_S; w^{\star}, m_F) = \frac{0.6419}{1} = 0.6419. \qquad (145)$$

whereas

$$R_J(\pi_F; w^{\star}, m_F) = \frac{0.6831}{3} = 0.2277. \qquad (146)$$

The full form is superior in terminal target probability, whereas the shorthand is superior in target-evidence gain per unit time. This is the sense in which the shorthand is more time-efficient in the present reduced calculation.

Now suppose instead that B is shorthand-unfamiliar, denoted by $m_U$, and that the shorthand increases the target probability only to

$$q_B(w^{\star} \mid \pi_S, m_U) = 0.55. \qquad (147)$$

Then

$$\Delta J(\pi_S; w^\star, m_U) = \ln 0.55 - \ln 0.50 = 0.0953. \quad (148)$$

and

$$R_J(\pi_S; w^\star, m_U) = 0.0953. \quad (149)$$

If the full form still yields the terminal posterior 0.99, then

$$R_J(\pi_F; w^\star, m_U) = 0.2277 > 0.0953 = R_J(\pi_S; w^\star, m_U). \quad (150)$$

Compression is therefore not intrinsically time-efficient. Its efficiency depends on the partner model:

$$\boxed{\begin{array}{c}\text{shorthand} - \text{familiar partner} \Rightarrow \text{shorthand may maximize} \\ \text{target-evidence gain per unit time}\end{array}} \quad (151)$$

whereas

$$\boxed{\begin{array}{c}\text{shorthand} - \text{unfamiliar partner} \Rightarrow \text{the full form may yield greater} \\ \text{target-evidence gain per unit time}\end{array}}. \quad (152)$$

These rate comparisons are diagnostic rather than a complete policy-selection rule. In particular, $R_J$ is not itself inserted into the policy posterior or into the contextual temporal factor. General policy selection still depends on the full score in Equation (98): prospective communicative consequences remain in G, learned historical propensity enters $E^{\text{hab}}$, and $\Phi^{\text{time}}$ supplies the context-dependent price of policy duration. The temporal factor therefore affects current policy support directly, but it may also have an indirect longer-term consequence for $E^{\text{hab}}$. If the same temporal context repeatedly favors and thereby produces the realized selection of a particular policy, those selections are accumulated by the habitual update rule and can gradually alter the learned habitual prior. Policy duration does not directly update $E^{\text{hab}}$; rather, duration-dependent selection can generate a persistent habitual carryover through repeated policy realization. Thus, Equations (139)–(152) establish a partner-dependent time-efficiency contrast, but do not yet determine A's policy posterior.

### 5.3.2. Policy selection under uncertain partner familiarity

We now embed the same qualitative contrast in a fuller outcome model in which A is uncertain about B's familiarity with the shorthand convention. The numerical specification below is not algebraically derived from Equations (139)-(152); it instantiates the same familiar-versus-unfamiliar distinction while allowing policy selection and partner-model updating to be computed explicitly.

Let A's persistent partner model include the uncertain hypothesis

$$m^B \in \{m_F = \text{shorthand-familiar},\ m_U = \text{shorthand-unfamiliar}\}. \qquad (153)$$

Before the current feedback, A's belief over B's partner type is

$$q_t(m_F) = 0.8, \qquad q_t(m_U) = 0.2. \qquad (154)$$

Possible outcomes of B's response are $j_C$ (correct understanding),$j_Q$ (clarification), and $j_E$ (error):

$$j \in \{j_C, j_Q, j_E\}. \qquad (155)$$

These outcomes are B-side response consequences represented within A's partner model. For shorthand $\pi_S$, let

$$P(j \mid \pi_S, m_F) = (0.90, 0.08, 0.02), \qquad (156)$$

and

$$P(j \mid \pi_S, m_U) = (0.40, 0.40, 0.20). \qquad (157)$$

For the full form $\pi_F$, let

$$P(j \mid \pi_F, m_F) = (0.96, 0.03, 0.01), \qquad (158)$$

and

$$P(j \mid \pi_F, m_U) = (0.85, 0.10, 0.05). \qquad (159)$$

Let A's preference distribution over these outcomes be

$$P^\star(j_C) = 0.75, \qquad P^\star(j_Q) = 0.20, \qquad P^\star(j_E) = 0.05. \qquad (160)$$

For either policy, A's predictive outcome distribution marginalizes over its current uncertainty about B's partner type:

$$q_t(j \mid \pi) = \sum_{m^B} q_t(m^B) P(j \mid \pi, m^B). \qquad (161)$$

For this example, the epistemic term is evaluated over uncertainty in the relatively persistent partner-model variable $m^B$. We therefore write

$$IG_t^m(\pi) = \sum_j q_t(j \mid \pi) D_{KL}[q_t(m^B \mid j, \pi) \parallel q_t(m^B)]. \qquad (162)$$

This is the partner-parameter/model-index analogue of the information-gain term in Equation (61), rather than information gain about a rapidly changing discourse state. The pragmatic-value term is

$$PV_t(\pi) = \sum_j q_t\,(j \mid \pi)\ln P^\star(j). \qquad (163)$$

For shorthand,

$$q_t(j \mid \pi_S) = 0.8(0.90,0.08,0.02) + 0.2(0.40,0.40,0.20) = (0.800,0.144,0.056). \qquad (164)$$

The resulting values are

$$IG_t^m(\pi_S) = 0.1079, \qquad (165)$$

$$PV_t(\pi_S) = -0.6297, \qquad (166)$$

and

$$G_t(\pi_S) = -IG_t^m(\pi_S) - PV_t(\pi_S) = 0.5218. \qquad (167)$$

For the full form,

$$q_t(j \mid \pi_F) = (0.938,0.044,0.018), \qquad (168)$$

$$IG_t^m(\pi_F) = 0.0138, \qquad (169)$$

$$PV_t(\pi_F) = -0.3946, \qquad (170)$$

and

$$G_t(\pi_F) = 0.3808. \qquad (171)$$

The full form therefore has lower expected free energy, even though shorthand is more informative about the partner-type variable itself. This is a useful epistemic-pragmatic trade-off: shorthand is more diagnostic of whether the convention is shared, whereas the full form more reliably produces the preferred communicative outcome.

Suppose, however, that repeated successful interaction has produced a strong learned shorthand habit:

$$E_t^{\text{hab}}(\pi_S) = 0.8, \qquad E_t^{\text{hab}}(\pi_F) = 0.2. \qquad (172)$$

To isolate partner-model updating from immediate time pressure, set $\Phi_t^{\text{time}}(\pi_S \mid c_t) = \Phi_t^{\text{time}}(\pi_F \mid c_t) = 1$ and hold the current-evidence term $F_t(\pi)$ common across the two alternatives. With A's policy precision

$$\gamma_A = 5, \qquad (173)$$

Equation (98) reduces for the present comparison to

$$q_t(\pi \mid c_t) \propto E_t^{\text{hab}}(\pi)\exp[-5G_t(\pi)]. \qquad (174)$$

This gives

$$q_t(\pi_S \mid c_t) = 0.6641, \qquad q_t(\pi_F \mid c_t) = 0.3359. \qquad (175)$$

Despite its higher EFE, shorthand remains more probable because of the learned habitual policy prior. The example therefore separates prospective policy evaluation from accumulated historical propensity: $G_t$ favors the full form, whereas $E_t^{\mathrm{hab}}$ initially favors shorthand.

**5.3.3. Clarification-driven partner updating and policy reversal**

Now suppose that B requests clarification after A uses shorthand, yielding outcome $j_Q$. Because clarification is much more likely under the shorthand-unfamiliar hypothesis, Bayes' rule gives

$$q_{t+1}(m_F \mid j_Q, \pi_S) = \frac{0.8(0.08)}{0.8(0.08) + 0.2(0.40)} = 0.4444. \qquad (176)$$

and

$$q_{t+1}(m_U \mid j_Q, \pi_S) = 0.5556. \qquad (177)$$

The unexpected clarification therefore substantially increases A's posterior probability that B is shorthand-unfamiliar. Recomputing expected free energy under this updated partner model gives

$$G_{t+1}(\pi_S) = 0.8070, \qquad (178)$$

and

$$G_{t+1}(\pi_F) = 0.4477. \qquad (179)$$

For this immediate partner-model update, hold the slower habitual component fixed at its pre-feedback value, $E_{t+1}^{\mathrm{hab}} = E_t^{\mathrm{hab}}$. The temporal factor also remains common and equal to one. The resulting policy posterior is

$$q_{t+1}(\pi_S \mid c_{t+1}) = 0.3988, \qquad (180)$$

and

$$q_{t+1}(\pi_F \mid c_{t+1}) = 0.6012. \qquad (181)$$

The policy ranking therefore reverses:

$$\boxed{q_t(\pi_S \mid c_t) = 0.6641 \rightarrow q_{t+1}(\pi_S \mid c_{t+1}) = 0.3988}. \qquad (182)$$

The computational path is

$$\boxed{\text{clarification} \rightarrow q(m^B) \rightarrow q(j \mid \pi) \rightarrow (IG^m, PV) \rightarrow G(\pi) \rightarrow q(\pi) \text{ reversal}}. \qquad (183)$$

Throughout this immediate reversal, the learned habitual baseline, the contextual temporal factor, and the current-evidence contribution are held fixed. The reversal is therefore attributable to the updated partner model and its consequences for predicted outcomes and EFE, rather than to a newly imposed duration cost.

More generally, the architecture distinguishes update targets with different temporal signatures:

$q_t(z_t^B)$: B-private pragmatic state and interaction-level pragmatic state presented by A;

$q_t(\theta_{A\leftarrow B})$: relatively persistent partner-model variables; including $\rho_B$ , $\eta_B$,$\gamma_B$ and $m^B$ ;

$E_t^{\mathrm{hab}}(\pi)$: learned historical policy propensity;

$\Phi_t^{\mathrm{time}}(\pi \mid c_t)$: transient context-dependent temporal modulation.

These quantities need not update on the same turn or at the same timescale. In the present example, a clarification primarily changes $q_t(\theta_{A\leftarrow B})$ through the partner-type component $m^B$; the habitual prior is deliberately held fixed over the immediate update, and the temporal factor is held constant.

The diagnostic result is the cross-over after partner-model updating: the same policy set and temporal context generate a different policy ranking because evidence about the partner changes the predicted outcome distribution. A direct test should therefore compare a model that updates a partner-specific convention or familiarity parameter with a model that merely adds a trial-specific cost while holding temporal context fixed. The partner-model account additionally predicts partner-specific carryover: greater support for the full form should persist when A next addresses the same shorthand-unfamiliar partner, but need not transfer to a new shorthand-familiar partner.

### 5.4 . Example III: Mutually stabilizing misconception under mis-specified partner models

Suppose that two agents use the name *Alex* for different people:

$$r_1 = \text{Alex Kim}, \qquad r_2 = \text{Alex Lee}. \qquad (184)$$

Agent $A$ intends $r_1$ and maintains

$$q_A(r_1) = 0.9, \qquad q_A(r_2) = 0.1. \qquad (185)$$

Agent $B$ interprets the same name as $r_2$:

$$q_B(r_1) = 0.1, \qquad q_B(r_2) = 0.9. \qquad (186)$$

Agent $A$ says, “Alex will give the presentation”, and $B$ replies, “Yes, I know”. Both assume that the other uses the same referent. Under $A$’s model,

$$P_A(\text{“yes”} \mid r_1) = P_A(\text{“yes”} \mid r_2) = 0.9, \qquad (187)$$

while under $B$'s model,

$$P_B(\text{"yes"} \mid r_1) = P_B(\text{"yes"} \mid r_2) = 0.9. \qquad (188)$$

At a tight variational bound, the observation contributes surprisal

$$-\ln 0.9 = 0.1054 \qquad (189)$$

for each agent. Both agents may therefore experience low variational free energy under their own generative models. Their beliefs are nevertheless strongly misaligned:

$$D_{\mathrm{KL}}[q_A(r) \parallel q_B(r)] = 0.9\ln\frac{0.9}{0.1} + 0.1\ln\frac{0.1}{0.9} = 1.7578 \text{ nats.} \qquad (190)$$

While KL divergence is not assumed to be symmetric in general, the reverse divergence $D_{\mathrm{KL}}[q_B(r) \parallel q_A(r)]$ also takes the same value because the two posteriors are mirror-symmetric in this toy case.

Thus,

$$\boxed{F_A \text{ may be low and } F_B \text{ may be low} \quad \text{while} \quad q_A(r) \neq q_B(r)}. \qquad (191)$$

The two agents can occupy a self-confirming, comparatively low-VFE regime in which each model predicts the observations generated by the other, despite systematic divergence between their referential beliefs and, in B's case, from A's intended referent. This is not necessarily a local minimum in a formally specified continuous landscape; it is a stable or metastable pattern under the stated update rules.

To generalize, let $p \in (\frac{1}{2}, 1)$ and define mirror-symmetric referential beliefs over $(r_1, r_2)$:

$$q^A(r) = (p, 1-p), \qquad q^B(r) = (1-p, p) \qquad (192)$$

Their directed belief divergence is

$$D_{\mathrm{KL}}[q_A(r) \parallel q_B(r)] = (2p-1)\ln\frac{p}{1-p}, \qquad (193)$$

which is zero at $p = 0.5$ and increases as the agents become more confident in opposing referents.

Figure 3B compares two likelihood structures. In the nondiagnostic condition used in the example,

$$P_i(o \mid r_1) = P_i(o \mid r_2) = \lambda = 0.9. \qquad (194)$$

The likelihood ratio is therefore one, the observed reply does not change the referential odds, and, at a tight variational bound,

$$F_i^* = -lnP_i(o) = -ln\,\lambda = 0.1054\,, \quad (195)$$

independently of p. For the idealized diagnostic comparator shown in Figure 3B,

$$P_i(o \mid r_i^*) = 1, \qquad P_i(o \mid r_i^o) = 0, \qquad (196)$$

where $r_i^*$ is agent $i$ ‘s current favored referent and $r_i^o$ is the alternative. Hence,

$$P_i(o) = p\,, \qquad F_i^* = -\ln p. \qquad (197)$$

At $p = 0.90$, the diagnostic and nondiagnostic observations can therefore have the same surprisal, 0.1054 nats, even though only the diagnostic likelihood distinguishes the referents. Diagnosticity is determined by the likelihood ratio across hypotheses, not by surprisal alone.

Low expected information gain for clarification is an additional consequence of the mis-specified generative model, rather than a consequence of low variational free energy alone. Specifically,

$$IG_i(\pi_{clarification}) \approx 0 \quad (198)$$

only when the agent predicts the same distribution of clarification outcomes under the competing referents:

$$P_i(\, o_\tau \mid r_1, \pi_{clarification}\,) = P_i(\, o_\tau \mid r_2,\ \pi_{clarification}\,)\ for\ all\ o_\tau\,. \quad (199)$$

If the predicted clarification outcomes differ across the referential hypotheses, clarification has positive expected information gain and the self-maintaining regime need not persist.

The misconception may remain stable when

1. each agent's current model assigns high likelihood to the observed response;
2. no observation has yet distinguished $r_1$ from $r_2$;
3. both agents assign high posterior confidence to their current referential hypotheses;
4. the mis-specified model predicts approximately state-independent clarification outcomes, such that $IG_i(\pi_{clarification}) \approx 0$.

The primary update target depends on the evidence. A uniquely identifying visual, documentary, third-party, or repair cue primarily updates the agent-specific current referential posterior $q_i(r_i)$. By contrast, repeated partner-specific mismatches may additionally support slower learning about the partner’s naming or likelihood mapping $\theta_{A\leftarrow B}^{\mathrm{lik}}$ or about partner reliability $\rho_B$. A single disambiguating observation should therefore not be assigned directly to persistent partner-model learning without an additional attributional assumption.

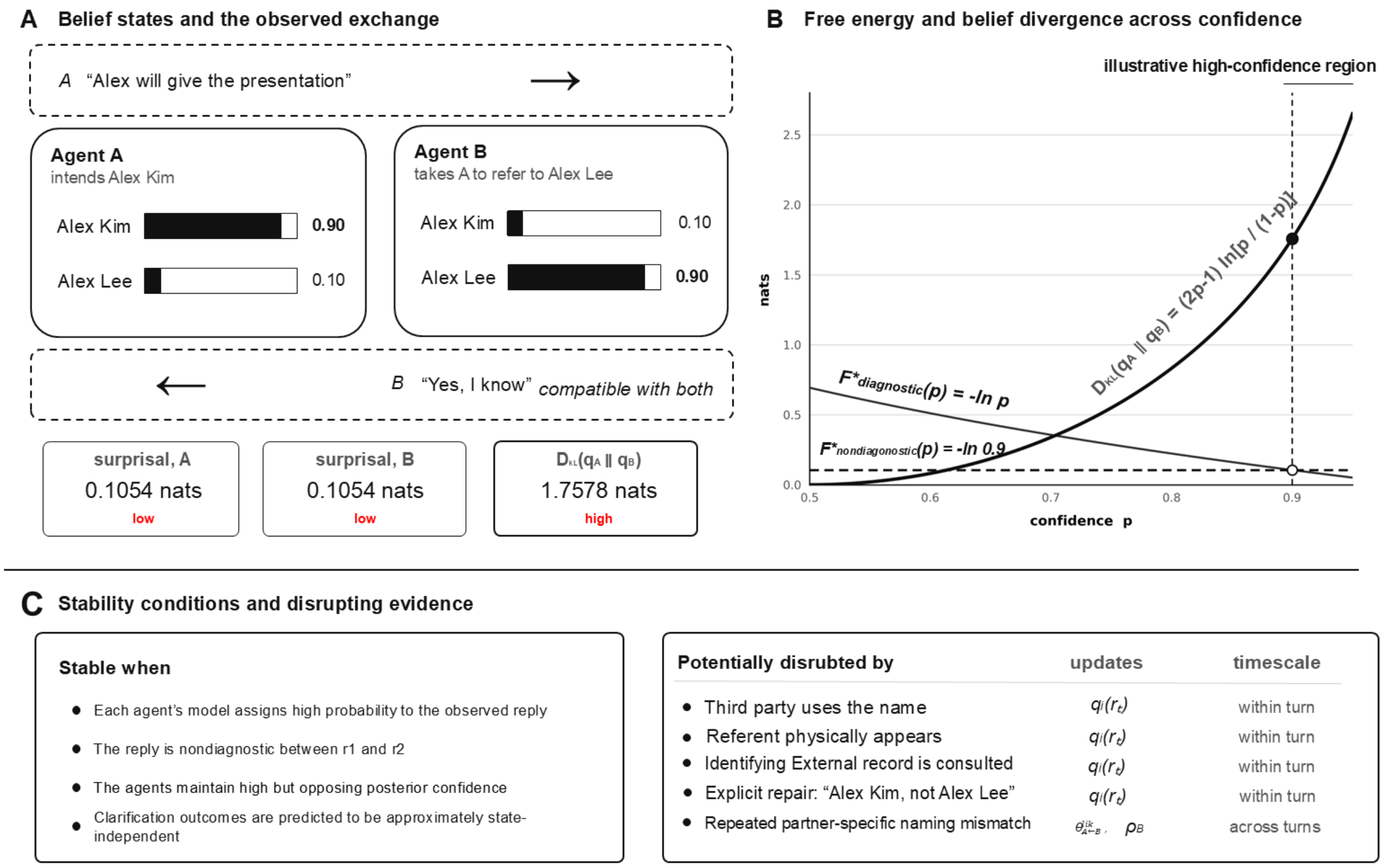

**Figure 3: Low free energy does not entail shared belief: non-disconfirmed misunderstanding under misspecified partner models. (A)** The worked case of Section 5.4. A intends *Alex Kim*, whereas B takes A to refer to *Alex Lee*; each assigns probability 0.90 to its own referent. B's reply is equally probable under either referential hypothesis. Thus, at a tight variational bound, each agent incurs a surprisal of 0.1054 nats–both generative models fit the observed exchange well–while their referential posteriors differ by $D_{\mathrm{KL}}$ $(q_A \parallel q_B) = 1.7578$ nats. Neither agent receives evidence from the reply that anything has gone wrong. **(B)** The example is generalized using mirror-symmetric referential beliefs, $q_A(r) = (p, 1-p)$ and $q_B(r) = (1-p, p)$. The thick solid curve shows their divergence, $D_{\mathrm{KL}}(q_A \parallel q_B) = (2p-1)\ln[p/(1-p)]$. The thin solid curve shows tight-bound free energy for the idealized diagnostic comparator, $F^*_{\mathrm{diagnostic}}(p) = -\ln p$, whereas the dashed curve shows the nondiagnostic condition used in Panel A, $F^*_{\mathrm{nondiagnostic}}(p) = -\ln 0.90$. At $p = 0.90$, the diagnostic and nondiagnostic observations have the same tight-bound free energy, 0.1054 nats, although only the diagnostic likelihood distinguishes the competing referents. Increasing opposing posterior confidence therefore increases belief divergence while leaving the surprisal of the nondiagnostic reply unchanged. The marked points indicate the values used in Section 5.4. **(C)** The misunderstanding can persist when each agent's model assigns high probability to the observed reply, the reply is nondiagnostic between the referents, the agents maintain high but opposing posterior confidence, and clarification outcomes are predicted to be approximately state-independent. It can be disrupted by evidence that separates the referents. Diagnostic third-party usage, the physical appearance of a referent, an identifying external record, or explicit referential repair primarily updates the current referential posterior $q_i(r_t)$ within a turn. Repeated partner-specific naming mismatches may additionally update persistent partner-model parameters, such as $\theta^{\mathrm{lik}}_{i \leftarrow j}$ or $\rho_j$, across turns. Low free energy therefore indicates model-relative fit rather than truth, shared reference, or convergence.

### 5.5. Example IV: Conversation stopping

At an early conversational turn, suppose that

$$G_t(\pi_{\text{stop}}) = 0.80, \qquad (200)$$

$$G_t(\pi_{\text{ask}}) = 0.30, \qquad (201)$$

and

$$T(\pi_{\text{ask}}) = 2. \qquad (202)$$

To illustrate the restricted rate form of the stopping criterion in Equation (116), let $T(\pi_{stop}) = 0$, $E_t^{hab}(\pi) = E_t^{hab}(\pi_{stop})$ and $F_t(\pi) = F_t(\pi_{stop})$. Then the prospective benefit per unit time is

$$R_t^G(\pi_{\text{ask}}) = \frac{G_t(\pi_{\text{stop}}) - G_t(\pi_{\text{ask}})}{T(\pi_{\text{ask}})} = 0.25. \qquad (203)$$

Suppose that the current context-dependent price of time is

$$\mu_t = 0.05. \qquad (204)$$

Because

$$0.25 > 0.05, \qquad (205)$$

the continuation policy outranks stopping under the MAP or high-policy-precision rate criterion. After several turns, uncertainty has decreased. Suppose that

$$G_{t+k}(\pi_{\text{stop}}) = 0.32, \qquad (206)$$

and

$$G_{t+k}(\pi_{\text{ask}}) = 0.26. \qquad (207)$$

Under the assumption that the duration of the question and the temporal opportunity rate remain unchanged, the new rate is

$$R_{t+k}^G(\pi_{\text{ask}}) = \frac{G_{t+k}(\pi_{\text{stop}}) - G_{t+k}(\pi_{\text{ask}})}{2} = 0.03, \qquad (208)$$

and now,

$$0.03 < 0.05. \qquad (209)$$

The stopping policy is therefore selected. Importantly, the question would still reduce expected free energy:

$$0.26 < 0.32. \qquad (210)$$

It is rejected because the reduction is too small relative to its duration.

$$\boxed{\text{prospective EFE improvement per unit time} < \text{current temporal opportunity rate}}. \qquad (211)$$

Stopping therefore does not require the prospective benefit of further clarification to vanish. The agent may terminate while a continuation policy would still improve expected free energy, provided that improvement is too small relative to its duration. Residual uncertainty may therefore remain at termination.

## 6. Linguistic and Empirical Implications

The proposed architecture matters for linguistics insofar as it sharpens distinctions that can be expressed in linguistic theory and tested in communicative behavior. Its basic claim is that every conversational move has a dual role: it is evidence about the agent who produced it and an intervention on the future state of the interaction. Consequently, pragmatic competence cannot be exhausted by computing a posterior over what a speaker meant. The posterior must alter predictions about uptake, repair, commitment, and action, and these predictions must shape the next policy.

This perspective connects several literatures that are often modeled separately. RSA supplies a principled distribution over choices under a specified semantics, alternative set, and utility. Dynamic pragmatics supplies interaction-level objects such as QUD and commitment stores. Dialogue research supplies the sequential organization of grounding, turn-taking, repair, and closing. Audience-design research supplies partner-specific learning and convention formation. Active inference supplies one architecture in which current-state inference, parameter learning, prospective policy evaluation, and habit can be separated while remaining coupled. The framework should be judged by whether this integration yields predictions that are unavailable from a simpler model, not by whether all components can be written under a free-energy objective.

### 6.1. Empirical dissociations

The causal allocation developed in Section 4 generates a concrete factorial program. The following manipulations can be matched on immediate utterance probability while testing different learning, persistence, and interaction signatures.

**1. One-off deadline**

*Location:* contextual temporal policy factor $\Phi_t^{\text{time}}$ .

*Immediate prediction:* when terminal communicative consequences and habitual support are matched, increasing the current temporal opportunity rate should shift probability toward shorter policies.

*Carryover:* if the deadline is removed before substantial repeated selection has accumulated, $\Phi_t^{\mathrm{time}}$ should return toward unity and the compression bias should largely disappear.

**2. Repeated deadline training**

*Location*: $\Phi_t^{\mathrm{time}}$ during each pressured choice, with indirect learning of $E_t^{\mathrm{h}ab}$ through the policies actually selected. Immediate prediction: repeated temporal pressure repeatedly favors shorter policies.

*Learning prediction*: if this bias produces repeated realized selection of the same short policy, those selections gradually increase its habitual prior support.

*Carryover*: after the deadline is removed and $\Phi_t^{\mathrm{time}}$ returns to unity, residual compression may persist through $E_t^{\mathrm{h}ab}$ and decay according to the habitual forgetting process. The temporal factor itself is not learned into the habitual prior: it changes which policies are realized, and those realized selections provide the history from which $E_t^{\mathrm{h}ab}$ is updated.

**3. Listener divided attention**

*Location:* the comprehension likelihood.

*Immediate prediction*: policies with high local processing load should become less likely to produce successful comprehension, favoring forms that distribute or reduce that load.

*Interaction*: the effect should depend on the current listener's capacity and local information density, with little carryover to a new attentive listener. Crucially, a capacity manipulation can alter policy ranking even when duration is held fixed, distinguishing it from a pure temporal-pressure effect.

**4. Listener familiarity with a partner-specific convention**

*Location:* the relatively persistent partner-model variable $m^B$ and the likelihoods conditioned on it.

*Immediate prediction:* shorthand should be favored when the current partner is believed to be shorthand-familiar.

*Updating prediction*: unexpected clarification after shorthand should lower the posterior probability of familiarity, change the predicted outcome distribution, and may reverse the policy ranking even when $E_t^{\mathrm{h}ab}$, $\Phi_t^{\mathrm{time}}$, and current policy evidence are held fixed. Carryover: the resulting adaptation should preferentially persist for the same partner rather than transfer indiscriminately to a new shorthand-familiar partner.

**5. Social risk or politeness stakes**

*Location:* prior preferences over predicted social outcomes and hence PV.

*Immediate prediction:* greater face, relationship, or strategic risk should favor more indirect, face-preserving, or deniable policies even when informativeness and duration are matched.

*Transfer:* the effect should generalize across policies with similar predicted social consequences rather than simply across policies of similar length.

**6. Production difficulty**

*Location:* not fixed a priori. Articulatory or execution failure should affect the relevant action or observation likelihood; aversive or energetically valued consequences may enter PV; and increased duration contributes through $\Phi_t^{\text{time}}$ only when taking longer carries a context-dependent opportunity cost.

*Immediate prediction:* difficult forms should be avoided, but the persistence and transfer pattern should depend on which causal component is manipulated.

*Discrimination:* articulatory difficulty, lexical retrieval difficulty, increased duration, and listener burden should therefore be manipulated separately rather than collapsed into a generic effort cost.

**7. Uncertainty about one latent factor**

*Location:* EFE information gain.

*Immediate prediction:* factor-targeted clarification.

*Diagnostic:* question form should reveal whether reference, force, QUD, or reliability is uncertain.

**8. Unexpected partner response**

*Location:* whichever representation best explains the discrepancy in $z_t^B$ and $\theta_{A\leftarrow B}$.

*Immediate prediction:* surprising response can reorganize the next policy by changing the inferred current partner state, the interaction-level context, or a persistent partner-model hypothesis.

*Carryover:* a current-state update should remain relatively local to the episode, an interaction-level update should track the ongoing discourse sequence or task, and a partner-model update should preferentially persist for the same partner.

**9. Persistent misunderstanding**

*Location:* divergent agent-specific posteriors under nondiagnostic observations.

*Prediction:* a response can be well predicted under both agents' current generative models while leaving their referential beliefs sharply divergent. Diagnostic evidence that discriminates the competing referents should reduce the divergence, whereas nondiagnostic uptake may leave it unchanged. Conversational fluency or locally well-predicted responses therefore do not by themselves establish shared reference.

**10. Conversation stopping.**

*Location:* comparison between the best continuation policy and the current temporal opportunity rate under the restricted rate formulation of Section 4.8.

*Prediction:* an agent may stop even when a continuation policy still has lower EFE than stopping, provided that its prospective EFE improvement per unit time falls below $\mu_t(c_t)$.

*Diagnostic*: termination therefore need not coincide with zero residual uncertainty or with the disappearance of all prospective benefit from further clarification.

### 6.2. Model-comparison requirements

The coupled model is more expressive than a standard one-shot RSA model, so model comparison must establish that its additional structure improves prediction rather than merely increasing flexibility. Empirical applications should therefore combine held-out predictive performance, posterior predictive checks, parameter-recovery simulations, and intervention-based validation. Relevant comparison models include at least: (i) a static RSA baseline; (ii) RSA with a flexible trial-level cost; (iii) RSA with partner-specific learning; (iv) an amortized speaker model; (v) a model with a single unconstrained dynamic policy prior $E_t$ ; (vi) an active-inference model in which temporal effects are represented within temporally extended EFE, for example through time-indexed preferences, discounting, or horizon structure; and (vii) the proposed coupled model with separately specified partner-state, partner-model, contextual temporal, and habitual-learning components. For comparison, we refer the reader to Hawkins et al. (2023).

Identifiability should be established before assigning psychological interpretations to fitted parameters. In particular, a single contemporaneous policy distribution cannot generally identify $E_t^{\mathrm{h}ab}$ and $\Phi_t^{\mathrm{time}}$ separately, because only their normalized product contributes to the current policy prior. Their distinction therefore requires temporal interventions and longitudinal observations: a one-off change in temporal context should alter choice immediately but disappear when that context is removed, whereas repeated realized selection may alter $E_t^{\mathrm{h}ab}$ and leave residual carryover. Likewise, the same unexpected response may be explained by a change in current partner state, interaction-level context, persistent partner model, preference, precision, or policy history. Experimental designs should therefore include selective interventions and transfer tests that discriminate these alternatives. Simulation-based recovery should specifically test whether temporal sensitivity $\mu_t(c_t)$, habit acquisition $\kappa_E$ , habitual forgetting $\omega$, policy precision $\gamma_t$, and persistent partner-model parameters can be estimated separately under the proposed design.

### 6.3. Relation to Relevance Theory and efficiency accounts

Our model also creates a disciplined comparison with relevance-theoretic and efficiency approaches. A policy may be attractive because it produces valued downstream consequences, resolves consequential uncertainty, can be processed by the current listener, or reaches a useful result rapidly. These influences resemble the intuitive trade-off between cognitive effects and effort, but they occupy different locations in the present architecture. Relevance-theoretic claims should therefore be translated into explicit hypotheses about which predicted outcomes are valued, which uncertainty matters, which contextual assumptions define the hypothesis space, and which causal process gives rise to the relevant form of effort. ‘Processing effort’ is not assigned a single active-inference variable by definition.

The time-efficiency calculation in Section 5 makes an additional distinction. The target-evidence rate $R_J = \Delta \mathrm{J}/\mathrm{T}$ is a descriptive diagnostic of how effectively a policy changes a designated target belief per unit time; it is neither the general policy-selection score nor an input inserted directly into $\Phi_t^{\mathrm{time}}$. The contextual temporal factor instead represents the current opportunity price of taking longer. Moreover, time efficiency is partner-relative: the same shorthand may provide high target-evidence gain per unit time for a shorthand-familiar partner and low gain per unit time for an unfamiliar one. Brevity is therefore not intrinsically efficient. Uniform-information-density and efficient-communication accounts motivate important predictions about the distribution of information over time, but further intervention is required to determine whether an observed adaptation arises from listener capacity, channel noise, production difficulty, temporal opportunity cost, or learned convention (Aylett & Turk, 2004; Gibson et al., 2019; Jaeger, 2010; Pate & Goldwater, 2015).

### 6.4. Relation to bounded and amortized pragmatics

The distinction between the exact posterior implied by a generative model and the tractable posterior represented by an agent should remain central. Natural discourse combines referents, lexical senses, force, QUD, commitments, social stances, partner models, and interaction histories too rapidly for exhaustive enumeration. Approximation may be achieved through restricted support, factorization, variational message passing, sampling-based inference, or a learned recognition mapping. Amortized RSA is therefore not an opponent of the present model but one candidate strategy for approximating part of the required computation. A learned context-to-utterance mapping can efficiently approximate speaker choice in familiar task distributions, whereas the coupled architecture additionally specifies which representations are revised after an unexpected response and how those revisions alter subsequent policy evaluation (see also Franke & Degen, 2016).

A learned amortized mapping should not, however, be identified with the habitual policy prior $E_t^{\mathrm{h}ab}$. The habitual term is a component of the generative decision model: it contributes prior support to policies and is updated from realized policy history. An amortized mapping is instead a computational mechanism for approximating or initializing an otherwise costly inference or policy calculation. The two can interact, for example, an amortized policy network may learn to approximate a posterior that itself contains habitual prior support, but they make different claims about where past experience enters the computation.

Nor must amortization be restricted to a direct context-to-utterance mapping. In the coupled architecture, learned approximations could operate at several levels: recognition of $z_t^B$ . inference over persistent partner-model variables $\theta_{A\leftarrow B}$, prediction of a partner's response, construction of counterfactual posteriors, or approximation of the policy posterior itself. This provides a natural route for bounding nested social inference without requiring exhaustive recursive computation on every turn. The empirical question is then not whether pragmatic reasoning is exact or amortized in general, but which parts of the coupled computation are approximated, and under what histories those approximations generalize or fail.

## 7. Conclusion

> "The universal cognitive tendency to maximise relevance makes it possible, at least to some extent, to predict and manipulate the mental states of others".
>
> — Wilson and Sperber (2004)

The models that currently dominate probabilistic pragmatics are theories of interpretation: given an utterance, infer what was meant. We have argued that this is only half of the problem. Once the speaker's choice rule is placed inside the listener's generative model, interpretation and production become one computation viewed from two angles, and every conversational move is simultaneously evidence about an interlocutor and an intervention on the future course of discourse. Pragmatics does not end when a listener has formed a posterior over what a speaker intended. That posterior changes the expected consequences of the next move, and a clarification question, acknowledgment, reformulation, indirect request, gesture, or closing is selected partly because the preceding action has changed the agent's model of its partner and of the interaction. The central prediction that follows is that identical utterance choices can conceal different computational causes and histories, which later turns, new partners, and targeted interventions will reveal.

We have proposed a coupled active-inference architecture for making this sequential dependence explicit. The likelihood of a partner's observed action is induced by the policy distribution attributed to that partner; current-state inference is separated from learning of partner-specific parameters; and prospective EFE is separated from policy priors. Within the prospective score, outcome-valued communicative consequences enter factorized PV, expected uncertainty resolution enters IG, and listener or channel constraints alter the relevant likelihood. Policy selection retains a single normalized prior $E_t(\pi|c_t)$, parameterized in the present model as the normalized product of a learned habitual baseline $E_t^{\mathrm{h}ab}(\pi)$ and a positive context-sensitive temporal factor $\Phi_t^{\mathrm{time}}(\pi|c_t)$. The temporal factor changes with the current opportunity structure, whereas the habitual component changes through accumulated realized policy selections. Temporal pressure can therefore leave a longer-term trace indirectly by changing which policies are repeatedly selected.

Under a restricted one-step construction, these terms can be mapped to an RSA utility containing informativeness, temporal cost, and additional values, while terms common across alternatives cancel under normalization. The familiar RSA speaker and listener distributions are thereby recovered as a limiting case. This static equivalence does not uniquely identify the causal allocation: at a single time and context, only the combined contribution of $E_t^{\mathrm{h}ab}$and $\Phi_t^{\mathrm{time}}$ to the normalized policy prior is behaviorally visible. The proposed decomposition instead derives empirical content from distinct intervention, learning, persistence, and transfer signatures.

For linguistics, the important consequence is a division of explanatory labor. Grammar and conventional meaning constrain which structured forms and interpretations are available. Dynamic pragmatics constrains QUD, commitments, and context change. Research on dialogue and conversational interaction constrains the sequential organization of grounding, repair, and closing. The coupled model proposes how uncertain beliefs about these objects, together with partner-specific expectations and socially consequential outcomes, may be updated and converted into later action. It therefore occupies the interface between linguistic representation and communicative control rather than reducing one to the other.

Our account is also *internalist* in a specific sense. The interlocutor never enters the model as a mind-external object; it enters as a component of the agent's own generative model, and the evidence that accumulates over a conversation is evidence for that component. Under the free-energy principle an agent's actions supply evidence for its own model, and since the coupled model contains a partner the same holds of the partner. Speaking is therefore not only an attempt to change the hearer but a way of sampling evidence *for or against one's model of the hearer*, and expected information gain about the partner enters the value of an utterance directly (Eq. 52). A predicted reply raises the evidence for the interlocutor as modelled; an unpredicted one lowers it. As we show, two agents can each accumulate evidence for a partner model that is wrong about the other, so what dialogue certifies is the internal representation of the other, not the other. More generally, we might say that we speak not only to communicate with others, but to test the very models through which others become intelligible to us: every utterance is therefore at once an intervention in another mind and an epistemic probe into our representation of it.

Our proposal remains deliberately vulnerable to a range of evidence types. Richer latent structure creates risks of non-identifiability and post hoc explanation, and multi-agent recursion remains tractable only through finite depth, factorization, support restriction, or learned approximation. Coupling does not guarantee truth, shared reference, or belief convergence: two agents may each assign low VFE to a locally coherent exchange while maintaining sharply divergent referential beliefs when the available observations are nondiagnostic. Likewise, discourse termination need not imply complete certainty. Under the restricted rate formulation developed above, an agent may stop even when a continuation policy would still improve prospective EFE, if that improvement is too small relative to the policy's duration and the current temporal opportunity rate. The numerical examples demonstrate these possibilities rather than establishing them as empirical effects.

Future work must therefore compare rival causal allocations against behavioral data, establish parameter recovery in simulation, and test selective interventions involving transient versus repeated temporal pressure, shorthand-familiar versus shorthand-unfamiliar partners, local processing load, social risk, unexpected feedback, transfer across partners and tasks, and the sequential uptake of repair and closing. The central empirical contribution of the framework is therefore to generate testable differences in immediate adaptation, persistence, transfer, misunderstanding, and stopping that distinguish the proposed architecture from simpler alternatives.

The apparent opposition between "language for thought" and "language for communication" viewpoints may therefore be drawn at the wrong explanatory level. Our model does not require language to be a "communication system"; it requires communication to be a system that can exploit linguistic structure.

More broadly, our framework complicates recent claims that language is primarily a tool for communication rather than thought (Fedorenko et al., 2024). Evidence that linguistic behavior is shaped by pressures for efficient information transfer does not establish that the computational system generating linguistic structure is itself a communication system. In our account, grammar and compositional semantics first make structured objects available; only downstream are those objects recruited by a distinct architecture of partner modelling, belief updating, policy evaluation, repair and action. Linguistic communication therefore presupposes structured language rather than deriving it, and communicative efficiency may characterize the use and externalization of linguistic objects without explaining their internal organization. More radically, even overt speech is not purely outward-facing: an utterance is also an epistemic action through which the speaker samples evidence for their own model of another mind. Language can therefore be used for communication without being reducible to communication. Showing that linguistic objects are exquisitely useful for communicative interaction is not the same thing as showing that language is for communication.

More generally, communicative efficiency cannot by itself identify communicative function. A central result of the present framework is that observationally equivalent utterance choices may arise from different causal allocations and histories. The same principle applies to linguistic form: an efficient surface pattern does not reveal whether its source lies in grammatical computation, externalization, processing constraints, cultural transmission, partner adaptation, or policy selection. Inferring the function of language from the efficiency of its outputs therefore risks mistaking a property of behavior for a property of the mechanism that makes that behavior possible.

The broader implication of our model is that communication is not simply the recovery of meaning, but a temporally extended process of acting on, learning about, and revising models of other agents. This develops the Gricean/RSA view of pragmatic inference by bringing it into contact with dynamic accounts of context and commitment, interactional work on grounding, repair, and partner adaptation, and active-inference accounts of interpersonal coupling. Linguistic structure constrains the objects available for interpretation, and dialogue determines how those objects become evidence and action. The outcome is a common causal architecture for interpretation, production, learning, and repair – one in which successful coordination need not imply perfect mental alignment, and every utterance both changes another mind and tests our model of it.

## Appendix A: Active-inference definitions and derivations

This appendix retains the full definitions and derivations underlying the compact quantities introduced in Section 2.

The free-energy principle, and its active inference process-theory implementations, provide a general framework for describing inference, learning, and action under uncertainty (Friston, 2010; Parr et al., 2022). The formalism begins with a generative model: a probabilistic model of how hidden states give rise to observable outcomes. Variational free energy (VFE) is then used to infer the hidden states responsible for observations that have already occurred, whereas expected free energy (EFE) is used to evaluate the possible future consequences of alternative policies.

These two quantities must be kept distinct. VFE, $F$, is retrospective in the limited sense that it is evaluated using current or past observations. EFE, $G$, is prospective and *counterfactual*: it is evaluated over outcomes that have not yet occurred, conditional on policies that may or may not be pursued. Both are defined relative to an agent's generative model, but they perform different formal roles (Parr et al., 2022).

Throughout this section, we will use $P_\theta$ to denote probabilities specified by a generative model with parameters $\theta$, whereas $q$ denotes an agent's current inferential or predictive beliefs. Unless otherwise stated, variables are treated as discrete and logarithms are natural logarithms. For continuous variables, the corresponding sums are replaced by integrals.

### A.1. Generative models and inverse inference

Let $o$ denote an observable outcome and $s$ a hidden state that could have generated it. A generative model specifies a joint probability distribution over these variables:

$$P_\theta(o, s) = P_\theta(o \mid s) P_\theta(s). \qquad \text{(A1)}$$

The two factors in Appendix Equation (A1) have different roles. The prior,

$$P_\theta(s), \qquad \text{(A2)}$$

expresses the probability assigned to each hidden state before the current observation is taken into account. The likelihood,

$$P_\theta(o \mid s), \qquad \text{(A3)}$$

expresses how probable each possible observation would be if a particular hidden state obtained.

The term *generative* refers to the direction represented by this factorization. Given the model, one could generate a sample by first drawing a hidden state from $P_\theta(s)$ and then drawing an observation from $P_\theta(o \mid s)$. This does not imply that the hidden state is directly available to the agent. On the contrary, only $o$ is observed; $s$ must be inferred.

Marginalizing the hidden state gives the model evidence:

$$P_\theta(o) = \sum_s P_\theta\,(o \mid s) P_\theta(s). \qquad \text{(A4)}$$

The evidence is the total probability assigned to $o$, after considering all hidden states that could have generated it. It therefore evaluates the observation under the model as a whole, rather than under one particular state hypothesis.

The surprisal of the observation is defined as

$$\mathcal{S}_\theta(o) = -\ln P_\theta(o). \qquad \text{(A5)}$$

An observation to which the model assigns high probability has low surprisal, whereas an observation assigned low probability has high surprisal. Surprisal is model-relative: the same observation can have different surprisal under different priors, likelihoods, or parameter settings. It is also an information-theoretic quantity and need not correspond to conscious surprise, novelty, or an overt startle response.

Inference proceeds in the reverse direction to generation. Given an observation, the agent seeks a distribution over the hidden states that might have produced it. Bayes' rule gives

$$P_\theta(s \mid o) = \frac{P_\theta(o \mid s) P_\theta(s)}{P_\theta(o)} = \frac{P_\theta(o \mid s) P_\theta(s)}{\sum_{s'} P_\theta\,(o \mid s') P_\theta(s')}. \qquad \text{(A6)}$$

The posterior $P_\theta(s \mid o)$ combines prior belief with the evidence provided by the observation. The denominator is not an additional hypothesis about the state, but is rather a normalization term ensuring that $\sum_s P_\theta\,(s \mid o) = 1$.

For a small discrete model, Appendix Equation (A6) can be evaluated exactly. We calculate the unnormalized posterior score

$$P_\theta(o \mid s) P_\theta(s) \qquad \text{(A7)}$$

for each state and divide each score by their sum. In this setting, no iterative variational procedure is mathematically required.

Exact inference becomes difficult when the hidden state is high-dimensional, when it contains many interacting factors, when the model is temporally or hierarchically deep, or when the posterior has a complex form. The difficulty arises primarily from the marginalization in Appendix Equation (A4): computing $P_\theta(o)$ may require summing or integrating over an extremely large state space. *Variational inference* introduces a tractable distribution,

$$q_\phi(s) \approx P_\theta(s \mid o), \qquad \text{(A8)}$$

where $\phi$ denotes the parameters of the approximate posterior. For notational economy, we generally write $q(s)$ and leave $\phi$ implicit. The distribution may be selected from a restricted variational family, for example a mean-field family,

$$q(s) = \prod_{i=1}^{N} q_i\,(s_i), \qquad \text{(A9)}$$

or it may be represented by a parameterized recognition model. These restrictions make inference tractable but may prevent $q$ from exactly representing the model posterior.

Three distributions should therefore be distinguished:

1. $P_\theta(s)$ is the prior specified by the generative model;
2. $P_\theta(s \mid o)$ is the exact posterior mathematically implied by that model;
3. $q(s)$ is the posterior belief actually represented or computed by the agent.

The exact posterior is 'exact' only relative to $P_\theta$; it is not guaranteed to recover an objective state of the external world if the model itself is misspecified. Likewise, $q$ need not always be approximate. In a small model, it may be set equal to the exactly normalized posterior. The notation $q$ is retained because it identifies the distribution over which variational optimization is performed and generalizes to cases in which exact inference is unavailable.

The distinction between $q$ and $\theta$ will be central below. Inference changes beliefs about the current hidden state, represented by $q$. Learning changes the relatively persistent parameters $\theta$ that determine the prior, likelihood, and, in temporal models, state transitions. Updating $q$ is therefore not the same operation as updating the model that generates $q$.

### A.2. Variational free energy

VFE is defined as

$$F[o, q, \theta] = \mathbb{E}_{q(s)}[\ln q(s) - \ln P_\theta(o, s)]. \qquad \text{(A10)}$$

For fixed $o$ and $\theta$, $F$ is a functional of the entire distribution $q$. Different candidate distributions over $s$ generally produce different values of $F$.

Appendix Equation (A10) can first be written as an energy-entropy decomposition:

$$F[o,q,\theta] = \mathbb{E}_q[-\ln P_\theta(o,s)] - \mathrm{H}[q(s)], \qquad \mathrm{H}[q(s)] = -\mathbb{E}_q \ln q(s). \quad \text{(A11)}$$

The first term assigns a high value when $q$ places probability mass on state-observation combinations that are improbable under the generative model. The second term is the entropy of $q$. The two terms cannot be optimized independently: a diffuse distribution has high entropy, but it will not have low free energy if it assigns probability to states that poorly explain the observation.

Two additional decompositions make the inferential role of $F$ more explicit.

### A.2.1. Posterior-divergence form

The joint distribution also admits the factorization

$$P_\theta(o,s) = P_\theta(s \mid o) P_\theta(o). \quad \text{(A12)}$$

Substituting Appendix Equation (A12) into Appendix Equation (A10) gives

$$\begin{aligned} F[o,q,\theta] &= \mathbb{E}_q[\ln q(s) - \ln P_\theta(s \mid o) - \ln P_\theta(o)] \\ &= \mathbb{E}_q[\ln \frac{q(s)}{P_\theta(s \mid o)}] - \ln P_\theta(o) \quad \text{(A13)} \\ &= D_{\mathrm{KL}}[q(s) \parallel P_\theta(s \mid o)] - \ln P_\theta(o). \end{aligned}$$

The first term is nonnegative:

$$D_{\mathrm{KL}}[q(s) \parallel P_\theta(s \mid o)] \geq 0. \quad \text{(A14)}$$

It follows that

$$F[o,q,\theta] \geq -\ln P_\theta(o). \quad \text{(A15)}$$

VFE is therefore an upper bound on the surprisal of the observation under the generative model (Parr et al., 2022).

For fixed $o$ and $\theta$, the surprisal term does not depend on $q$. Minimizing $F$ with respect to $q$ consequently minimizes the divergence between $q(s)$ and $P_\theta(s \mid o)$:

$$q^\star = \underset{q \in Q}{\mathrm{argmin}} F[o,q,\theta], \quad \text{(A16)}$$

where $Q$ denotes the selected variational family. If the family contains the exact posterior, then

$$q^\star(s) = P_\theta(s \mid o) \quad \text{and} \quad F[o,q^\star,\theta] = -\ln P_\theta(o). \quad \text{(A17)}$$

The bound is then tight. If the variational family is restricted, a residual gap remains:

$$\min_{q \in Q} F[o,q,\theta] = -\ln P_\theta(o) + \min_{q \in Q} D_{\mathrm{KL}}[q(s) \parallel P_\theta(s \mid o)]. \quad \text{(A18)}$$

This gap quantifies the inferential error induced by the approximation family and its optimization.

An important qualification follows. For fixed $o$ and $\theta$, changing $q$ does not change the model evidence $P_\theta(o)$, because the observation has already occurred and the model is fixed. Perceptual inference instead makes the free-energy bound tighter by bringing $q$ closer to the model posterior. Model evidence can be changed by learning $\theta$, while future observations can be changed by action.

### A.2.2. Inaccuracy-complexity form

Using the generative factorization in Appendix Equation (A1) gives

$$\begin{aligned} F[o,q,\theta] &= \mathbb{E}_q[\ln q(s) - \ln P_\theta(o \mid s) - \ln P_\theta(s)] \\ &= \underbrace{-\mathbb{E}_q \ln P_\theta(o \mid s)}_{\text{inaccuracy}} + \underbrace{D_{\mathrm{KL}}[q(s) \parallel P_\theta(s)]}_{\text{complexity}}. \end{aligned} \quad \text{(A19)}$$

Equivalently, if accuracy is defined as

$$\text{Accuracy} = \mathbb{E}_q \ln P_\theta(o \mid s), \quad \text{(A20)}$$

then

$$F = \text{Complexity} - \text{Accuracy}. \quad \text{(A21)}$$

The inaccuracy term is the expected negative log likelihood. It is low when $q$ assigns probability to states that predict the observation well. The complexity term measures the information-theoretic difference between the posterior belief $q(s)$ and the prior $P_\theta(s)$. It therefore quantifies how extensively beliefs must be revised in order to explain the observation.

This decomposition does not imply that an agent should avoid all belief updating. An informative observation may justify a substantial posterior departure from the prior. Such a departure increases complexity, but it can still decrease total free energy if the corresponding gain in accuracy is larger. Free-energy minimization regulates the trade-off between the two terms; it does not minimize either term in isolation.

Likewise, variational complexity should not automatically be equated with reaction time, subjective difficulty, conscious effort, neural firing rate, metabolic expenditure, or the number of computational operations (McCarty et al., 2023; Murphy, 2024, 2025, 2026; Murphy et al., 2022, 2024a, 2024b). It is specifically the KL divergence

$$D_{\mathrm{KL}}[q(s) \parallel P_\theta(s)]. \quad \text{(A22)}$$

Connecting this quantity to a psychological or neurobiological cost requires an additional process theory specifying how the relevant probability distributions are represented and updated.

VFE is also the negative evidence lower bound:

$$
\begin{aligned}
\mathcal{L}_{\mathrm{ELBO}}[q,\theta] &= \mathbb{E}_q[\ln P_\theta(o,s) - \ln q(s)], \\
F[o,q,\theta] &= -\mathcal{L}_{\mathrm{ELBO}}[q,\theta].
\end{aligned} \qquad \text{(A23)}
$$

Thus, minimizing VFE is equivalent to maximizing the ELBO. Under additional assumptions, such as Gaussian likelihoods and a Laplace approximation, free energy may reduce to a sum of precision-weighted prediction errors. In the general case, however, VFE should not be identified with raw prediction error alone: it includes both expected log likelihood and the divergence between posterior and prior beliefs.

### A.3. Inference, learning, and action as distinct updates

Inference, learning, and action can all be formulated in relation to free-energy minimization, but they optimize different variables. Their formal unification therefore does not make them the same operation.

#### A.3.1. Inference

At time $t$, holding the observation $o_t$ and parameters $\theta_t$ fixed, inference optimizes the state belief:

$$
q_t^\star = \underset{q_t \in Q}{\operatorname{argmin}} F_t[o_t, q_t, \theta_t]. \qquad \text{(A24)}
$$

If $q_t$ is parameterized by $\phi_t$, an iterative implementation may take the form

$$
\phi_t \leftarrow \phi_t - \eta_\phi \nabla_{\phi_t} F_t[o_t, q_{\phi_t}, \theta_t]. \qquad \text{(A25)}
$$

This iterative gradient flow is one possible process for finding the posterior. It is not part of the definition of Bayesian inference itself. In a small categorical model, Appendix Equation (A6) can instead be evaluated directly in a single normalized update. Iteration becomes necessary only when required by the representation or optimization method.

#### A.3.2. Learning

Learning changes the parameters that define the generative model. For a sequence of observations, point-estimate learning can be written as

$$
\theta^\star = \underset{\theta}{\operatorname{argmin}} \sum_{t=1}^{N} F_t\,[o_t, q_t, \theta]. \qquad \text{(A26)}
$$

A stochastic gradient update is

$$
\theta_{t+1} = \theta_t - \eta_\theta \nabla_\theta F_t[o_t, q_t, \theta_t]. \qquad \text{(A27)}
$$

Whereas inference changes the current distribution over hidden states, learning changes the mappings and regularities encoded by the model. For example, learning may alter the likelihood $P_\theta(o \mid s)$, the prior $P_\theta(s)$, or the transition probabilities of a temporal model.

Appendix Equations (A24) and (A27) can be implemented as alternating or simultaneous updates. Holding one variable fixed in the equations isolates the formal role of each optimization; it does not require that biological or artificial systems complete one process before beginning the other. In a fully Bayesian treatment, parameters can themselves be treated as hidden variables with a posterior $q(\theta)$. The point-estimate notation above is retained because it makes the distinction between state inference and parameter learning transparent.

Inference is often associated with relatively fast changes in state beliefs, while learning is associated with slower changes in parameters. This difference in timescale is common and useful, but the fundamental distinction is the variable being updated, not an absolute temporal boundary.

### A.3.3. Action

Action changes the observations available to the agent by changing its relation to the environment. In continuous formulations, an action variable $a$ may be updated through the dependence of sensory observations on action:

$$\dot{a} \propto -\frac{\partial F[o(a), q, \theta]}{\partial a}. \qquad \text{(A28)}$$

Inference changes beliefs so that they better explain sampled observations; action changes sampling so that observations become more consistent with the agent's predictions and preferences – thus providing evidence for one's own (model's) existence.

Appendix Equation (A28) concerns an action that can be adjusted with respect to currently sampled sensory data. Planning requires an additional prospective calculation. Before a future observation is available, an agent must compare the observations expected under different possible sequences of actions. In discrete active inference, these action sequences are policies, and their prospective evaluation is performed using EFE.

## A.4. A discrete POMDP generative model

To evaluate policies, the static model in Appendix Equation (A1) must be extended through time. A partially observable Markov decision process (POMDP) specifies how hidden states evolve, how they generate observations, and how their transitions depend on actions.

Let $s_\tau$ be the hidden state and $o_\tau$ the observation at time $\tau$. Let $u_\tau$ denote an action, and let

$$\pi = (u_t, u_{t+1}, \dots, u_{T-1}) \qquad \text{(A29)}$$

denote a policy extending from the current time $t$ to a finite horizon $T$. Conditional on a policy, a basic POMDP generative model factorizes as

$$P_\theta(o_{1:T}, s_{1:T} \mid \pi) = P_\theta(s_1) \prod_{\tau=1}^{T} P_\theta\,(o_\tau \mid s_\tau) \prod_{\tau=1}^{T-1} P_\theta\,(s_{\tau+1} \mid s_\tau, u_\tau). \qquad \text{(A30)}$$

The first factor is the prior over the initial state. The first product contains the likelihood mappings from hidden states to observations. The second product contains action-dependent state transitions.

For categorical variables, these components are commonly represented by the arrays $A$, $B$, and $D$:

$$A_{ij} = P_\theta(o_\tau = i \mid s_\tau = j), \qquad \text{(A31)}$$
$$B^u_{kj} = P_\theta(s_{\tau+1} = k \mid s_\tau = j, u_\tau = u), \qquad \text{(A32)}$$
$$D_j = P_\theta(s_1 = j). \qquad \text{(A33)}$$

The model is *partially observable* because the hidden state is not observed directly. Different states may generate the same observation, and a single state may generate different observations probabilistically. The model is a *decision process* because actions influence state transitions and therefore influence future observations.

The generative model in Appendix Equation (A30) represents the agent's beliefs about state transitions and observation generation. It should be distinguished from the generative process in the external environment. The two may coincide in an idealized model, but an agent can also infer and act under an inaccurate generative model.

Suppose that inference over the current state has produced $q_t(s_t)$. For a candidate policy, future state beliefs are predicted recursively:

$$q(s_t \mid \pi) = q_t(s_t),$$
$$q(s_{\tau+1} \mid \pi) = \sum_{s_\tau} P_\theta\,(s_{\tau+1} \mid s_\tau, u_\tau) q(s_\tau \mid \pi), \qquad \tau \geq t. \qquad \text{(A34)}$$

These are predictive beliefs: they describe what state the agent expects to occupy if policy $\pi$ is pursued. The predicted outcome distribution is obtained by passing the predictive state belief through the likelihood:

$$q(o_\tau \mid \pi) = \sum_{s_\tau} P_\theta\,(o_\tau \mid s_\tau) q(s_\tau \mid \pi). \qquad \text{(A35)}$$

The corresponding predictive joint distribution is

$$q(o_\tau, s_\tau \mid \pi) = P_\theta(o_\tau \mid s_\tau) q(s_\tau \mid \pi). \qquad \text{(A36)}$$

The use of $q$ here is deliberate. These distributions do not describe observed future events. They are the agent's counterfactual predictions under a candidate policy, constructed using the likelihood supplied by $P_\theta$. For each possible future observation, the agent can also construct the posterior that it would hold if that observation were encountered:

$$q(s_\tau \mid o_\tau, \pi) = \frac{P_\theta(o_\tau \mid s_\tau) q(s_\tau \mid \pi)}{q(o_\tau \mid \pi)}. \qquad \text{(A37)}$$

Appendix Equation (A37) is a counterfactual Bayesian update. Before choosing a policy, the agent considers each possible future observation, the probability of receiving it, and the posterior belief that would follow from it. To make the temporal roles explicit, the main text denotes the predictive distribution in Appendix Equation (A34) by $q_\tau^-(s_\tau \mid \pi)$ and the counterfactual posterior in Appendix Equation (A37) by $q_\tau^+(s_\tau \mid o_\tau, \pi)$. The superscripts indicate before and after incorporation of the possible observation; they do not indicate valence.

If the counterfactual update is intractable, $q_\tau^+$ is defined as the optimized variational posterior $\mathrm{argmin}_{q \in Q} F_\tau[q; o_\tau, \pi]$ and only approximates the exact model posterior. EFE then uses this inner inference result to evaluate how informative and preferred the possible future consequences of each policy would be. It should therefore be distinguished from a simple expectation of the ordinary VFE under future observations.

Active-inference models additionally specify prior preferences over future outcomes:

$$P_\tau^\star(o_\tau), \qquad \sum_{o_\tau} P_\tau^\star(o_\tau) = 1. \qquad \text{(A38)}$$

The preference distribution is distinct from both the likelihood and the predictive outcome distribution:

1. $P_\theta(o_\tau \mid s_\tau)$ describes which outcomes a state is expected to generate;
2. $q(o_\tau \mid \pi)$ describes which outcomes a policy is predicted to produce;
3. $P_\tau^\star(o_\tau)$ describes which outcomes are preferred or expected a priori.

Prior preferences provide the model with a probabilistic representation of goal states or characteristic outcomes. They need not describe how frequently those outcomes currently occur in the environment.

### A.5. Expected free energy and policy selection

VFE evaluates beliefs in relation to observations that are already available. EFE evaluates a policy using predicted future observations. For each future time $\tau > t$, it combines two properties of a policy: whether the policy is expected to reduce uncertainty and whether it is expected to realize preferred outcomes (Parr et al., 2022).

#### A.5.1. Expected information gain

The expected information gain of policy $\pi$ at time $\tau$ is

$$\mathrm{IG}_\tau(\pi) = \mathbb{E}_{q(o_\tau \mid \pi)} D_{\mathrm{KL}}[q(s_\tau \mid o_\tau, \pi) \;\|\; q(s_\tau \mid \pi)]. \qquad \text{(A39)}$$

For a particular possible observation $o_\tau$, the inner KL divergence compares two distributions:

1. $q(s_\tau \mid \pi)$, the belief about the future state before receiving the future observation;
2. $q(s_\tau \mid o_\tau, \pi)$, the belief that would follow after receiving that observation.

The outer expectation averages this divergence over all possible future observations, weighted by their predicted probabilities $q(o_\tau \mid \pi)$. Thus, information gain asks how much the agent expects its state beliefs to change as a consequence of observing the outcomes generated by a policy. Because KL divergence is nonnegative,

$$\mathrm{IG}_\tau(\pi) \geq 0, \qquad \text{(A40)}$$

It is zero when every possible future observation leaves the state belief unchanged:

$$q(s_\tau \mid o_\tau, \pi) = q(s_\tau \mid \pi). \qquad \text{(A41)}$$

It is large when predicted observations are expected to discriminate among currently plausible hidden states. Under the predictive joint distribution (Appendix Equation 36), expected information gain is the mutual information between future states and future observations:

$$\mathrm{IG}_\tau(\pi) = I_q(s_\tau; o_\tau \mid \pi). \qquad \text{(A42)}$$

It has two equivalent entropy decompositions:

$$\begin{aligned} \mathrm{IG}_\tau(\pi) &= \mathrm{H}[q(s_\tau \mid \pi)] - \mathbb{E}_{q(o_\tau \mid \pi)} \mathrm{H}[q(s_\tau \mid o_\tau, \pi)] && \text{(A43)} \\ &= \mathrm{H}[q(o_\tau \mid \pi)] - \mathbb{E}_{q(s_\tau \mid \pi)} \mathrm{H}[P_\theta(o_\tau \mid s_\tau)]. && \text{(A44)} \end{aligned}$$

Appendix Equation (A43) says that information gain is the expected reduction in uncertainty about hidden states. Appendix Equation (A44) says that informative policies produce outcomes with substantial predictive variation, but only when that variation is diagnostic of the hidden state rather than being generated by observation noise.

### A.5.2. Pragmatic value

We define *pragmatic value*[3], hereafter $PV$, as:

$$PV_\tau(\pi) = \mathbb{E}_{q(o_\tau \mid \pi)} \ln P_\tau^\star(o_\tau). \qquad \text{(A45)}$$

[3] A terminological clarification is warranted here. Throughout this paper, the term pragmatic value has been used in the broad sense standard in the active-inference literature, where it refers to the value of predicted outcomes under an agent's prior preferences. It should therefore not be understood as synonymous with pragmatic in the narrower linguistic sense of pragmatics as the study of context-dependent meaning and communicative inference. Whenever we use the expression pragmatic value, we refer specifically to the active-inference quantity PV defined in terms of prior preferences.

For discrete outcomes,

$$PV_\tau(\pi) = \sum_{o_\tau} q\,(o_\tau \mid \pi) \ln P_\tau^\star(o_\tau). \qquad \text{(A46)}$$

This quantity evaluates predicted outcomes under the agent's prior preferences. Since probabilities are no greater than one, $\ln P_\tau^\star(o_\tau) \leq 0$, and $PV$ is generally nonpositive. A less negative, and therefore larger, value indicates that the policy is expected to produce more preferred outcomes. If prior preferences are uniform, then

$$P_\tau^\star(o_\tau) = \frac{1}{|\mathcal{O}|} \qquad \text{(A47)}$$

for every outcome. In that case, $PV$ is identical across policies and does not influence their relative ranking. Policy selection is then driven by epistemic differences alone.

#### A.5.3. The epistemic-preference decomposition

The expected free energy at time $\tau$ is

$$G_\tau(\pi) = -\mathrm{IG}_\tau(\pi) - PV_\tau(\pi). \qquad \text{(A48)}$$

For a finite policy horizon,

$$G_t(\pi) = \sum_{\tau=t+1}^{T} G_\tau\,(\pi). \qquad \text{(A49)}$$

The signs in Appendix Equation (A48) are important. Information gain is nonnegative and appears with a negative sign. Policies expected to produce large belief updates therefore have lower expected free energy. $PV$ is generally nonpositive; subtracting it penalizes policies whose predicted outcomes have low prior preference. Consequently, minimizing $G$ favors policies that are both informative and consistent with preferred outcomes.

The two terms are not independently appended exploration and reward bonuses. They arise as complementary decompositions of the same prospective functional. This can be seen by deriving the risk-ambiguity form. From Appendix Equations (A35) and (A37),

$$\frac{q(s_\tau \mid o_\tau, \pi)}{q(s_\tau \mid \pi)} = \frac{P_\theta(o_\tau \mid s_\tau)}{q(o_\tau \mid \pi)}. \qquad \text{(A50)}$$

Taking logarithms gives

$$\ln \frac{q(s_\tau \mid o_\tau, \pi)}{q(s_\tau \mid \pi)} = \ln P_\theta(o_\tau \mid s_\tau) - \ln q(o_\tau \mid \pi). \qquad \text{(A51)}$$

Substituting this relation into Appendix Equation (A48) yields

$$
\begin{aligned}
G_\tau(\pi) &= -\mathbb{E}_{q(o_\tau,s_\tau \mid \pi)} \ln \frac{q(s_\tau \mid o_\tau,\pi)}{q(s_\tau \mid \pi)} - \mathbb{E}_{q(o_\tau \mid \pi)} \ln P_\tau^\star(o_\tau) \\
&= \mathbb{E}_{q(o_\tau,s_\tau \mid \pi)}[\ln q(o_\tau \mid \pi) - \ln P_\theta(o_\tau \mid s_\tau) - \ln P_\tau^\star(o_\tau)] \qquad \text{(A52)} \\
&= \underbrace{D_{\mathrm{KL}}[q(o_\tau \mid \pi) \parallel P_\tau^\star(o_\tau)]}_{\text{risk}} + \underbrace{\mathbb{E}_{q(s_\tau \mid \pi)} \mathrm{H}[P_\theta(o_\tau \mid s_\tau)]}_{\text{expected ambiguity}}.
\end{aligned}
$$

Thus, under the predictive factorization specified above,

$$
\boxed{-\mathrm{IG}_\tau(\pi) - PV_\tau(\pi) = \mathrm{Risk}_\tau(\pi) + \mathrm{Ambiguity}_\tau(\pi)}. \qquad \text{(A53)}
$$

Risk is

$$
\mathrm{Risk}_\tau(\pi) = D_{\mathrm{KL}}[q(o_\tau \mid \pi) \parallel P_\tau^\star(o_\tau)]. \qquad \text{(A54)}
$$

This measures the discrepancy between the distribution of outcomes predicted under the policy and the distribution of outcomes preferred by the agent. Risk in this technical sense is not identical to physical danger or conscious concern. A deterministic policy can have high risk if its predicted outcome is strongly dispreferred. Conversely, a probabilistic policy has low risk when its distribution of possible outcomes closely matches the prior-preference distribution.

Expected ambiguity is

$$
\mathrm{Ambiguity}_\tau(\pi) = \mathbb{E}_{q(s_\tau \mid \pi)} \mathrm{H}[P_\theta(o_\tau \mid s_\tau)]. \qquad \text{(A55)}
$$

For a particular state,

$$
\mathrm{H}[P_\theta(o_\tau \mid s_\tau)] = -\sum_{o_\tau} P_\theta(o_\tau \mid s_\tau) \ln P_\theta(o_\tau \mid s_\tau). \qquad \text{(A56)}
$$

Ambiguity therefore concerns uncertainty in the state-to-outcome mapping. It asks how uncertain the observation would remain even if the hidden state were known. A deterministic likelihood has zero ambiguity. A likelihood in which the same state generates several observations with similar probability has high ambiguity.

Ambiguity should be distinguished from uncertainty about the hidden state. The entropy

$$
\mathrm{H}[q(s_\tau \mid \pi)] \qquad \text{(A57)}
$$

measures uncertainty over which state will obtain. By contrast, Appendix Equation (A55) measures uncertainty over which observation a known state will generate. Additional observations may resolve state uncertainty when the likelihood is diagnostic. They cannot resolve irreducible ambiguity when observations are statistically independent of the state.

Appendix Equation (A44) makes this distinction explicit:

$$
\mathrm{IG}_\tau(\pi) = \underbrace{\mathrm{H}[q(o_\tau \mid \pi)]}_{\text{predictive outcome uncertainty}} - \underbrace{\mathbb{E}_{q(s_\tau \mid \pi)} \mathrm{H}[P_\theta(o_\tau \mid s_\tau)]}_{\text{expected ambiguity}}. \qquad \text{(A58)}
$$

A high-entropy future is not necessarily informative. Random observations may be highly unpredictable while providing no evidence about hidden states. Information gain is high only when there is uncertainty about what will be observed and the eventual observation is expected to discriminate among state hypotheses.

### A.5.4 Why complexity in VFE and information gain in EFE have opposite signs

The complexity term in VFE is

$$+D_{\mathrm{KL}}[q(s \mid o) \parallel P_\theta(s)]. \qquad \text{(A59)}$$

It measures the belief revision required to explain an observation that has already been obtained. By contrast, EFE contains

$$-\mathbb{E}_{q(o_\tau \mid \pi)} D_{\mathrm{KL}}[q(s_\tau \mid o_\tau, \pi) \parallel q(s_\tau \mid \pi)]. \qquad \text{(A60)}$$

Before an observation has been obtained, the possibility of a belief-changing observation is valuable because it can resolve uncertainty. There is therefore no contradiction. VFE evaluates how economically current evidence can be explained, whereas EFE evaluates how useful future evidence is expected to be.

### A.5.5 Policy priors and policy posteriors

EFE supplies a prospective bias over policies. Let $E_t(\pi)$ be a normalized prior bias over policies:

$$E_t(\pi) \geq 0, \qquad \sum_{\pi} E_t(\pi) = 1. \qquad \text{(A61)}$$

In discrete active inference, $E_t(\pi)$ is commonly interpreted as a fixed or learned habit term. It represents a policy bias that is not itself computed from the current expected free energy. If no such bias is assumed, $E_t(\pi)$ can be uniform.

A prior over policies incorporating EFE can be written as

$$P_t(\pi) = \mathrm{softmax}[\ln E_t(\pi) - \gamma G_t(\pi)], \qquad \text{(A62)}$$

where $\gamma > 0$ is policy precision. A large $\gamma$ makes policy beliefs highly sensitive to differences in expected free energy; a small $\gamma$ produces a flatter distribution over policies.

Current and past observations may additionally provide policy-specific evidence. Let $F_{\pi,t}$ denote the VFE of the observations available at time $t$, evaluated under policy $\pi$. The posterior over policies is then

$$\begin{aligned} q_t(\pi) &\propto P_t(\pi)\exp[-F_{\pi,t}] \\ &= \mathrm{softmax}[\ln E_t(\pi) - F_{\pi,t} - \gamma G_t(\pi)]. \end{aligned} \qquad \text{(A63)}$$

The terms have distinct functions:

1. $E_t(\pi)$ is a prior or habitual bias over policies;
2. $F_{\pi,t}$ scores how well each policy accounts for observations that are already available;
3. $G_t(\pi)$ scores the predicted future consequences of each policy;
4. $q_t(\pi)$ is the resulting posterior belief over policies.

When the currently observed data are equally compatible with all prospective alternatives, $F_{\pi,t}$ is common across policies and cancels under normalization. Policy selection then reduces to

$$q_t(\pi) \propto E_t(\pi)\exp[-\gamma G_t(\pi)]. \qquad \text{(A64)}$$

This simplified expression should not be interpreted as identifying habitual bias with prospective evaluation. A frequently selected policy can receive a large $E_t(\pi)$ even if its current expected free energy is not minimal; conversely, a policy with weak habitual support can become probable when it has sufficiently low $G_t(\pi)$.

The framework developed in this section is domain-general. Its explanatory content depends on how a particular model instantiates the hidden states, observations, likelihoods, transitions, prior preferences, and admissible policies. EFE minimization alone does not determine these variables. A substantive application must specify them independently and show how their interactions generate the phenomenon to be explained.

## Declaration of Generative AI Use

The authors utilized ChatGPT with the GPT-5.6 Sol model (OpenAI) to assist with equation layout, translation, and language editing, as well as with checking the validity of algebraic derivations, notation and conditioning consistency, and the logical and formal connections among equations. All content was critically reviewed by the authors before incorporation into the manuscript.

## References

Aylett, M., & Turk, A. (2004). The smooth signal redundancy hypothesis: A functional explanation for relationships between redundancy, prosodic prominence, and duration in spontaneous speech. *Language and Speech*, *47*(1), 31–56.

Barr, D. J. (2008). Pragmatic expectations and linguistic evidence: Listeners anticipate but do not integrate common ground. *Cognition*, *109*(1), 18–40. https://doi.org/10.1016/j.cognition.2008.07.005

Bergen, L., Levy, R., & Goodman, N. D. (2016). Pragmatic reasoning through semantic inference. *Semantics and Pragmatics*, *9*, Article 20. https://doi.org/10.3765/sp.9.20

Bohn, M., & Frank, M. C. (2025). Pragmatics as social inference about intentional action. *Open Mind*, *9*, 290–304. https://doi.org/10.1162/opmi_a_00191

Brennan, S. E., & Clark, H. H. (1996). Conceptual pacts and lexical choice in conversation. *Journal of Experimental Psychology: Learning*, *Memory*, and Cognition, 22(6), 1482–1493. https://doi.org/10.1037/0278-7393.22.6.1482

Brown-Schmidt, S. (2012). Beyond common and privileged: Gradient representations of common ground in real-time language use. *Language and Cognitive Processes*, *27*(1), 62–89. https://doi.org/10.1080/01690965.2010.543363

Buz, E., Tanenhaus, M. K., & Jaeger, T. F. (2016). Dynamically adapted context-specific hyper-articulation: Feedback from interlocutors affects speakers' subsequent pronunciations. *Journal of memory and language*, *89*, 68–86. https://doi.org/10.1016/j.jml.2015.12.009

Clark, B. (2013). *Relevance Theory*. Cambridge University Press. https://doi.org/10.1017/cbo9781139034104

Clark, H. H., & Schaefer, E. F. (1989). Contributing to discourse. *Cognitive Science*, *13*(2), 259–294. https://doi.org/10.1207/s15516709cog1302_7

Constant, A., Ramstead, M. J. D., Veissière, S. P. L., Campbell, J. O., & Friston, K. J. (2018). A variational approach to niche construction. *Journal of the Royal Society Interface*, *15*(141), 20170685. https://doi.org/10.1098/rsif.2017.0685

Constant, A., Ramstead, M. J. D., Veissière, S. P. L., & Friston, K. J. (2019). Regimes of expectations: An active inference model of social conformity and human decision making. *Frontiers in psychology*, *10*, 679. https://doi.org/10.3389/fpsyg.2019.00679

Cummins, C. (2025). Noncooperative communication. *Annual Review of Linguistics*, *11*, 35–52. https://doi.org/10.1146/annurev-linguistics-011724-121451

Da Costa, L., Parr, T., Sajid, N., Veselic, S., Neacsu, V., & Friston, K. J. (2020). Active inference on discrete state-spaces: A synthesis. *Journal of Mathematical Psychology*, *99*, 102447. https://doi.org/10.1016/j.jmp.2020.102447

Degen, J. (2023). The rational speech act framework. *Annual Review of Linguistics*, *9*, 519–540. https://doi.org/10.1146/annurev-linguistics-031220-010811

Degen, J., Hawkins, R. D., Graf, C., Kreiss, E., & Goodman, N. D. (2020). When redundancy is useful: A Bayesian approach to "overinformative" referring expressions. *Psychological Review*, *127*(4), 591–621. https://doi.org/10.1037/rev0000186

Dingemanse, M., Roberts, S. G., Baranova, J., Blythe, J., Drew, P., Floyd, S., Gisladottir, R. S., Kendrick, K. H., Levinson, S. C., Manrique, E., Rossi, G., & Enfield, N. J. (2015). Universal principles in the repair of communication problems. *PLOS ONE*, *10*(9), e0136100. https://doi.org/10.1371/journal.pone.0136100

Elder, C.-H., & Haugh, M. (2024). The role of inference and inferencing in pragmatic models of communication. *Journal of Pragmatics*, *229*, 71–76. https://doi.org/10.1016/j.pragma.2024.06.002

Farkas, D. F., & Bruce, K. B. (2010). On reacting to assertions and polar questions. *Journal of Semantics*, *27*(1), 81–118. https://doi.org/10.1093/jos/ffp010

Fedorenko, E., Piantadosi, S.T., & Gibson, E.A.F. (2024). Language is primarily a tool for communication rather than thought. *Nature*, 630, 575–586. https://doi.org/10.1038/s41586-024-07522-w

Fedzechkina, M., Jaeger, T. F., & Newport, E. L. (2012). Language learners restructure their input to facilitate efficient communication. *Proceedings of the National Academy of Sciences*, *109*(44), 17897–17902. https://doi.org/10.1073/pnas.1215776109

Franke, M., & Degen, J. (2016). Reasoning in reference games: Individual- vs. population-level probabilistic modeling. *PLoS ONE*, 11(5), e0154854. https://doi.org/10.1371/journal.pone.0154854

Frank, M. C., & Goodman, N. D. (2012). Predicting pragmatic reasoning in language games. *Science*, *336*(6084), 998–998. https://doi.org/10.1126/science.1218633

Franke, M. (2011). Quantity implicatures, exhaustive interpretation, and rational conversation. *Semantics and Pragmatics*, *4*, Article 1. https://doi.org/10.3765/sp.4.1

Franke, M., & Jäger, G. (2016). Probabilistic pragmatics, or why Bayes' rule is probably important for pragmatics. *Zeitschrift für Sprachwissenschaft*, *35*(1), 3–44. https://doi.org/10.1515/zfs-2016-0002

Friston, K. J. (2010). The free-energy principle: A unified brain theory? Nature Reviews Neuroscience, 11(2), 127–138. https://doi.org/10.1038/nrn2787

Friston, K. J., & Frith, C. D. (2015a). Active inference, communication and hermeneutics. *Cortex*, *68*, 129–143. https://doi.org/10.1016/j.cortex.2015.03.025

Friston, K. J., & Frith, C. D. (2015b). A duet for one. *Consciousness and Cognition*, *36*, 390–405. https://doi.org/10.1016/j.concog.2014.12.003

Friston, K. J., Parr, T., Yufik, Y., Sajid, N., Price, C. J., & Holmes, E. (2020). Generative models, linguistic communication and active inference. *Neuroscience & Biobehavioral Reviews*, *118*, 42–64. https://doi.org/10.1016/j.neubiorev.2020.07.005

Gallagher, S., & Allen, M. (2018). Active inference, enactivism and the hermeneutics of social cognition. *Synthese*, *195*, 2627–2648. https://doi.org/10.1007/s11229-016-1269-8

Garrod, S., & Pickering, M. J. (2004). Why is conversation so easy? Trends in Cognitive Sciences, 8(1), 8-11. https://doi.org/10.1016/j.tics.2003.10.016
Gibson, E., Futrell, R., Piantadosi, S. T., Dautriche, I., Mahowald, K., Bergen, L., & Levy, R. (2019). How efficiency shapes human language. *Trends in cognitive sciences*, *23*(5), 389–407. https://doi.org/10.1016/j.tics.2019.02.003
Goodman, N. D., & Frank, M. C. (2016). Pragmatic language interpretation as probabilistic inference. *Trends in cognitive sciences*, *20*(11), 818–829. https://doi.org/10.1016/j.tics.2016.08.005
Goodman, N. D., & Stuhlmüller, A. (2013). Knowledge and implicature: Modeling language understanding as social cognition. *Topics in Cognitive Science*, *5*(1), 173–184. https://doi.org/10.1111/tops.12007
Hawkins, R. D., Frank, M. C., & Goodman, N. D. (2020). Characterizing the dynamics of learning in repeated reference games. *Cognitive Science*, *44*(6), e12845. https://doi.org/10.1111/cogs.12845
Hawkins, R. D., Franke, M., Frank, M. C., Goldberg, A. E., Smith, K., Griffiths, T. L., & Goodman, N. D. (2023). From partners to populations: A hierarchical Bayesian account of coordination and convention. *Psychological Review*, 130(4), 977–1016. https://doi.org/10.1037/rev0000348
Hipólito, I., & van Es, T. (2022). Enactive-dynamic social cognition and active inference. *Frontiers in psychology*, *13*, 855074. https://doi.org/10.3389/fpsyg.2022.855074
Jaeger, T. F. (2010). Redundancy and reduction: Speakers manage syntactic information density. *Cognitive Psychology*, *61*(1), 23–62. https://doi.org/10.1016/j.cogpsych.2010.02.002
Lee, J. J., & Pinker, S. (2010). Rationales for indirect speech: The theory of the strategic speaker. *Psychological Review*, *117*(3), 785–807. https://doi.org/10.1037/a0019688
Levinson, S. C. (2016). Turn-taking in human communication: Origins and implications for language processing. *Trends in cognitive sciences*, *20*(1), 6–14. https://doi.org/10.1016/j.tics.2015.10.010
Lewis, D. (1979). Scorekeeping in a language game. *Journal of Philosophical Logic*, *8*(1), 339–359. https://doi.org/10.1007/bf00258436
McCarty, M. J., Murphy, E., Scherschligt, X., Woolnough, O., Morse, C. W., Snyder, K., Mahon, B. Z., & Tandon, N. (2023). Intraoperative cortical localization of music and language reveals signatures of structural complexity in posterior temporal cortex. *iScience*, 26(7), 107223. https://doi.org/10.1016/j.isci.2023.107223
Metzing, C., & Brennan, S. E. (2003). When conceptual pacts are broken: Partner-specific effects on the comprehension of referring expressions. *Journal of memory and language*, *49*(2), 201–213. https://doi.org/10.1016/s0749-596x(03)00028-7
Millidge, B., Tschantz, A., & Buckley, C. L. (2021). Whence the expected free energy? Neural Computation, 33(2), 447-482. https://doi.org/10.1162/neco_a_01354
Murphy, E. (2019). No country for Oldowan men: emerging factors in language evolution. *Frontiers in Psychology*, *10*, 1448. https://doi.org/10.3389/fpsyg.2019.01448
Murphy, E. (2020). Language design and communicative competence: The minimalist perspective. *Glossa: a journal of general linguistics*, *5*(1), 2. 1–20. https://doi.org/10.5334/gjgl.1081
Murphy, E. (2021). *Linguistic representation and processing of copredication.* PhD dissertation. University College London.
Murphy, E. (2023). The citadel itself: Defending semantic internalism. *Global Philosophy*, *33*, 7. https://doi.org/10.1007/s10516-023-09669-z
Murphy, E. (2024). ROSE: A neurocomputational architecture for syntax. *Journal of Neurolinguistics*, *70*, 101180.
Murphy, E. (2025). ROSE: A Universal Neural Grammar. *Cognitive Neuroscience*, *16*(1–4), 49–80.
Murphy, E. (2026). Formal properties of language as constraints on neural dynamics. arXiv:2609.14384.
Murphy, E., Holmes, E., & Friston, K. (2024a). Natural language syntax complies with the free-energy principle. *Synthese*, *203*, 154. https://doi.org/10.1007/s11229-024-04566-3
Murphy, E., Rollo, P.S., Segaert, K., Hagort, P., & Tandon, N. (2024). Multiple dimensions of syntactic structure are resolved earliest in posterior temporal cortex. *Progress in Neurobiology*, *241*, 102669. https://doi.org/10.1016/j.pneurobio.2024.102669
Murphy, E., Woolnough, O., Rollo, P. S., Roccaforte, Z. J., Segaert, K., Hagoort, P., & Tandon, N. (2022). Minimal Phrase Composition Revealed by Intracranial Recordings. *The Journal of Neuroscience*, *42*(15), 3216–3227.
Murray, S. E. (2014). Varieties of update. *Semantics and Pragmatics*, *7*, Article 2, 1–53. https://doi.org/10.3765/sp.7.2

Noveck, I. A., & Reboul, A. (2008). Experimental pragmatics: A Gricean turn in the study of language. *Trends in cognitive sciences*, *12*(11), 425–431. https://doi.org/10.1016/j.tics.2008.02.009
Parr, T., Pezzulo, G., & Friston, K. J. (2022). *Active inference: The free energy principle in mind, brain, and behavior*. The MIT Press.
Pate, J. K., & Goldwater, S. (2015). Talkers account for listener and channel characteristics to communicate efficiently. *Journal of memory and language*, *78*, 1–17. https://doi.org/10.1016/j.jml.2014.10.003
Pickering, M. J., & Garrod, S. (2004). Toward a mechanistic psychology of dialogue. *Behavioral and Brain Sciences*, *27*(2), 169–190. https://doi.org/10.1017/s0140525x04000056
Pinker, S., Nowak, M. A., & Lee, J. J. (2008). The logic of indirect speech. *Proceedings of the National Academy of Sciences*, *105*(3), 833–838. https://doi.org/10.1073/pnas.0707192105
Qing, C., & Franke, M. (2015). Variations on a Bayesian theme: Comparing Bayesian models of referential reasoning. *Linguistics and Philosophy*, *38*, 165–201. https://doi.org/10.1007/s10988-014-9168-y
Ramstead, M. J. D., Veissière, S. P. L., & Kirmayer, L. J. (2016). Cultural affordances: Scaffolding local worlds through shared intentionality and regimes of attention. *Frontiers in psychology*, *7*, 1090. https://doi.org/10.3389/fpsyg.2016.01090
Roberts, C. (2012). Information structure in discourse: Towards an integrated formal theory of pragmatics. *Semantics and Pragmatics*, *5*, Article 6, 1–69. https://doi.org/10.3765/sp.5.6
Sajid, N., Ball, P. J., Parr, T., & Friston, K. J. (2021). Active inference: Demystified and compared. *Neural Computation*, *33*(3), 674–712. https://doi.org/10.1162/neco_a_01357
Sauerland, U., & Schumacher, P. B. (2016). Pragmatic processing: A review. *PLOS Biology*, *14*(12), e1002601. https://doi.org/10.1371/journal.pbio.1002601
Schegloff, E. A., Jefferson, G., & Sacks, H. (1977). The preference for self-correction in the organization of repair in conversation. *Language*, *53*(2), 361–382. https://doi.org/10.2307/413107
Schegloff, E. A., & Sacks, H. (1973). Opening up closings. *Semiotica*, *8*(4), 289–327. https://doi.org/10.1515/semi.1973.8.4.289
Schuster, S., & Degen, J. (2020). I know what you're probably going to say: Listener adaptation to variable use of uncertainty expressions. *Cognition*, *203*, 104285. https://doi.org/10.1016/j.cognition.2020.104285
Scott-Phillips, T. C. (2015). *Speaking our minds: why human communication is different, and how language evolved to make it special*. Palgrave Macmillan.
Smith, R., Friston, K. J., & Whyte, C. J. (2022). A step-by-step tutorial on active inference and its application to empirical data. *Journal of Mathematical Psychology*, *107*, 102632. https://doi.org/10.1016/j.jmp.2021.102632
Sperber, D., & Wilson, D. (1995). *Relevance: communication and cognition*. Blackwell.
Tison, R., & Poirier, P. (2021). Active inference and cooperative communication: An ecological alternative to the alignment view. *Frontiers in psychology*, *12*, 708780. https://doi.org/10.3389/fpsyg.2021.708780
Unger, C., & Buschmeier, H. (2025). *Conversational implicatures: modeling relevance theory probabilistically*. https://arxiv.org/abs/2509.22354
Vasil, J., Badcock, P. B., Constant, A., Friston, K. J., & Ramstead, M. J. D. (2020). A world unto itself: Human communication as active inference. *Frontiers in psychology*, *11*, 417. https://doi.org/10.3389/fpsyg.2020.00417
Veissière, S. P. L., Constant, A., Ramstead, M. J. D., Friston, K. J., & Kirmayer, L. J. (2020). Thinking through other minds: A variational approach to cognition and culture. *Behavioral and Brain Sciences*, *43*, e90. https://doi.org/10.1017/s0140525x19001213
Wilson, D., & Sperber, D. (2004). Relevance theory. In L.R. Horn & G. Ward (Eds.), *The Handbook of Pragmatics* (pp. 607–632). Blackwell.
Yoon, E. J., Tessler, M. H., Goodman, N. D., & Frank, M. C. (2020). Polite speech emerges from competing social goals. *Open Mind*, *4*, 71–87. https://doi.org/10.1162/opmi_a_00035